\pdfoutput=1
\documentclass[9pt,twocolumn,twoside]{pnas-new_arxiv}

\setboolean{displaywatermark}{false}

\usepackage{multirow}
\usepackage{wrapfig}
\usepackage{booktabs}
\usepackage{float}

\makeatletter
\def\thickhline{%
  \noalign{\ifnum0=`}\fi\hrule \@height \thickarrayrulewidth \futurelet
   \reserved@a\@xthickhline}
\def\@xthickhline{\ifx\reserved@a\thickhline
               \vskip\doublerulesep
               \vskip-\thickarrayrulewidth
             \fi
      \ifnum0=`{\fi}}
\makeatother

\newlength{\thickarrayrulewidth}
\templatetype{pnasresearcharticle} 

\title{Deep Neural Networks Improve Radiologists' Performance in Breast Cancer Screening} 

\author[b]{Nan Wu}
\author[b]{Jason Phang} 
\author[b]{Jungkyu Park}
\author[b]{Yiqiu Shen}
\author[b]{Zhe Huang}
\author[h*]{Masha Zorin}
\author[i]{Stanis\l{}aw Jastrz\k{e}bski}
\author[b]{Thibault F\'{e}vry}
\author[f]{Joe Katsnelson}
\author[a]{Eric Kim}
\author[a]{Stacey Wolfson}
\author[a]{Ujas Parikh}
\author[a]{Sushma Gaddam}
\author[a]{Leng Leng Young Lin}
\author[j*]{Kara Ho}
\author[a]{Joshua D. Weinstein}
\author[a,d]{Beatriu Reig}
\author[a,d]{Yiming Gao}
\author[a,d]{Hildegard Toth}
\author[a,d]{Kristine Pysarenko}
\author[a,d]{Alana Lewin}
\author[a,d]{Jiyon Lee}
\author[a]{Krystal Airola}
\author[a]{Eralda Mema}
\author[a]{Stephanie Chung}
\author[a]{Esther Hwang}
\author[a]{Naziya Samreen}
\author[a,d,e]{S. Gene Kim}
\author[a,d]{Laura Heacock}
\author[a,d,e]{Linda Moy}
\author[b,c,g]{Kyunghyun Cho}
\author[a,b,e,1]{Krzysztof J. Geras}

\affil[a]{Department of Radiology, New York University School of Medicine, 660 First Ave, New York, NY 10016, USA}
\affil[b]{Center for Data Science, New York University, 60 5th Ave, New York, NY 10011, USA}
\affil[c]{Courant Institute of Mathematical Sciences, New York University, 251 Mercer St, New York, NY 10012, USA}
\affil[d]{Perlmutter Cancer Center, NYU Langone Health, 160 E 34th St, New York, NY 10016, USA}
\affil[e]{Center for Advanced Imaging Innovation and Research, NYU Langone Health, 660 First Ave, New York, NY 10016, USA}
\affil[f]{Department of Information Technology, NYU Langone Health, 360 Park Ave South, New York, NY 10010, USA}
\affil[g]{CIFAR Global Scholar}
\affil[h]{Department of Computer Science and Technology, University of Cambridge, William Gates Building, 15 JJ Thomson Ave, Cambridge CB3 0FD, UK}
\affil[i]{Faculty of Mathematics and Information Technologies, Jagiellonian University, \L{}ojasiewicza 6, 30-348 Krak\'ow, Poland}
\affil[j]{SUNY Downstate College of Medicine, 450 Clarkson Ave, New York, NY 11203, USA}

\leadauthor{Wu} 
\significancestatement{Breast cancer is the second leading cancer-related cause of death for women in the US. Although screening mammography has reduced breast cancer mortality, its specificity remains low. The development of deep convolutional neural networks to aid in the evaluation of screening mammography would save significant health care costs. In this work, we propose a novel neural network architecture and an appropriate two-stage training procedure to efficiently handle a large dataset of high-resolution breast mammograms with biopsy-proven labels. We experimentally show that our model is as accurate as an experienced radiologist and that it can improve the accuracy of radiologists’ diagnoses when used as a second reader.}

\authorcontributions{Author contributions: KJG, KC, NW, LM, LH and SGK designed research; KJG, NW, JP, JP, YS, ZH, MZ, TF and SJ performed research; JK optimized the experimental framework; EK, SW, UP, SG, LLYL, KH, JDW, BR, YG, HT, KP, AL, JL, KA, EM, SC, NS, LH and LM collected and analyzed experimental data; KJG, NW, JP, JP and YS wrote the paper.}
\authordeclaration{The authors declare no conflict of interest.}
\correspondingauthor{\textsuperscript{1}To whom correspondence should be addressed. E-mail: k.j.geras@nyu.edu.\\
\textsuperscript{*}Work done while visiting NYU.}

\keywords{deep learning $|$ deep convolutional neural networks $|$ breast cancer screening $|$ mammography} 

\begin{abstract}
We present a deep convolutional neural network for breast cancer screening exam classification, trained and evaluated on over 200,000 exams (over 1,000,000 images). Our network achieves an AUC of 0.895
in predicting whether there is a cancer in the breast, when tested on the screening population. We attribute the high accuracy of our model to a two-stage training procedure, which allows us to use a very high-capacity patch-level network to learn from pixel-level labels alongside a network learning from macroscopic breast-level labels. To validate our model, we conducted a reader study with 14 readers, each reading 720 screening mammogram exams, and find our model to be as accurate as experienced radiologists when presented with the same data. Finally, we show that a hybrid model, averaging probability of malignancy predicted by a radiologist with a prediction of our neural network, is more accurate than either of the two separately. To better understand our results, we conduct a thorough analysis of our network's performance on different subpopulations of the screening population, model design, training procedure, errors, and properties of its internal representations.
\end{abstract}

\dates{This manuscript was compiled on \today}
\doi{\url{www.pnas.org/cgi/doi/10.1073/pnas.XXXXXXXXXX}}

\newcolumntype{K}[1]{>{\centering\arraybackslash}p{#1}}

\begin{document}

\maketitle
\thispagestyle{firststyle}
\ifthenelse{\boolean{shortarticle}}{\ifthenelse{\boolean{singlecolumn}}{\abscontentformatted}{\abscontent}}{}

\dropcap{B}reast cancer is the second leading cancer-related cause of death among women in the US. In 2014, over 39 million screening and diagnostic mammography exams were performed in the US. It is estimated that in 2015 232,000 women were diagnosed with breast cancer and approximately 40,000 died from it \cite{RN53}. Although mammography is the only imaging test that has reduced breast cancer mortality \cite{RN38, RN40, RN41}, there has been discussion regarding the potential harms of screening, including false positive recalls and associated false positive biopsies. The vast majority of the 10--15\% of women asked to return following an inconclusive screening mammogram undergo another mammogram and/or ultrasound for clarification. After the additional imaging exams, many of these findings are determined as benign and only 10--20\% are recommended to undergo a needle biopsy for further work-up. Among these, only 20--40\% yield a diagnosis of cancer \cite{RN43}. Evidently, there is an unmet need to shift the balance of routine breast cancer screening towards more benefit and less harm.

Traditional computer-aided detection (CAD) in mammography is routinely used by radiologists to assist with image interpretation, despite multicenter studies showing these CAD programs do not improve their diagnostic performance \cite{with_and_without_CAD}. These CAD programs typically use handcrafted features to mark sites on a mammogram that appear distinct from normal tissue structures. The radiologist decides whether to recall these findings, determining clinical significance and actionability. Recent developments in deep learning \cite{deep_learning}---in particular, deep convolutional neural networks (CNNs) \cite{convnet, alexnet, vggnet, resnet, densenet}---open possibilities for creating a new generation of CAD-like tools. 

This paper makes several contributions. Primarily, we train and evaluate a set of strong neural networks on a mammography dataset, with biopsy-proven labels, that is of a massive size by the standards of medical image analysis, let alone breast cancer screening. We use two complimentary types of labels: breast-level labels indicating whether there is a benign or malignant finding in each breast, and pixel-level labels indicating the location of biopsied malignant and benign findings.
To quantify the value of pixel-level labels, we compare a model using only breast-level labels 
against a model using both breast-level and pixel-level labels. Our best model, trained on both breast-level and pixel-level labels, achieves an AUC of 0.895 in identifying malignant cases and 0.756 in identifying benign cases on a non-enriched test set reflecting the screening population.

In the reader study, we compared the performance of our best model to that of radiologists and found our model to be as accurate as radiologists both in terms of area under ROC curve (AUC) and area under precision-recall curve (PRAUC). We also found that a hybrid model, taking the average of the probabilities of malignancy predicted by a radiologist and by our neural network, yields more accurate predictions than either of the two separately. This suggests that our network and radiologists learned different aspects of the task and that our model could be effective as a tool providing radiologists a second reader. Finally, we have made the code and weights of our best models available at \href{https://github.com/nyukat/breast_cancer_classifier}{https://github.com/nyukat/breast\_cancer\_classifier}. With this contribution, research groups that are working on improving screening mammography, which may not have access to a large training dataset like ours, will be able to directly use our model in their research or to use our pretrained weights as an initialization to train models with less data. By making our models public, we invite other groups to validate our results and test their robustness to shifts in the data distribution.

\section*{Data}

\begin{figure}[ht]
    \centering
     \begin{tabular}{c c c c }
    \hspace{-2mm}R-CC & \hspace{-4.5mm}L-CC & \hspace{-4.5mm}R-MLO & \hspace{-4.5mm}L-MLO \\
    \hspace{-2mm}\includegraphics[width=0.245\linewidth]{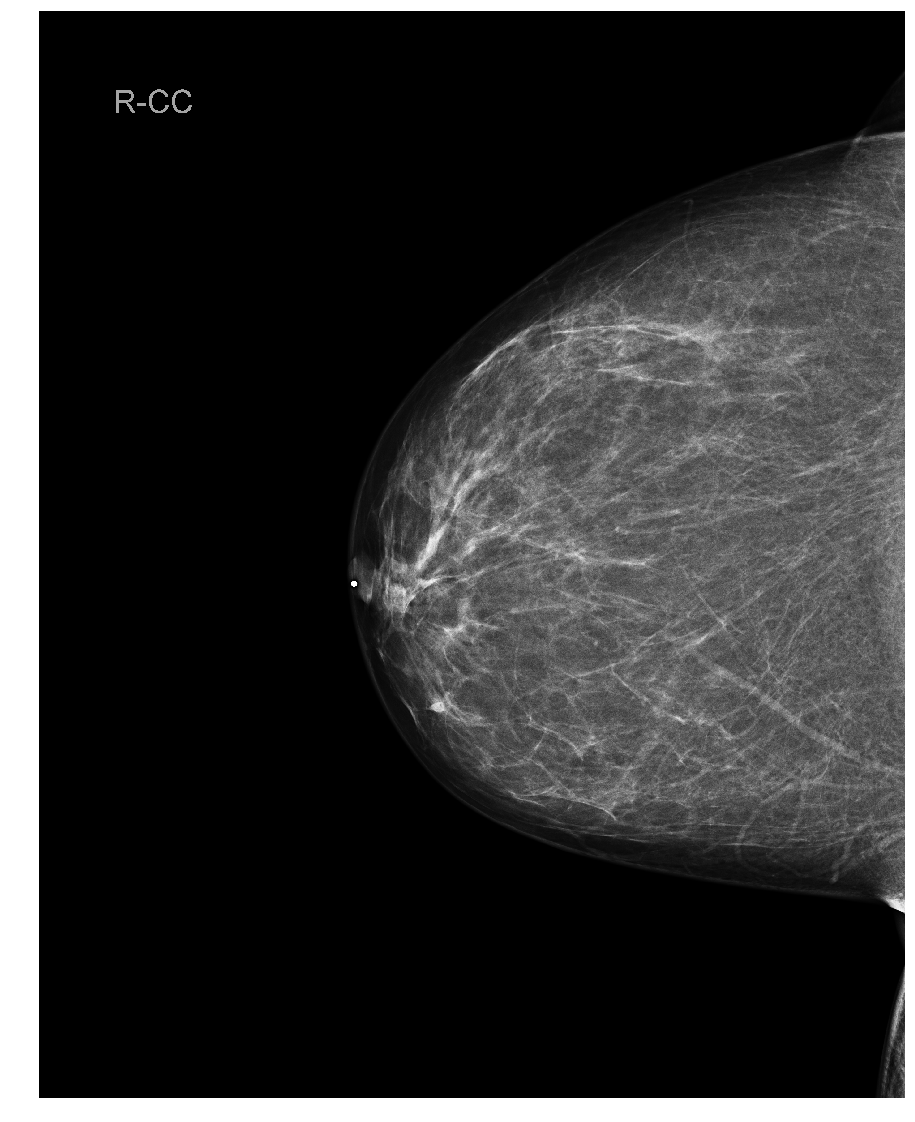} & 
    \hspace{-4.5mm}\includegraphics[width=0.245\linewidth]{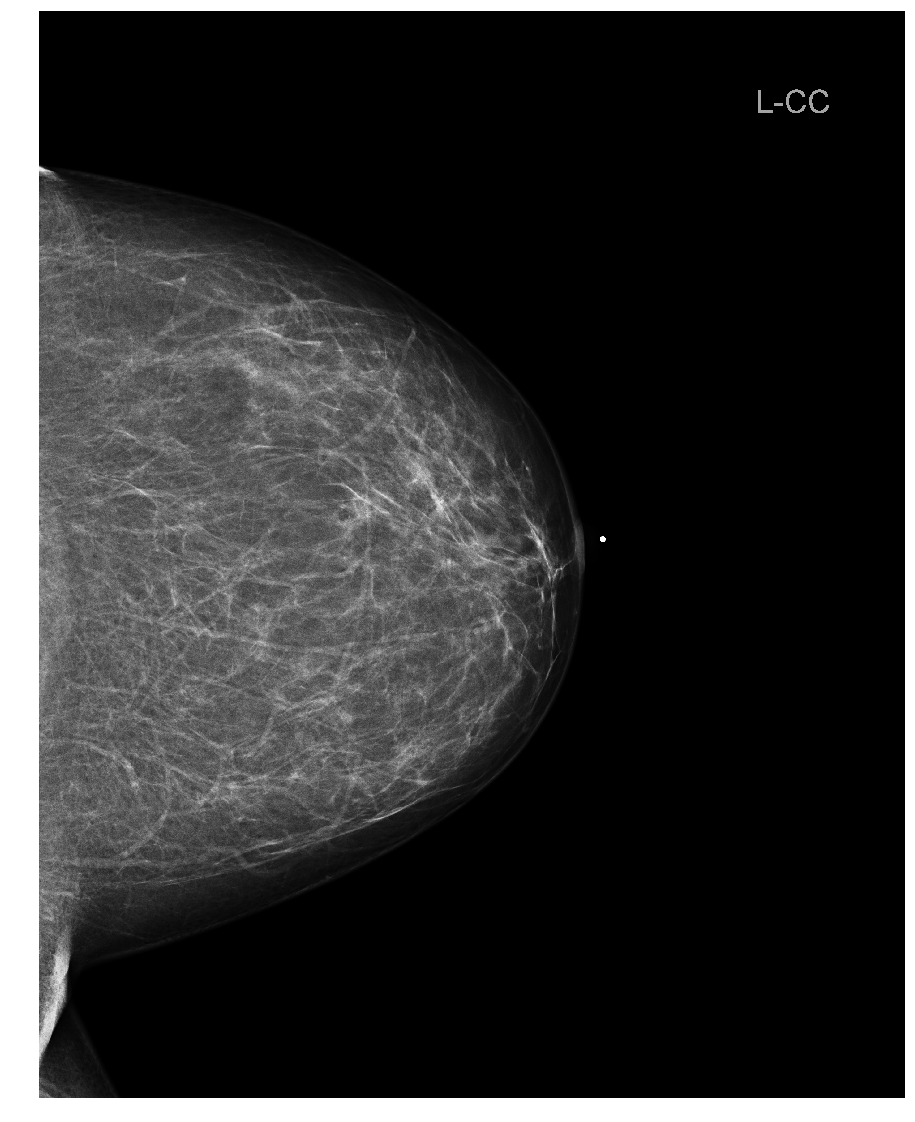} &
    \hspace{-4.5mm}\includegraphics[width=0.245\linewidth]{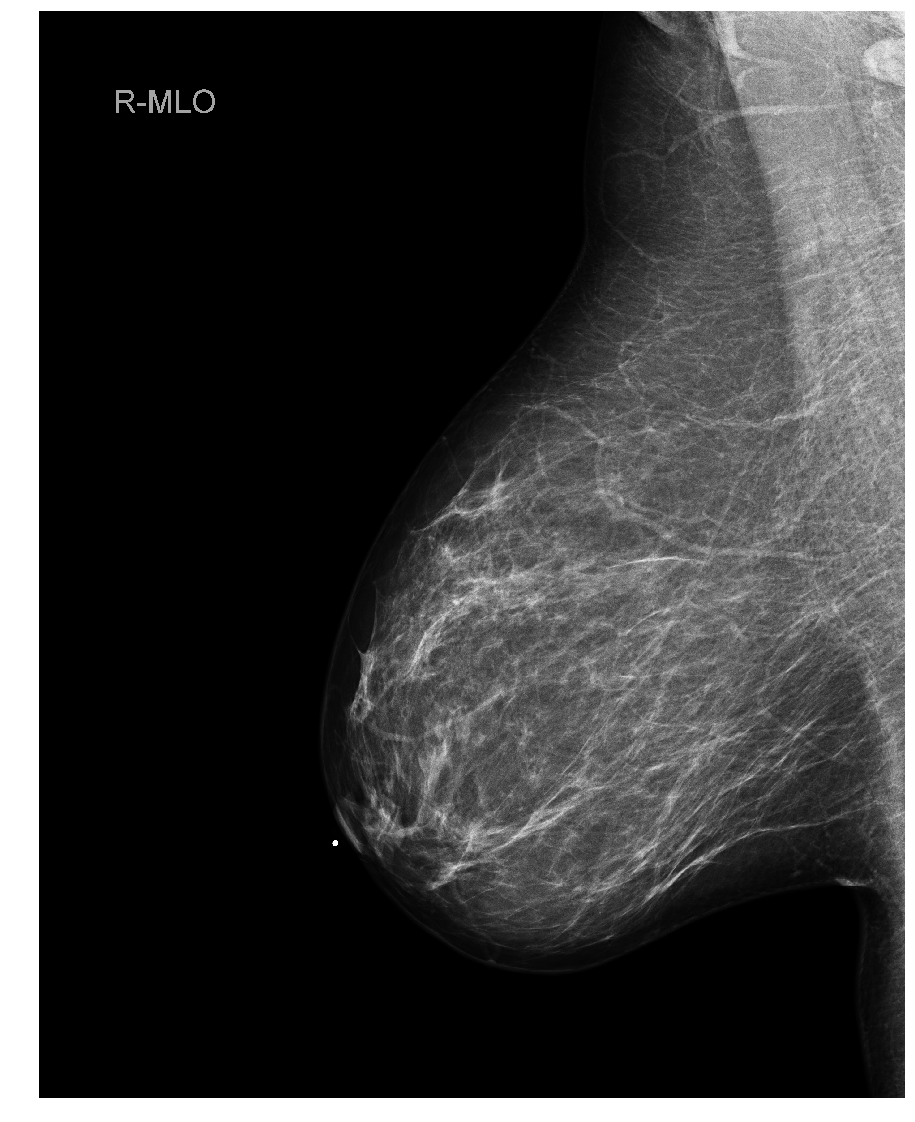} &
    \hspace{-4.5mm}\includegraphics[width=0.245\linewidth]{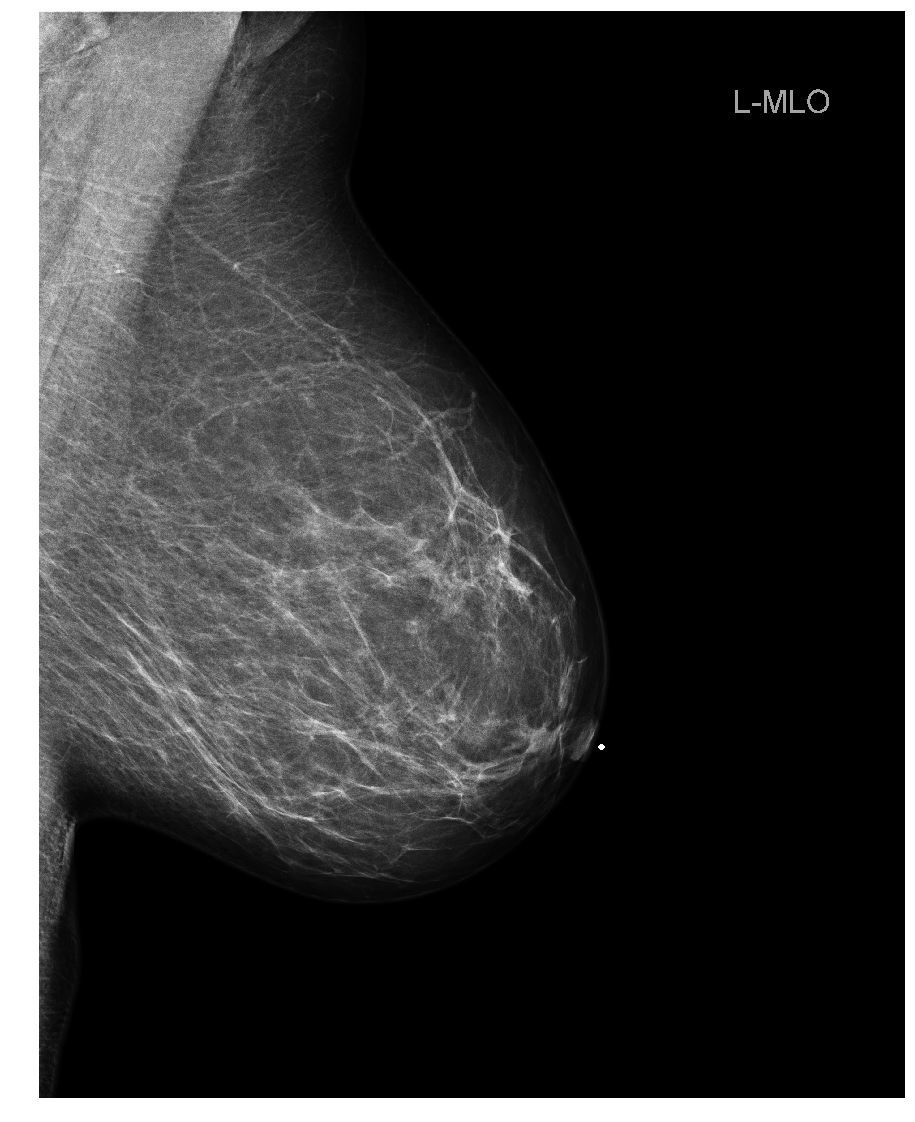} \\ \vspace{-3mm} \\
    \hspace{-2mm}\includegraphics[width=0.245\linewidth]{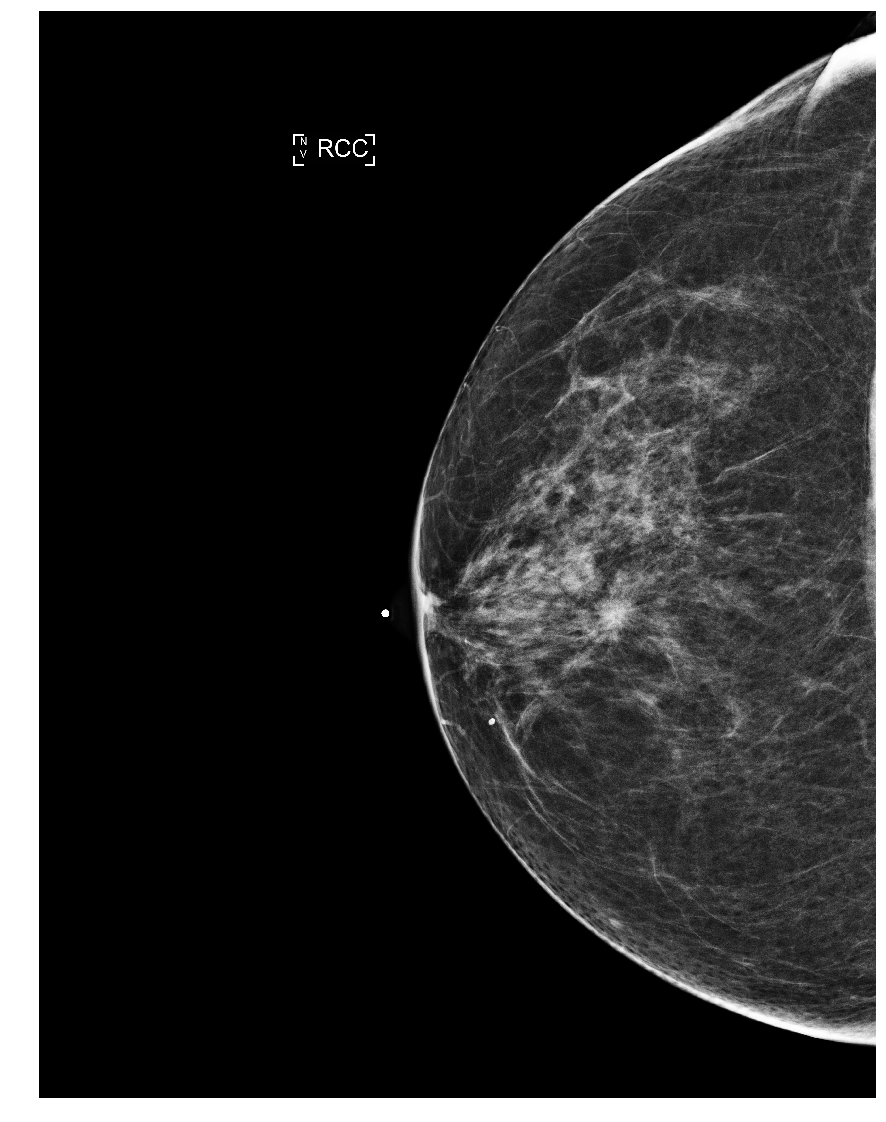} & 
    \hspace{-4.5mm}\includegraphics[width=0.245\linewidth]{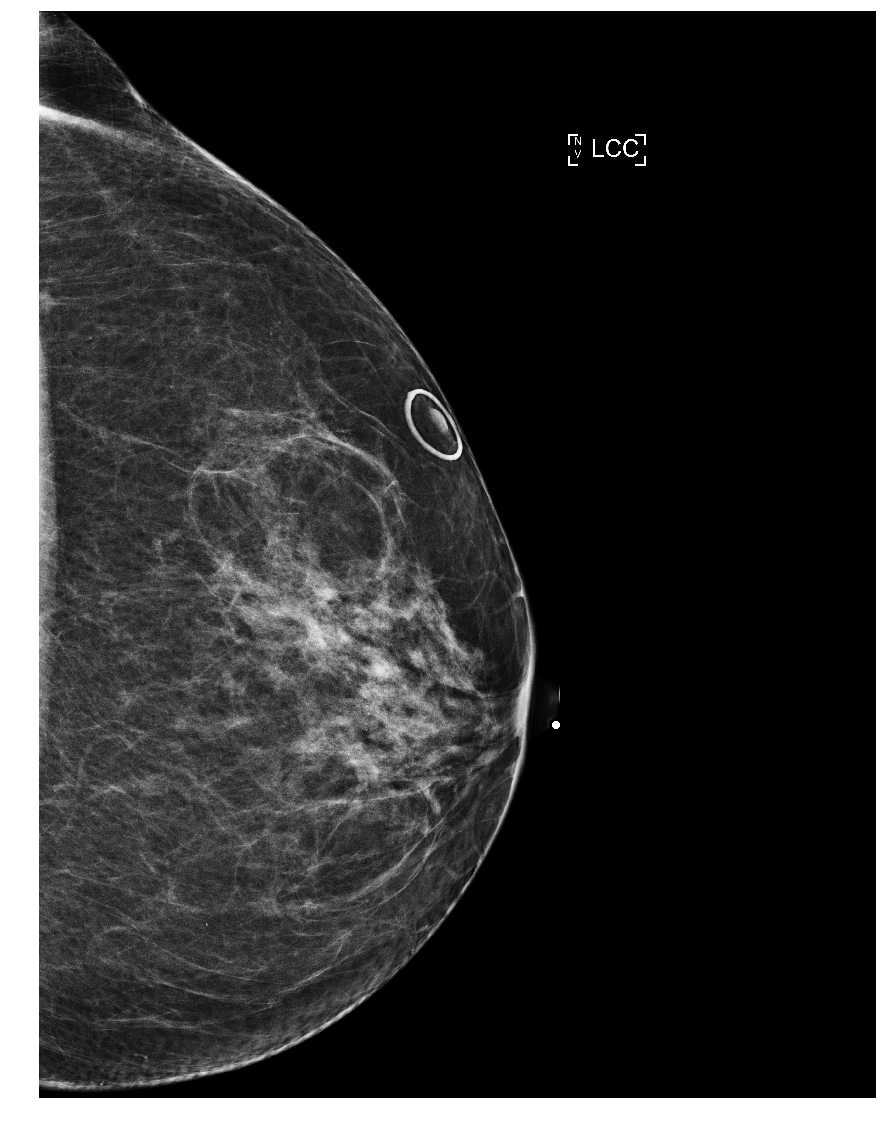} &
    \hspace{-4.5mm}\includegraphics[width=0.245\linewidth]{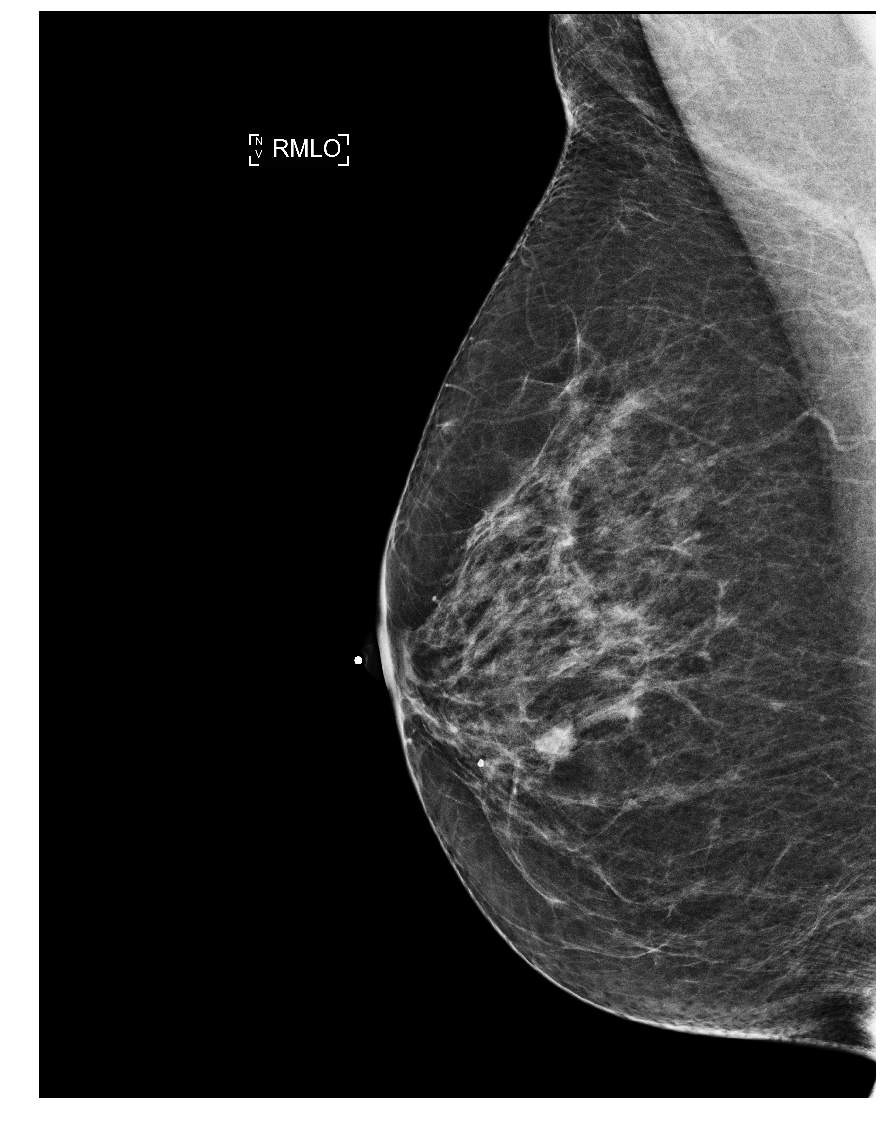} &
    \hspace{-4.5mm}\includegraphics[width=0.245\linewidth]{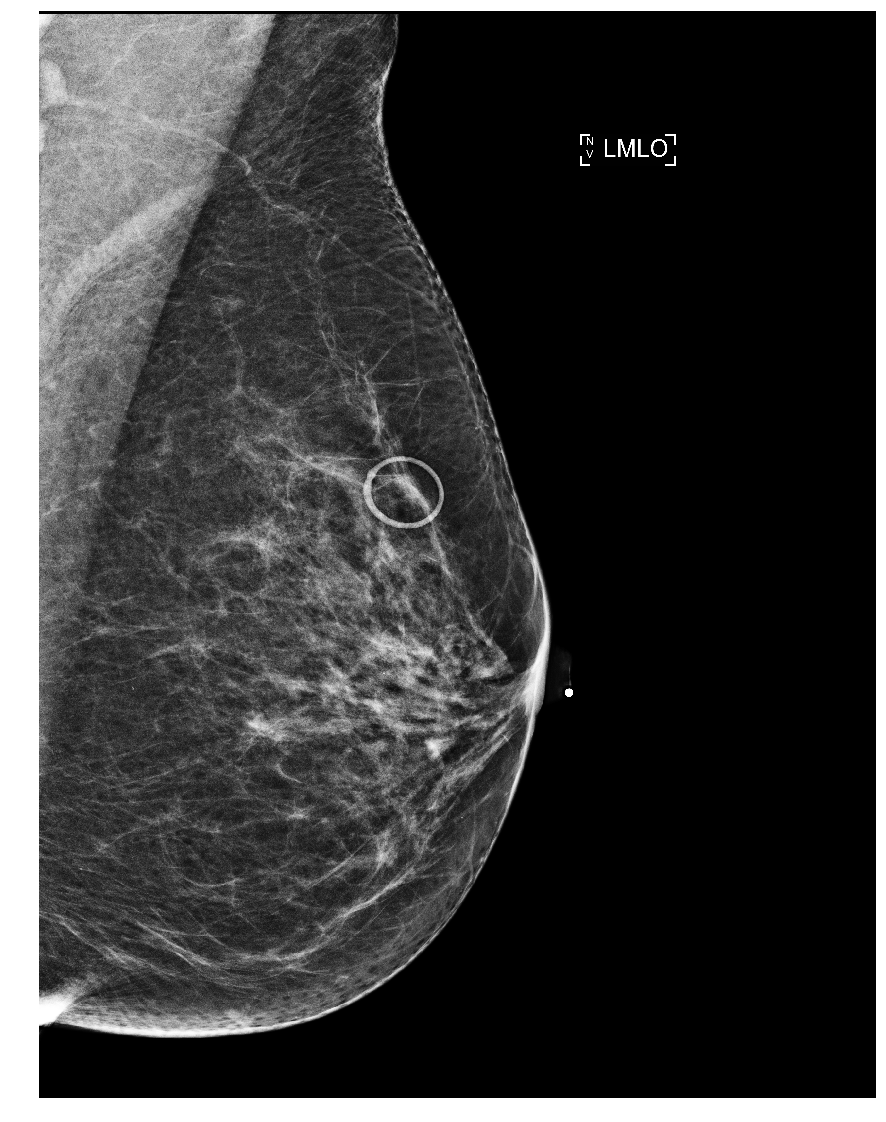} \\ \vspace{-3mm} \\ 
    \hspace{-2mm}\includegraphics[width=0.245\linewidth]{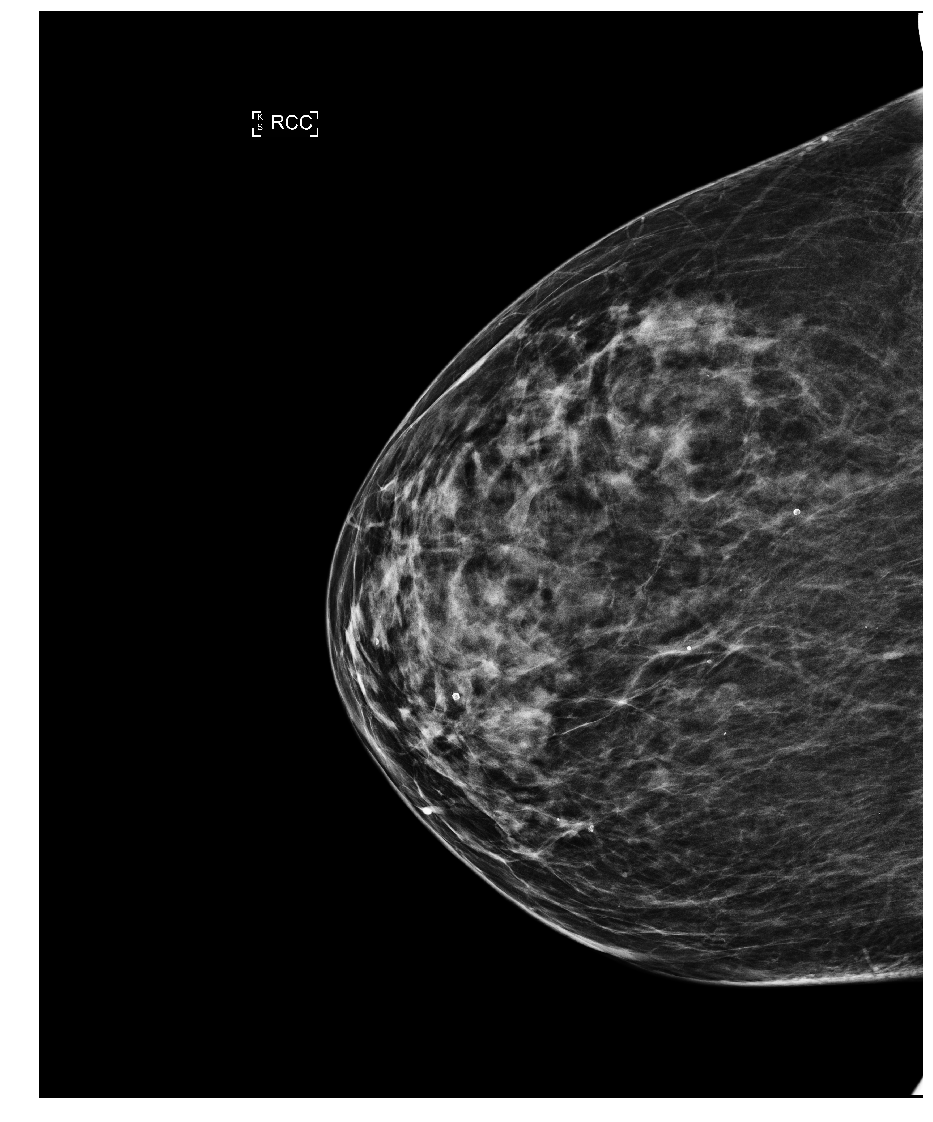} &
    \hspace{-4.5mm}\includegraphics[width=0.245\linewidth]{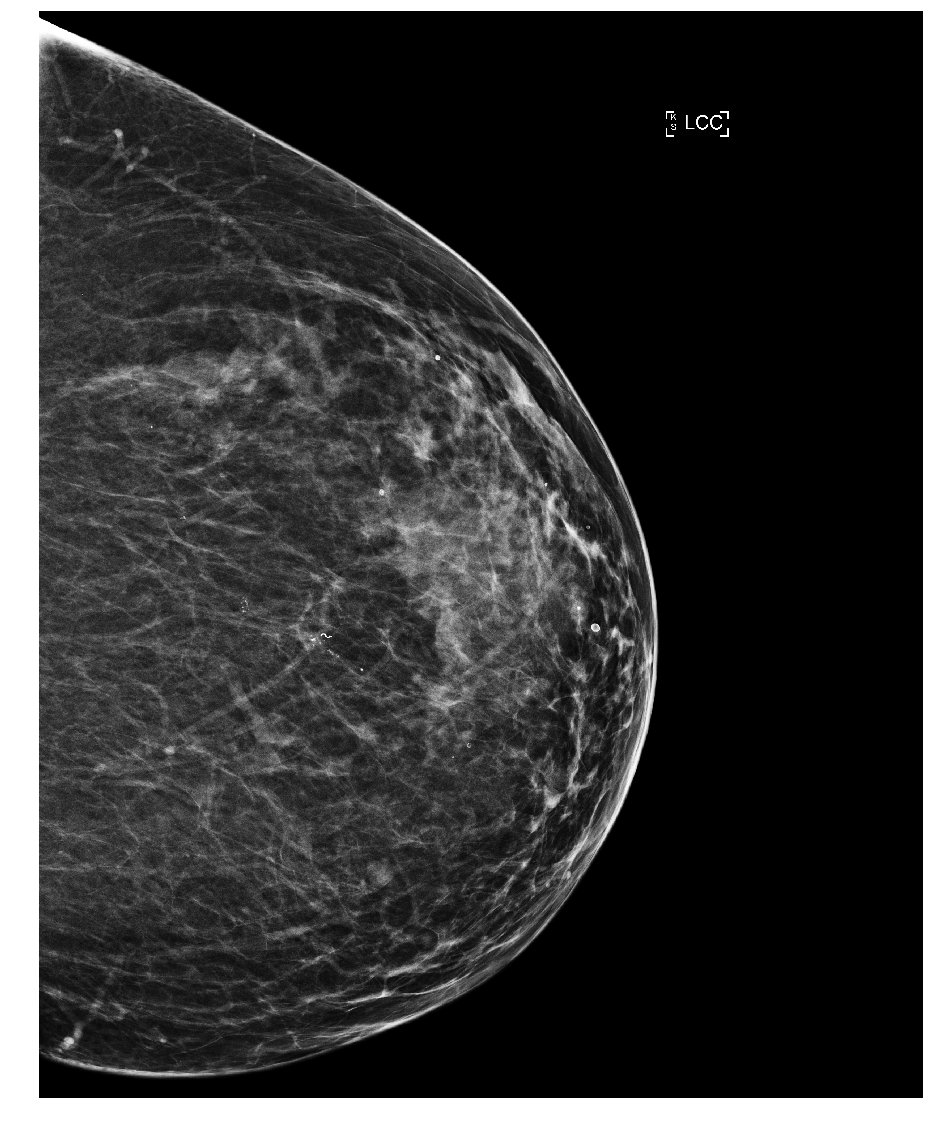} &
    \hspace{-4.5mm}\includegraphics[width=0.245\linewidth]{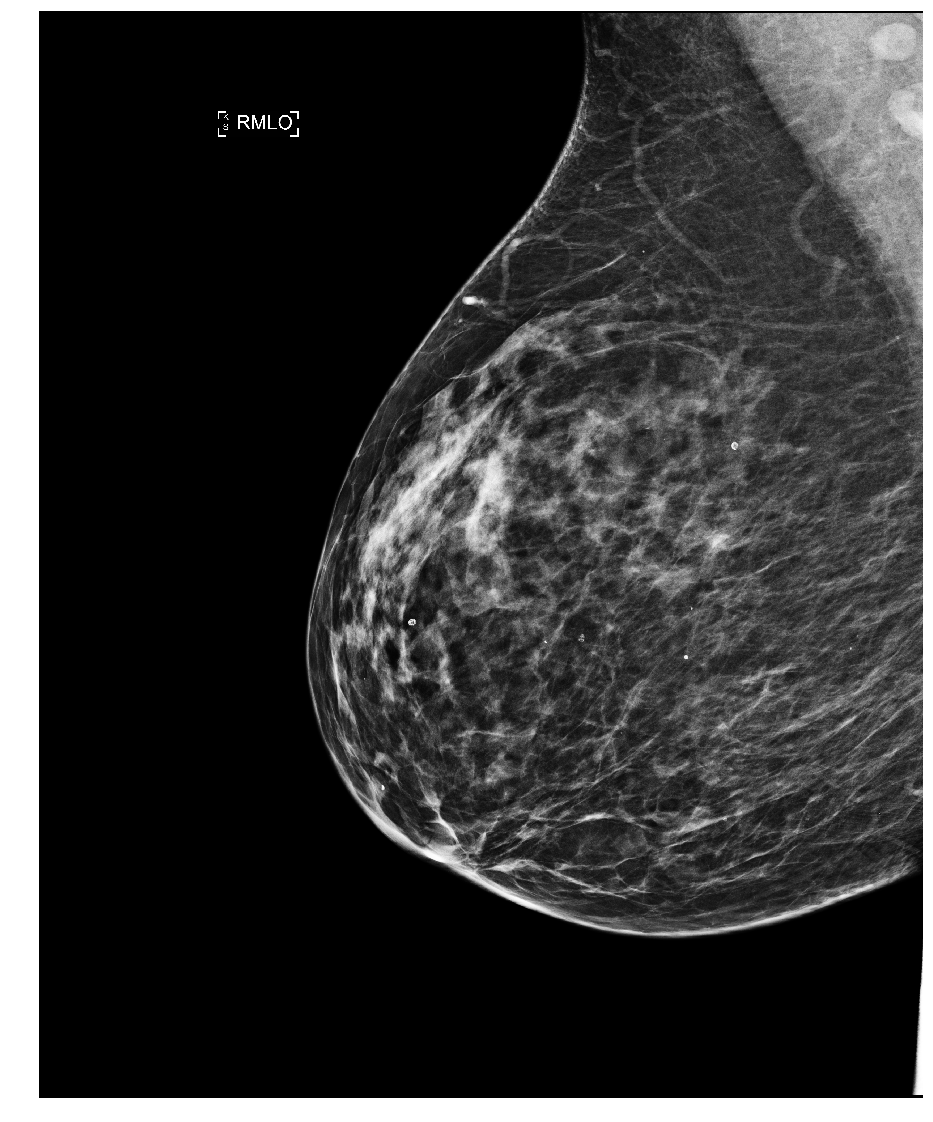} &
    \hspace{-4.5mm}\includegraphics[width=0.245\linewidth]{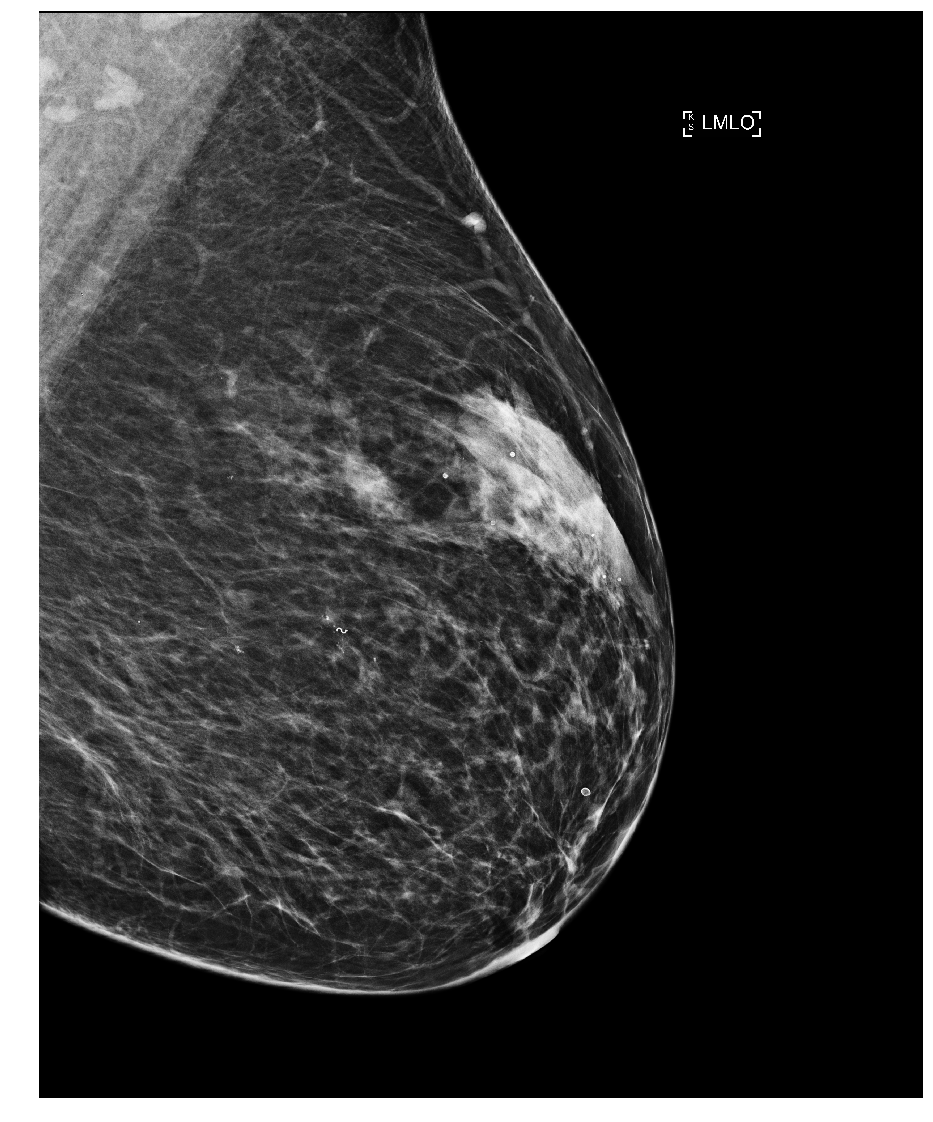} 
    \end{tabular}
    \vspace{-4mm}
    \caption{Examples of breast cancer screening exams. First row: both breasts without any findings; second row: left breast with no findings and right breast with a malignant finding; third row: left breast with a benign finding and right breast with no findings.
    }
    \label{fig:example_exams}
\end{figure}

Our retrospective study was approved by our institutional review board and was compliant with the Health Insurance Portability and Accountability Act. Informed consent was waived. This dataset\footnote{Details of its statistics and how it was extracted can be found in a separate technical report \cite{NYU_dataset}.} is a larger and more carefully curated version of a dataset used in our earlier work \cite{high_resolution, breast_density}. The dataset includes 229,426 digital screening mammography exams (1,001,093 images) from 141,473 patients. Each exam contains at least four images,\footnote{Some exams contain more than one image per view as technologists may need to repeat an image or provide a supplemental view to completely image the breast in a screening examination.} corresponding to the four standard views used in screening mammography: R-CC (right craniocaudal), L-CC (left craniocaudal), R-MLO (right mediolateral oblique) and L-MLO (left mediolateral oblique). A few examples of exams are shown in Figure~\ref{fig:example_exams}. 

To extract labels indicating whether each breast of the patient was found to have malignant or benign findings at the end of the diagnostic pipeline, we relied on pathology reports from biopsies. We have 5,832 exams with at least one biopsy performed within 120 days of the screening mammogram. Among these, biopsies confirmed malignant findings for 985 (8.4\%) breasts and benign findings for 5,556 (47.6\%) breasts. 234 (2.0\%) breasts had both malignant and benign findings. For the remaining screening exams that were not matched with a biopsy, we assigned labels corresponding to the absence of malignant and benign findings in both breasts. 

\begin{figure}[ht]
\begin{minipage}{.22\textwidth}
    \centering
    \begin{tabular}{c c}
    \hspace{-2mm}\includegraphics[width=0.48\linewidth]{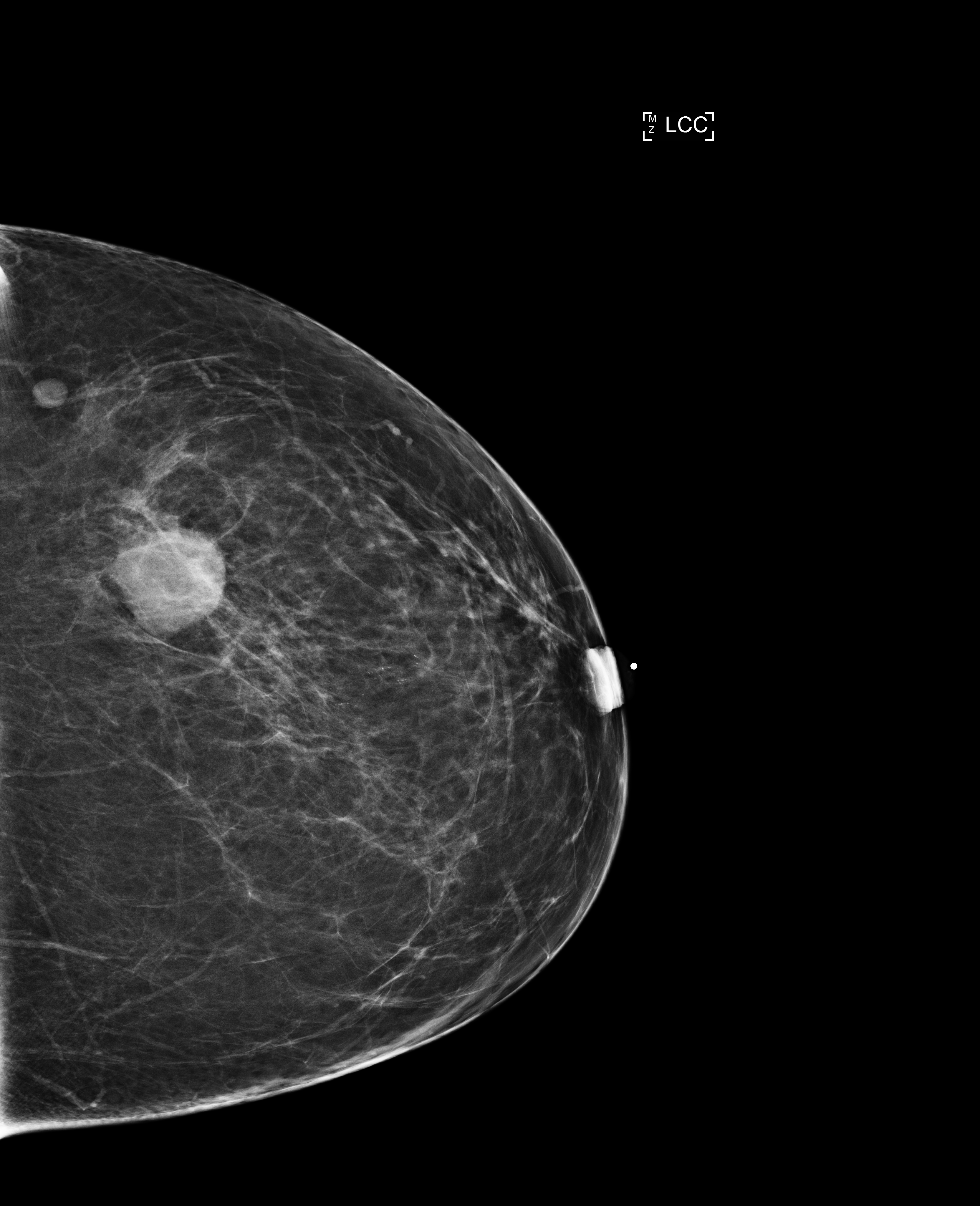} & \hspace{-4mm} \includegraphics[width=0.48\linewidth]{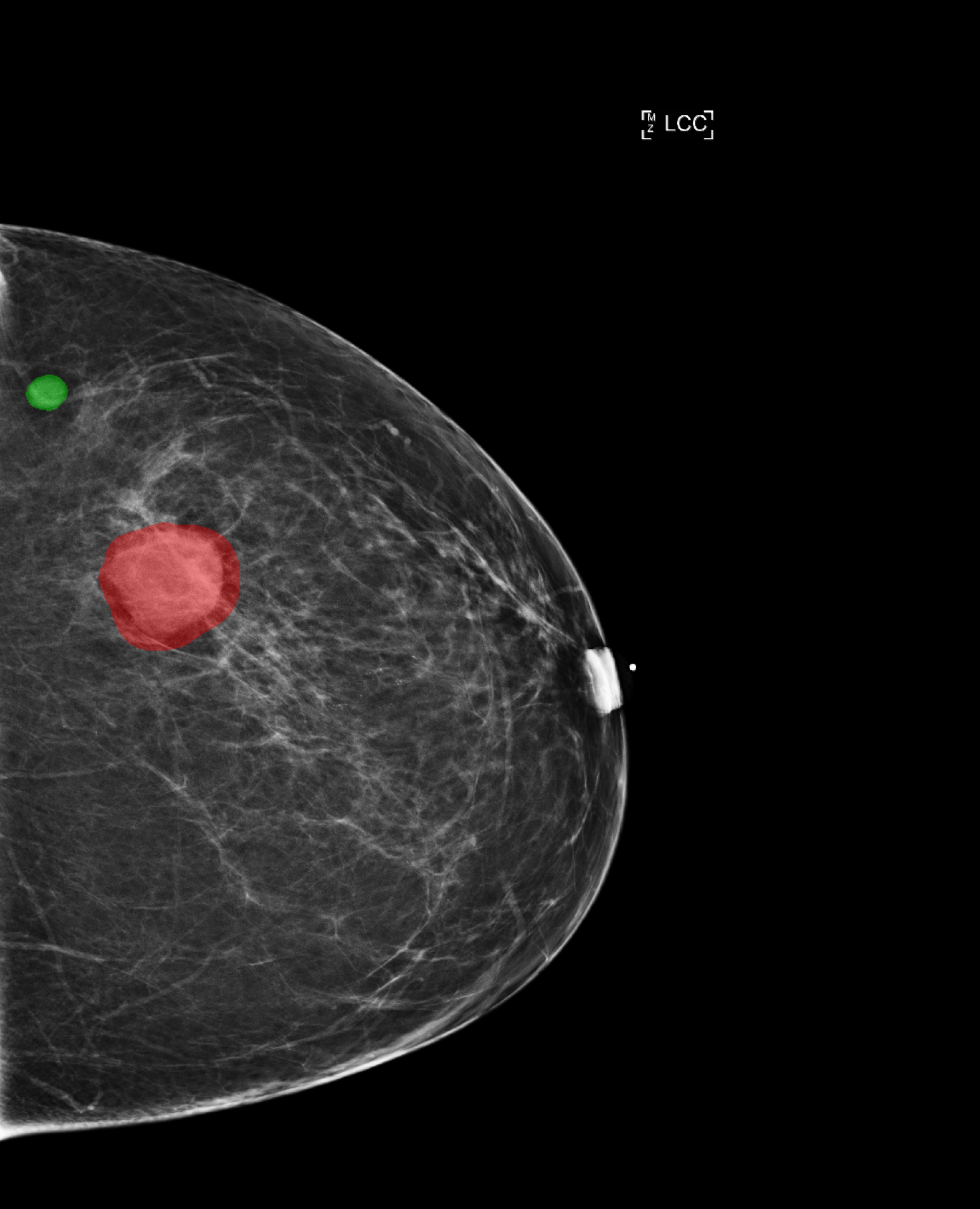} 
    \end{tabular}
    \vspace{-3mm}
    \caption{An example of a segmentation performed by a radiologist. Left: the original image. Right: the image with lesions requiring a biopsy highlighted. The malignant finding is highlighted with red and benign finding with green.}
    \label{fig:example_segmentation}
\end{minipage}
\hfill
\begin{minipage}{.25\textwidth}
    \centering
    \vspace{-1.5mm}
    \includegraphics[width=1\linewidth,trim={2cm 2.2cm 2.15cm 3.5cm},clip]{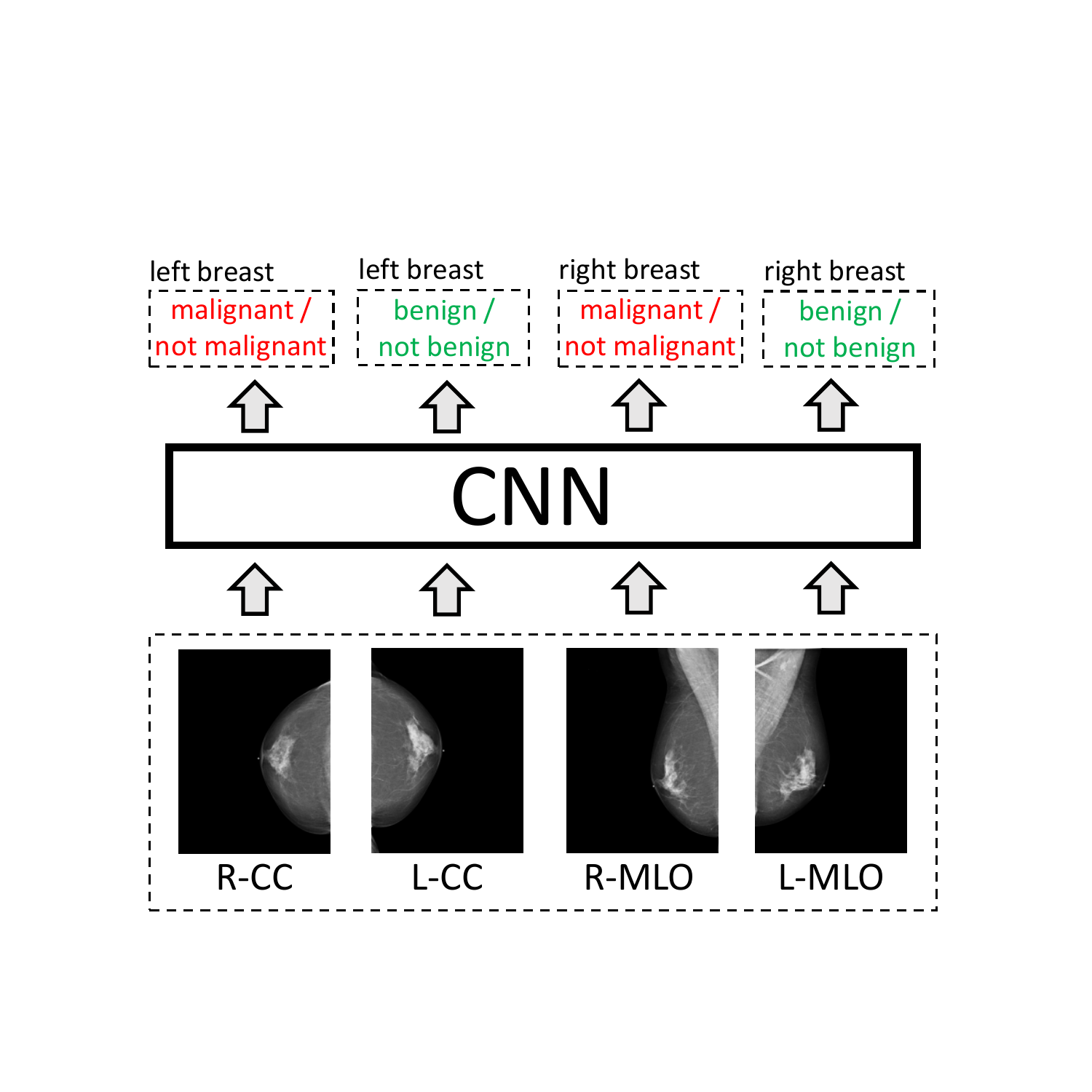}
    \vspace{-7mm}
    \caption{A schematic representation of how we formulated breast cancer exam classification as a learning task.}
    \label{fig:schematic}
\end{minipage}
\vspace{-3mm}
\end{figure}

For all exams matched with biopsies, we asked a group of radiologists (provided with the corresponding pathology reports) to retrospectively indicate the location of the biopsied lesions at a pixel level. An example of such a segmentation is shown in Figure~\ref{fig:example_segmentation}. We found that, according to the radiologists, approximately 32.8\% of exams were mammographically occult, i.e., the lesions that were biopsied were not visible on mammography, even retrospectively, and were identified using other imaging modalities: ultrasound or MRI.

\section*{Deep CNNs for cancer classification}

\subsection*{Problem definition}
For each breast, we assign two binary labels: the absence/presence of malignant findings in a breast, and the absence/presence of benign findings in a breast. With left and right breasts, each exam has a total of four binary labels.
Our goal is to produce four predictions corresponding to the four labels for each exam.
As input, we take four high-resolution images corresponding to the four standard screening mammography views. 
We crop each image to a fixed size of $2677\times1942$ pixels for CC views and $2974\times1748$ pixels for MLO views. See \autoref{fig:schematic} for a schematic representation.

\subsection*{Model architecture}

\begin{figure}[ht]
\begin{minipage}{0.72\linewidth}
    \centering
    \includegraphics[width=0.88\textwidth,trim={2.3cm 2.4cm 2.7cm 3.8cm}, clip]{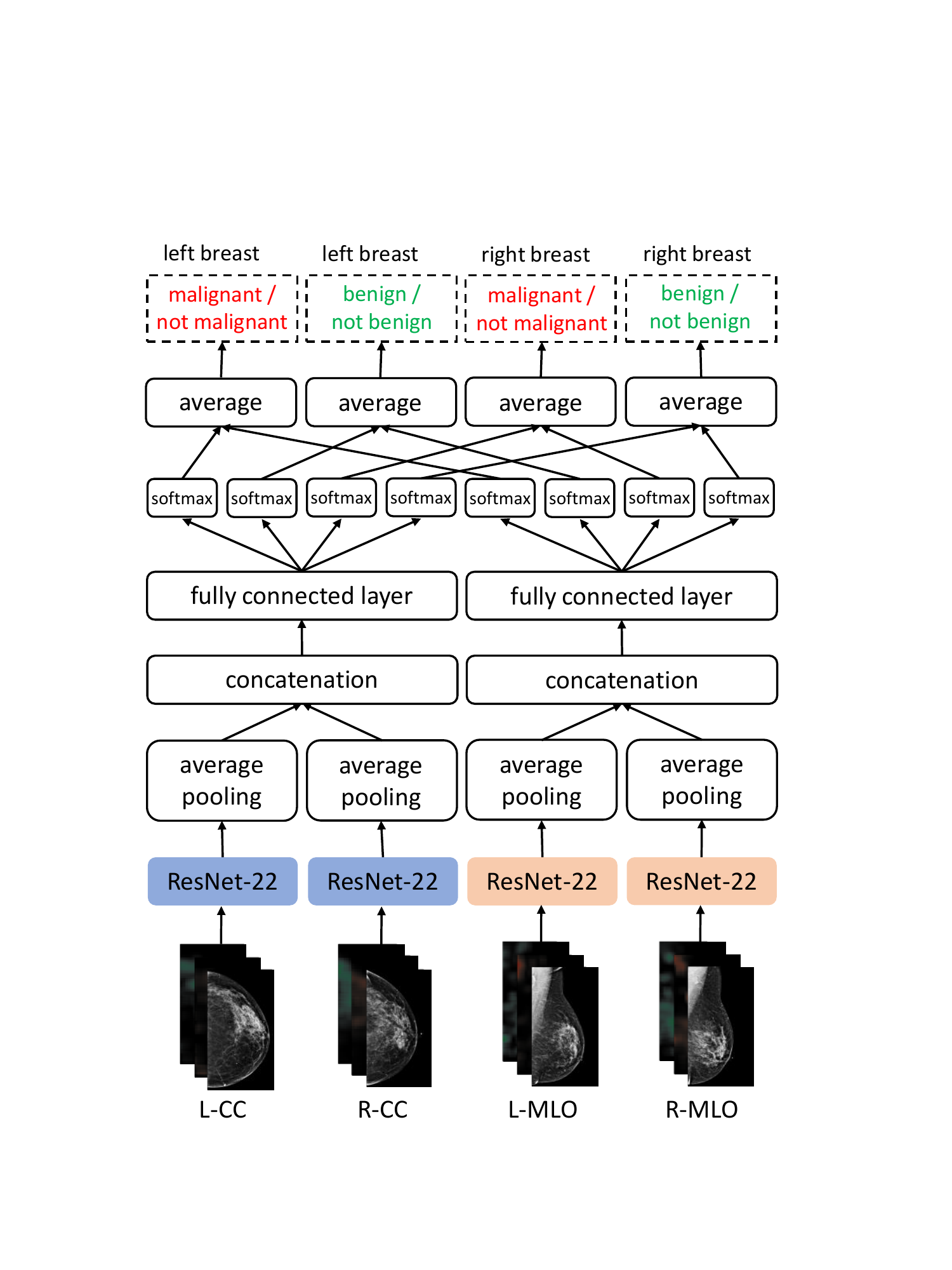}
    \vspace{-2.5mm}
    \caption{
        Architecture of our model.
        Four ResNet-22 columns take the four views as input. 
        The architecture is divided into CC and MLO branches. In each branch, the corresponding left and right representations from the ResNets are individually average-pooled spatially and concatenated, and two fully connected layers are applied to compute the predictions for the four outputs. The predictions are averaged between the CC and MLO branches. Weights are shared between L-CC/R-CC columns and L-MLO/R-MLO columns. When heatmaps are added as additional channels to corresponding inputs, the first layers of the columns are modified accordingly.
    }
    \label{fig:architectures}
\end{minipage}\hfill
\begin{minipage}{0.24\linewidth}
    \centering
    \begin{tabular}{c}
    \includegraphics[width = 0.9\textwidth,trim={23mm 12mm 0mm 0mm}]{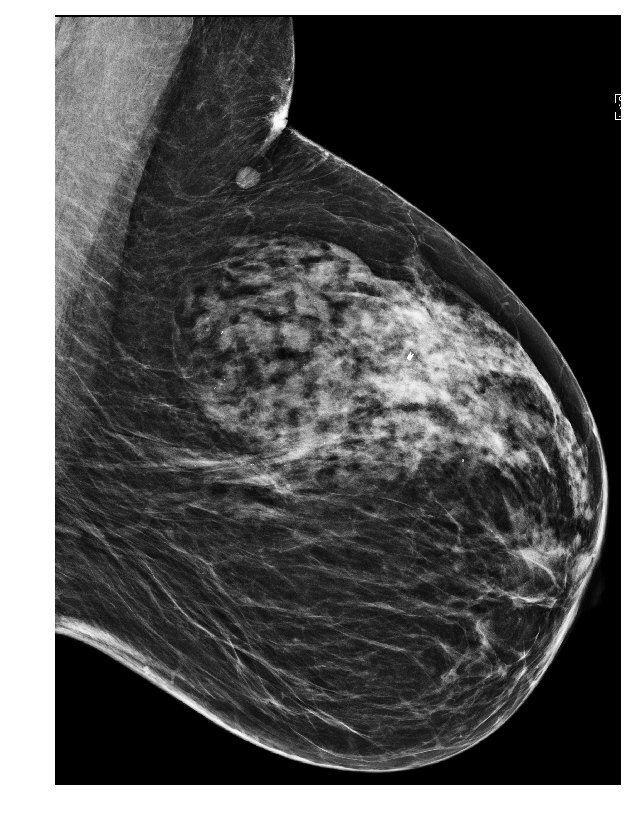}\\
    \includegraphics[width = 0.9\textwidth,trim={23mm 12mm 0mm 0mm}]{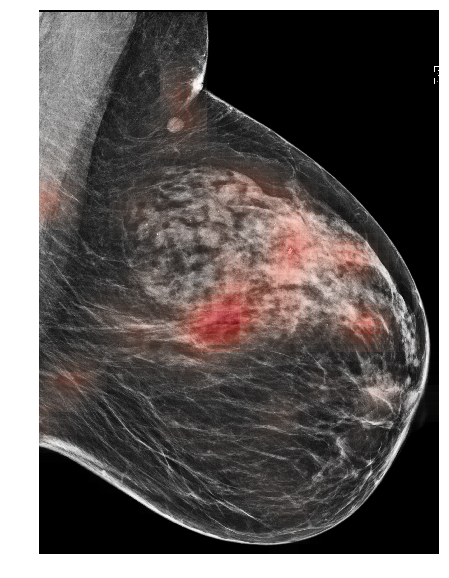}\\ 
    \includegraphics[width = 0.9\textwidth,trim={23mm 12mm 0mm 0mm}]{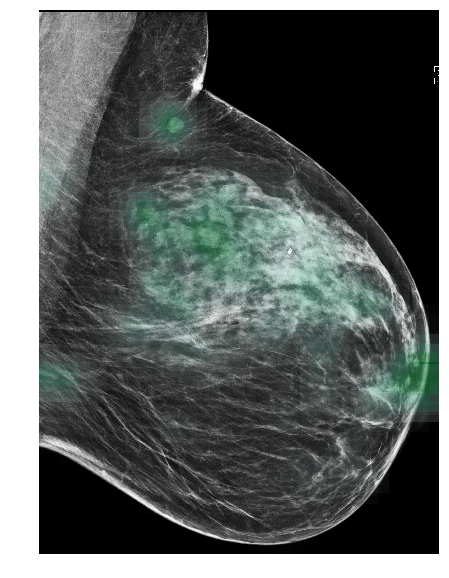}\\
    \end{tabular}
    \vspace{-2mm}
    \caption{The original image, the `malignant` heatmap over the image and the `benign` heatmap over the image. 
    }
    \label{fig:heatmaps}
\end{minipage}
\vspace{-5mm}
\end{figure}

We trained a deep multi-view CNN of architecture shown in \autoref{fig:architectures}, inspired by \cite{high_resolution}. The overall network consists of two core modules: (i) four view-specific columns, each based on the ResNet architecture \cite{resnet} that outputs a fixed-dimension hidden representation for each mammography view, and (ii) two fully connected layers to map from the computed hidden representations to the output predictions. We used four ResNet-22\footnote{\textit{ResNet-22} refers to our version of a 22-layer ResNet, with additional modifications such as a larger kernel in the first convolutional layer. Details can be found in the SI.} columns to compute a 256-dimension hidden representation vector of each view. The columns applied to L-CC/R-CC views share their weights. The columns applied to L-MLO/R-MLO views share their weights too. We concatenate the L-CC and R-CC representations into a 512-dimension vector, and apply two fully connected layers to generate predictions for the four outputs. We do the same for the L-MLO and R-MLO views. We average the probabilities predicted by the CC and MLO branches of the model to obtain our final predictions.

\subsubsection*{Auxiliary patch-level classification model and heatmaps}
    The high resolution of the images and the limited memory of modern GPUs constrain us to use relatively shallow ResNets within our model when using full-resolution images as inputs.
    To further take advantage of the fine-grained detail in mammograms, we trained an auxiliary model to classify $256 \times 256$-pixel patches of mammograms, predicting two labels: the presence or absence of malignant and benign findings in a given patch. The labels for these patches are produced based on the pixel-level segmentations of the corresponding mammograms produced by clinicians. We refer to this model as a \textit{patch-level} model, in contrast to the \textit{breast-level} model described in the section above which operates on images of the whole breast.
    
    Subsequently, we apply this auxiliary network to the full resolution mammograms in a sliding window fashion to create two \textit{heatmaps} for each image (an example in \autoref{fig:heatmaps}), one containing an estimated probability of a malignant finding for each pixel, and the other containing an estimated probability of a benign finding. These patch classification heatmaps can be used as additional input channels to the breast-level model to provide supplementary fine-grained information.
    
    Using separate breast- and pixel-level models as described above differentiates our work from approaches which utilize pixel-level labels in a single differentiable network \cite{multi_scale} or models based on the variations of R-CNN \cite{breast_cancer_rcnn}. Our approach allows us to use a very deep auxiliary network at the patch level, as this network does not have to process the entire high-resolution image at once. Adding the heatmaps produced by the patch-level classifier as additional input channels allows the main classifier to get the benefit from pixel-level labels, while the heavy computation necessary to produce the pixel-level predictions does not need to be repeated each time an example is used for learning. We can also initialize weights of the patch-level classifier using weights of networks pretrained on large off-domain data sets such as ImageNet \cite{imagenet}.\footnote{To finetune a network pretrained on RGB images with grayscale images, we duplicate the grayscale images across the RGB channels.}
    Hereafter, we refer to the model using only breast-level labels as the \textit{image-only} model, and the model using breast-level labels and the heatmaps as the \textit{image-and-heatmaps} model.

\section*{Experiments}

In all experiments, we used the training set for optimizing parameters of our model and the validation set for tuning hyperparameters of the model and the training procedure. Unless otherwise specified, results were computed across the screening population. To obtain predictions for each test example, we apply random transformations to the input 10 times, apply the model to each of the 10 samples separately and then average the 10 predictions (details in the SI).

To further improve our results, we employed the technique of model ensembling \cite{ensemble}, wherein the predictions of several different models are averaged to produce the overall prediction of the ensemble. In our case, we trained five copies of each model with different random initializations of the weights in the fully connected layers. The remaining weights are initialized with the weights of the model pretrained on BI-RADS classification, giving our model a significant boost in performance (details in the SI). For each model, we report the results from a single network (mean and standard deviation across five random initializations) and from an ensemble. 

\subsection*{Test populations} In the experiments below, we evaluate our model on several populations to test different hypotheses:
(i) \textit{screening population}, including all exams from the test set without subsampling; 
(ii) \textit{biopsied subpopulation}, which is subset of the screening population, only including exams from the screening population containing breasts which underwent a biopsy;
(iii) \textit{reader study subpopulation}, which consists of the biopsied subpopulation and a subset of randomly sampled exams from the screening population without any findings. 

\subsection*{Evaluation metrics} We evaluated our models primarily in terms of AUC (area under the ROC curve) for malignant/not malignant and benign/not benign classification tasks on the breast level. The model and readers' responses on the subset for reader study are evaluated in terms of AUC as well as precision-recall AUC (PRAUC), which are commonly used metrics in evaluation of radiologists' performance.

ROC and PRAUC capture different aspects of performance of a predictive model. The ROC curve summarizes the trade-off between the true positive rate and false positive rate for a model using different probability thresholds. The precision-recall curve summarizes the trade-off between the true positive rate (recall) and the positive predictive value (precision) for a model using different probability thresholds.

\subsection*{Screening population}
    
In this section we present the results on the screening population, which approximates the distribution of patients who undergo routine screening. Results are shown in the first two rows of \autoref{tab:cancer_pred}. The model ensemble using only mammogram images achieved an AUC of 0.840
for malignant/not malignant classification and an AUC of 0.743
for benign/not benign classification. The image-and-heatmaps model ensemble using both the images and the heatmaps achieved an AUC of 0.895
for malignant/not malignant and 0.756 for benign/not benign classification, outperforming the image-only model on both tasks.
The discrepancy in performance of our models between these two tasks can be largely explained by the fact that a larger fraction of benign findings than malignant findings are mammographically-occult (Table 2 in \cite{NYU_dataset}). Additionally, there 
can be noise in the benign/not benign labels associated with radiologists' confidence in their diagnoses. For the same exam, one radiologist might discard a finding as obviously not malignant without requesting a biopsy, while another radiologist might ask for a biopsy.
    
 We find that the image-and-heatmaps model performs better than the image-only model on both tasks. Moreover, the image-and-heatmaps model improves more strongly in malignant/not malignant classification than benign/not benign classification. We also find that ensembling is beneficial across all models, leading to a small but consistent increase in AUC.

    \begin{table}[t]
        \centering
        \caption{
            AUCs of our models on screening and biopsied populations.
        }
        \vspace{-2mm}
        \resizebox{.485\textwidth}{!}{
        \begin{tabular}{| l | c | c | c | c |}
        \cline{2-5}
        \multicolumn{1}{c|}{} & \multicolumn{2}{c|}{single} & \multicolumn{2}{c|}{5x ensemble} \\
        \cline{2-5}
        \multicolumn{1}{c|}{} & malignant & benign & malignant & benign \\ 

        \cline{2-5}
        \hline
        \multicolumn{5}{|c|}{\cellcolor{gray!20} {\textbf{screening population}} } \\ \hline
        image-only & 0.827$\pm$0.008 & 0.731$\pm$0.004 & 0.840 & 0.743 \\ \hline
        image-and-heatmaps & \textbf{0.886}$\pm$\textbf{0.003} & \textbf{0.747}$\pm$\textbf{0.002} & \textbf{0.895} & \textbf{0.756} \\ \hline
        \multicolumn{5}{|c|}{\cellcolor{gray!20} {\textbf{biopsied subpopulation}} } \\ \hline
        image-only & 0.781$\pm$0.006 & 0.673$\pm$0.003 & 0.791 & 0.682 \\ \hline
        image-and-heatmap & \textbf{0.843}$\pm$\textbf{0.004} & \textbf{0.690}$\pm$\textbf{0.002} & \textbf{0.850} & \textbf{0.696} \\ \hline
        \end{tabular}
        }
        \label{tab:cancer_pred}
        \vspace{-2mm}
    \end{table}

\subsection*{Biopsied subpopulation}
    We show the results of our models evaluated only on the biopsied subpopulation, in the last two rows of \autoref{tab:cancer_pred}. Within our test set, this corresponds to 401 breasts: 339 with benign findings, 45 with malignant findings, and 17 with both. This subpopulation that underwent biopsy with at least one imaging finding differs markedly from the overall screening population, which consists of largely healthy individuals undergoing routine annual screening without recall for additional imaging or biopsy. Compared to the results on the screening population, AUCs on the biopsied population are markedly lower across all the model variants.

    On the biopsied subpopulation, we observed a consistent difference between the performance of image-only and image-and-heatmaps models. The ensemble of image-and-heatmaps models performs best on both malignant/not malignant classification, attaining an AUC of 0.850,
    and on benign/not benign classification, attaining an AUC of 0.696.
    The markedly lower AUCs attained for the biopsied subpopulation, in comparison to the screening population, can be explained by the fact that exams that require a recall for diagnostic imaging and that subsequently need a biopsy are more challenging for both radiologists and our model.\footnote{More precisely, this difference in AUC can be explained by the fact that while adding or subtracting negative examples to the test population does not change the true positive rate, it alters the false positive rate. False positive rate is computed as a ratio of false positive and negative. Therefore, when adding easy negative examples to the test set, the number of false positives will be growing slower than the number of all negatives, which will lead to an increase in AUC. On the other hand, removing easy negative examples will have a reverse effect and the AUC will be lower.}

\subsection*{Results across ages and breast densities}
\begin{figure}[ht]
    \vspace{-3mm}
    \begin{tabular}{c c}
    \includegraphics[width = 0.22\textwidth]{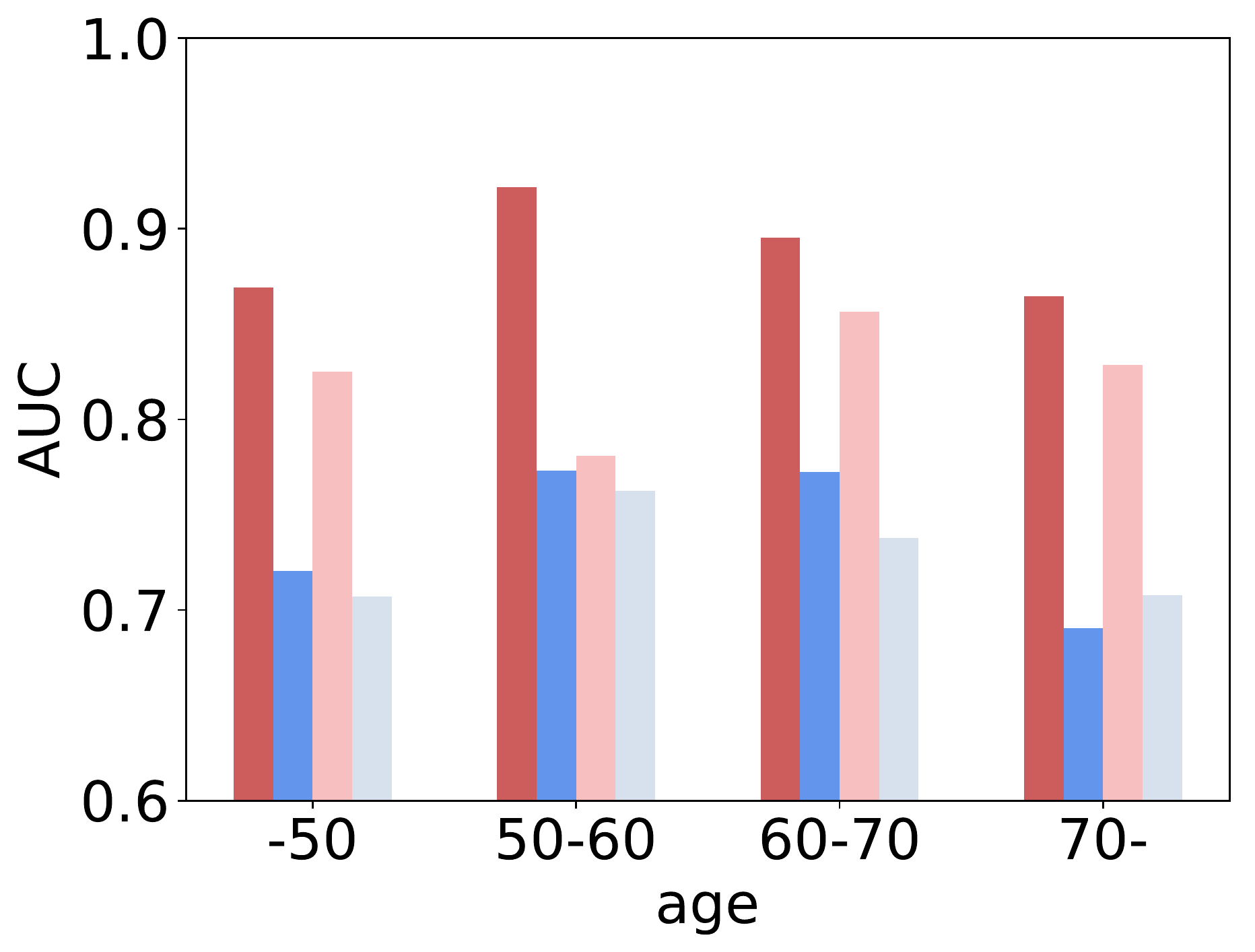} & \hspace{-5mm}
    \includegraphics[width = 0.255\textwidth]{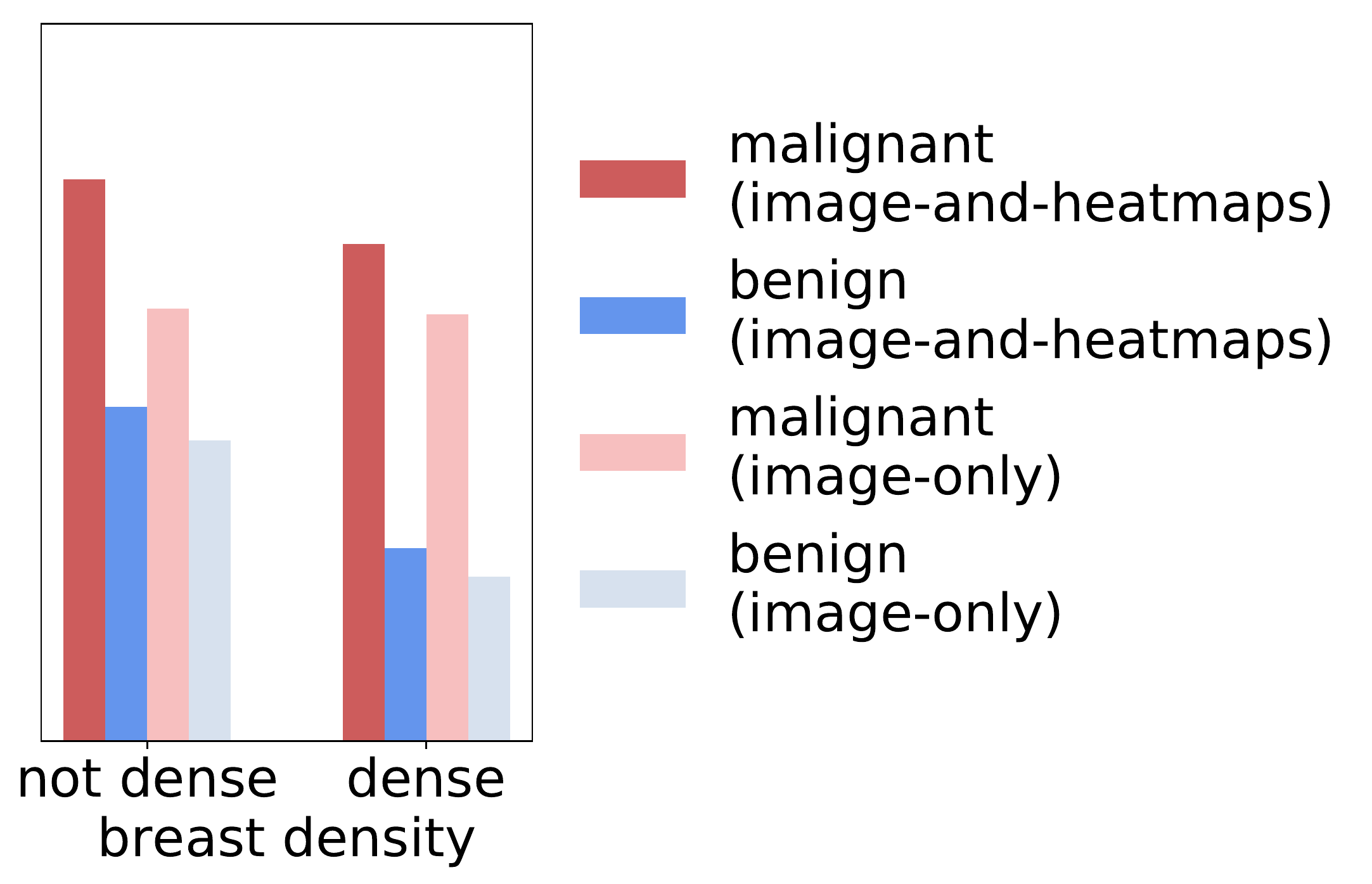}
    \end{tabular}
    \vspace{-3mm}
    \caption{AUCs for patients grouped by age and by breast density.}
    \vspace{-4mm}
    \label{fig:age_den}
    \end{figure}

    We divide the test set by patient age and breast density and evaluate our model on each subpopulation, as shown in \autoref{fig:age_den}. We observe that the performance of both the image-only and the image-and-heatmaps models varies across age groups. We also find that both models perform worse on dense breasts (``heterogeneously dense'' and ``extremely dense'') than on fattier ones (``almost entirely fatty'' and ``scattered areas of fibroglandular density''), which is consistent with the decreased sensitivity of radiologists for patients with denser breasts. Differences in the model's performance in benign/not benign classification is larger than in malignant/not malignant classification. We hypothesize that this is due to age and breast density influencing the level of noise in benign/not benign labels, associated with radiologists' confidence in their diagnoses.

\begin{figure*}[ht]
    \begin{tabular}{c c c c c c}
    \hspace{-3.5mm}\includegraphics[width = 0.165\textwidth]{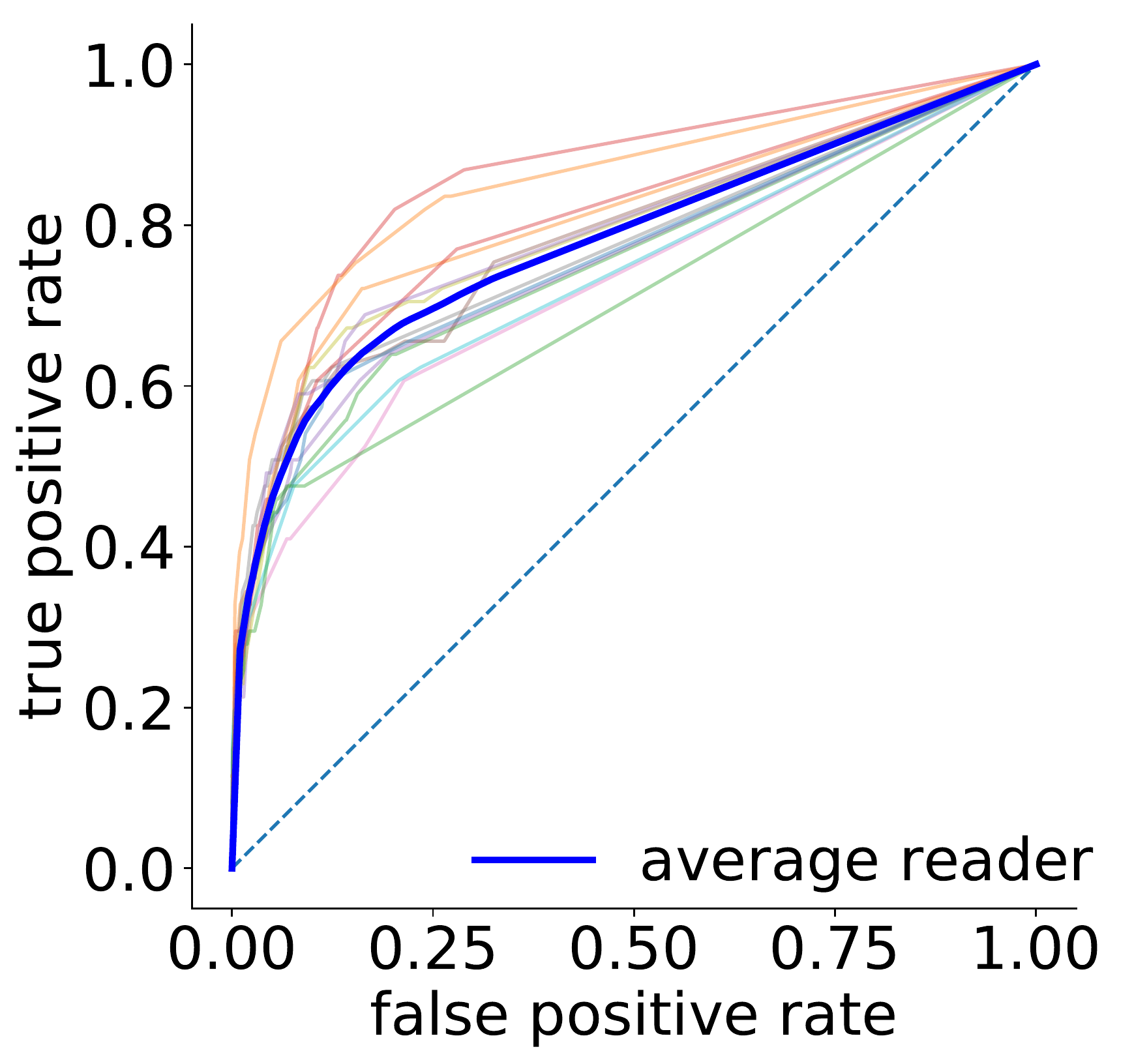}& 
    \hspace{-3.5mm}\includegraphics[width = 0.165\textwidth]{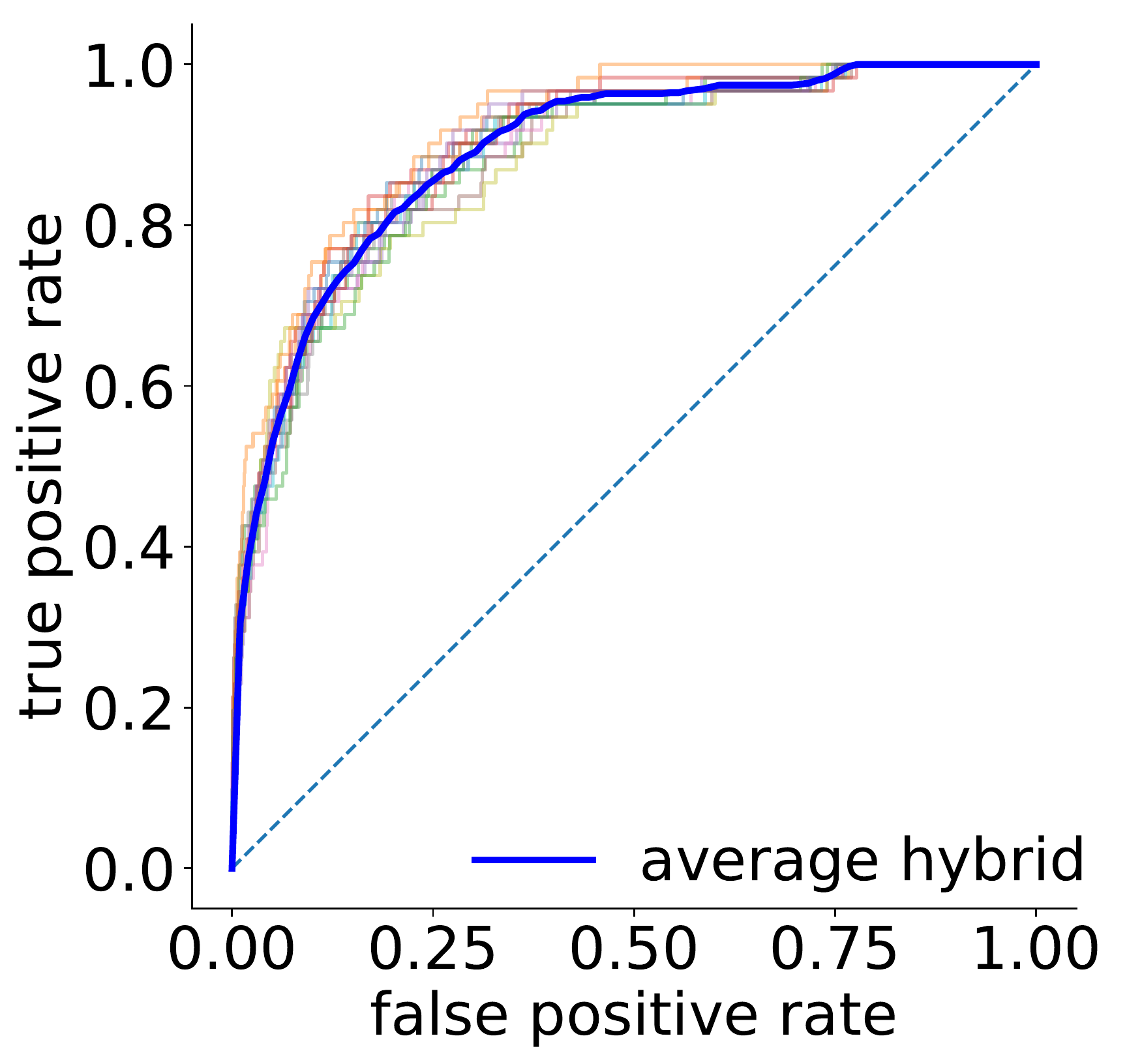} & 
    \hspace{-3.5mm}\includegraphics[width = 0.165\textwidth]{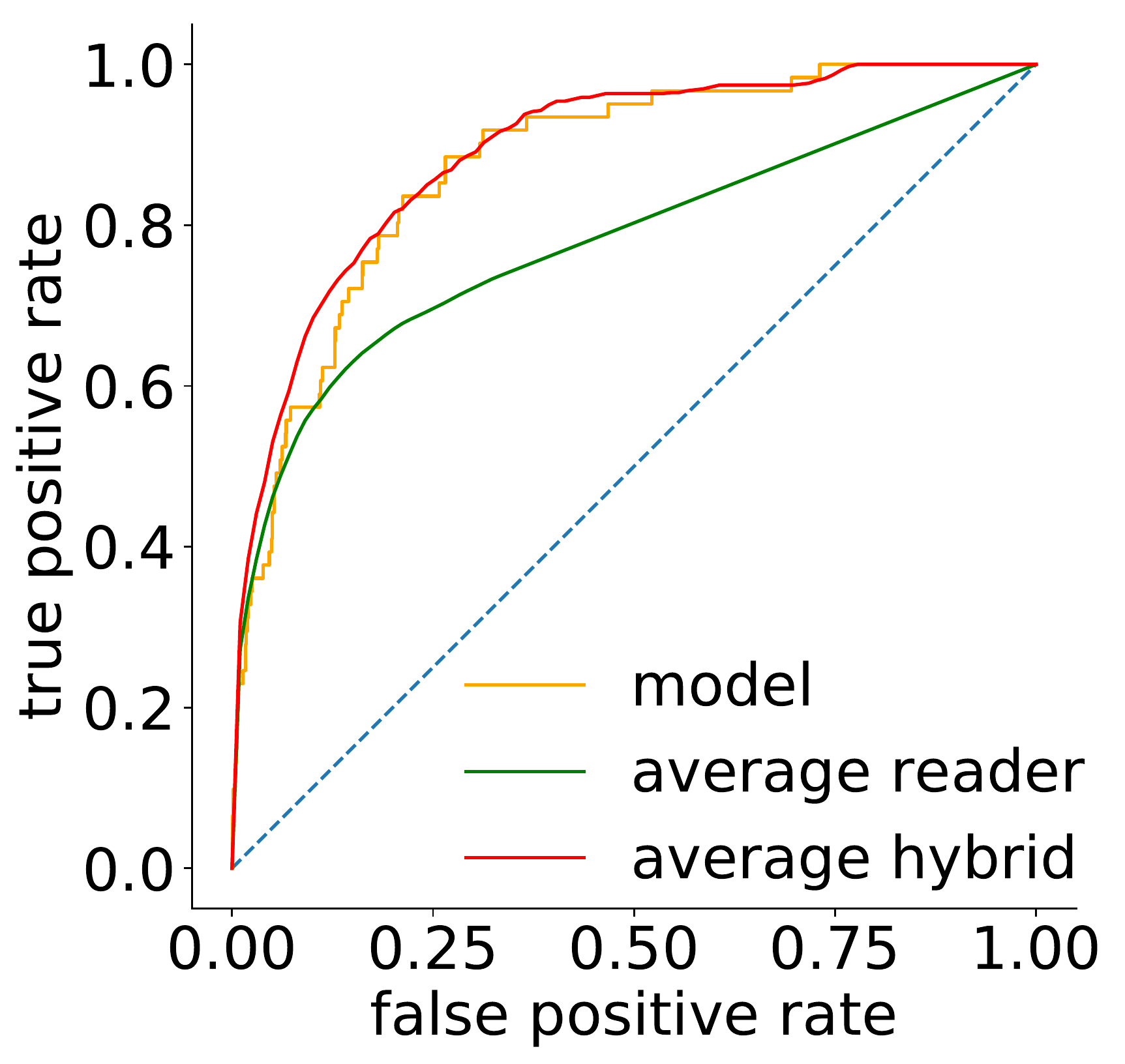} & 
     \hspace{-3.5mm}\includegraphics[width = 0.165\textwidth]{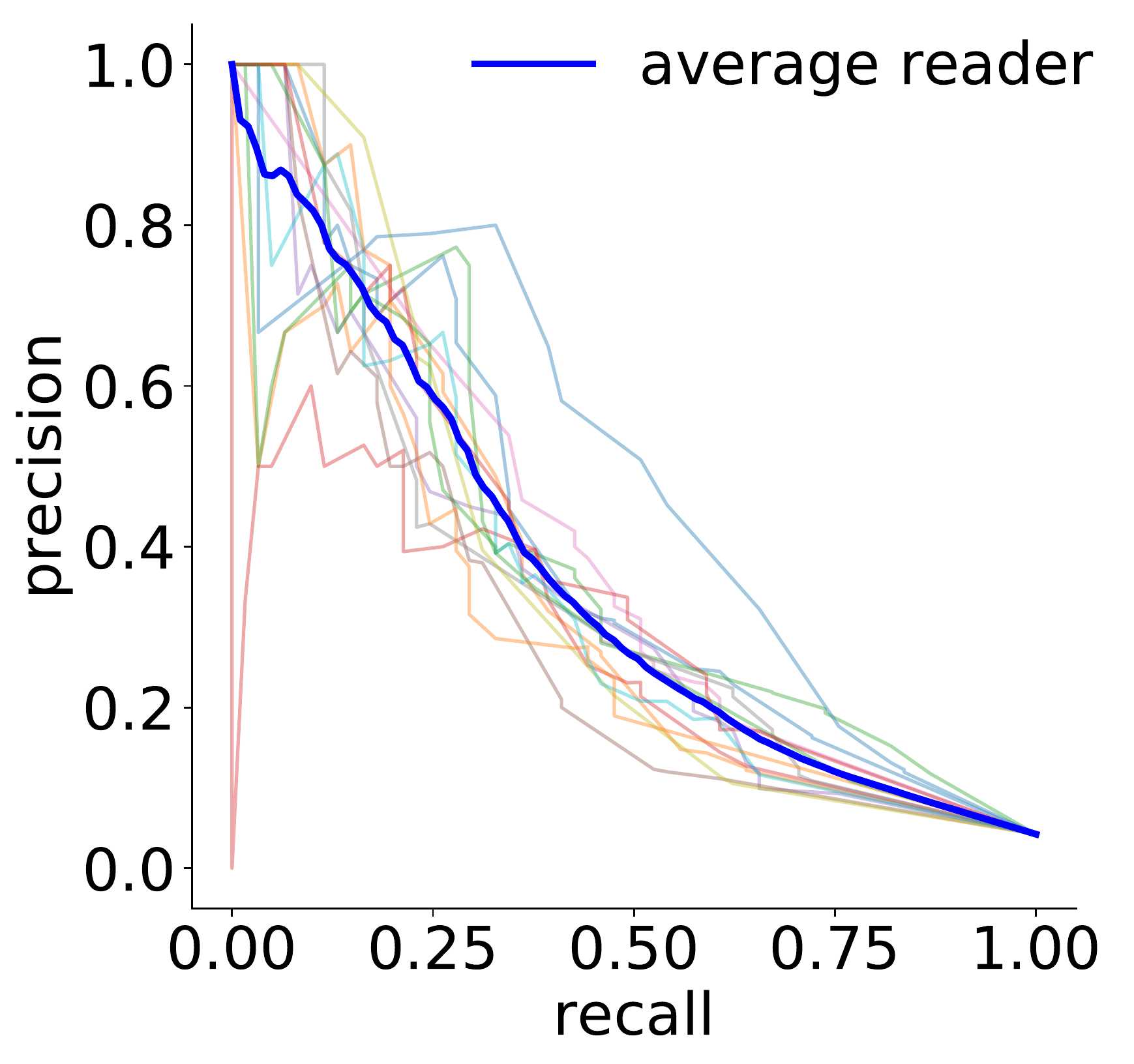}& \hspace{-3.5mm}\includegraphics[width = 0.165\textwidth]{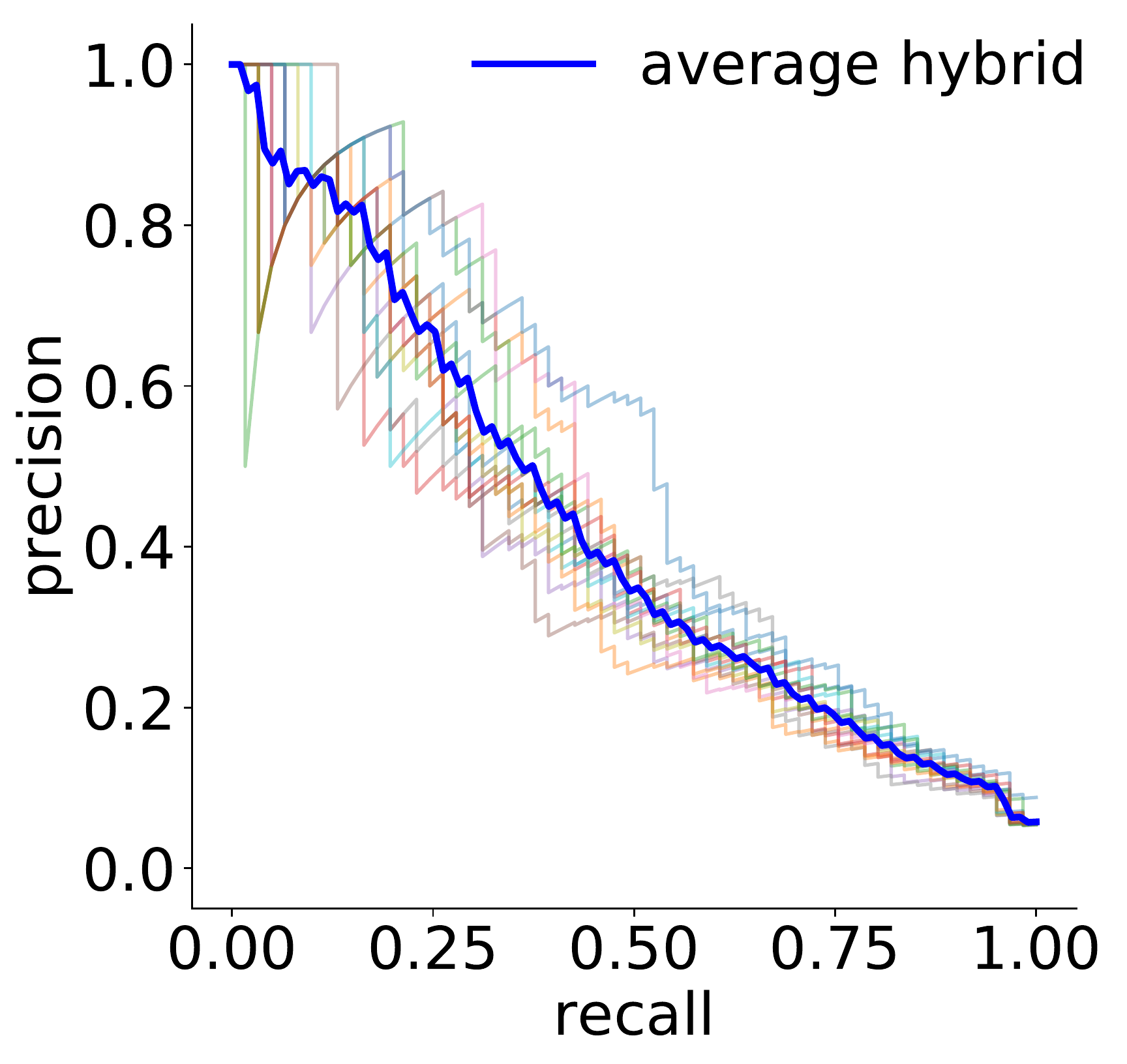} & 
    \hspace{-3.5mm}\includegraphics[width = 0.165\textwidth]{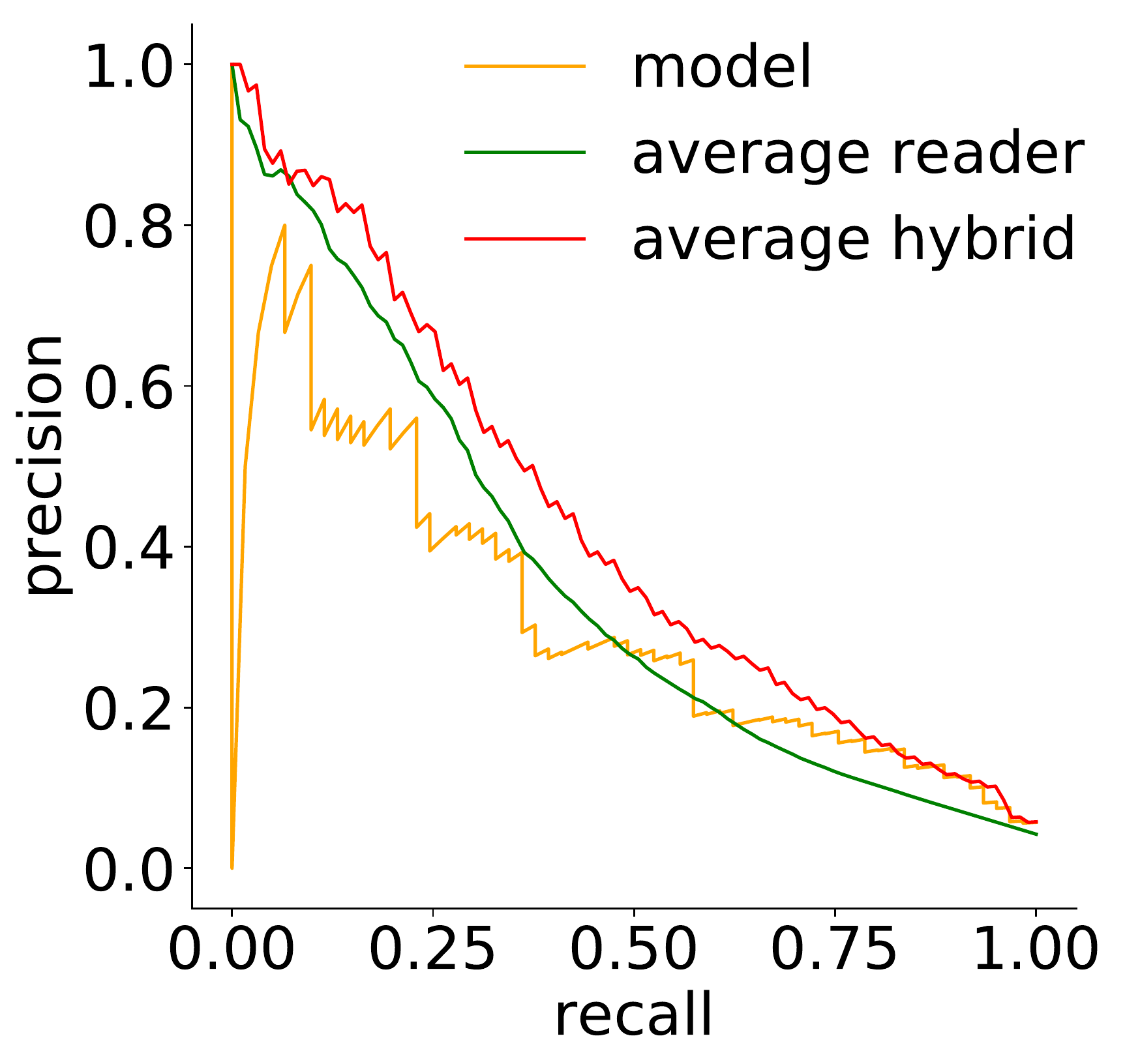}\\
    \vspace{-5mm}\\
    \footnotesize{(a)} & \footnotesize{(b)} & \footnotesize{(c)} &
    \footnotesize{(a*)} & \footnotesize{(b*)} & \footnotesize{(c*)} \\
    \end{tabular}
    \vspace{-2mm}
    \caption{ROC curves ((a), (b), (c)) and Precision-Recall curves ((a*), (b*), (c*)) on the subset of the test set used for reader study. (a) \& (a*): curves for all 14 readers. Their average performance are highlighted in blue. (b) \& (b*): curves for hybrid of the image-and-heatmaps ensemble with each single reader. Curve highlighted in blue indicates the average performance of all hybrids. (c) \& (c*): comparison among the image-and-heatmaps ensemble, average reader and average hybrid.}
    \label{fig:human_ai_comparison}
    \vspace{-3mm}
\end{figure*}
\section*{Reader study}

To compare the performance of our image-and-heatmaps ensemble (hereafter referred to as \textit{the model}) to human radiologists, we performed a reader study with 14 readers---12 attending radiologists at various levels of experience (between 2 and 25 years), a resident and a medical student---each reading 740 exams from the test set (1,480 breasts): 368 exams randomly selected from the biopsied subpopulation 
and 372 exams randomly selected from exams not matched with any biopsy. Exams were shuffled before being given to the readers. Readers were asked to provide a probability estimate of malignancy on a 0\%-100\% scale for each breast in an exam. As some breasts contain multiple suspicious findings, readers were asked to give their assessment of the most suspicious finding.

We used the first 20 exams as a practice set to familiarize readers with the format of the reader study--these were excluded from the analysis.\footnote{The readers were shown the images and asked to give their assessment. We confirmed the correctness of the format in which they returned their answers but we did not provide them with feedback on the accuracy of their predictions.} On the remaining 720 exams, we evaluated the model's and readers' performance on malignancy classification. Among the 1,440 breasts, there are 62 breasts labeled as malignant and 356 breasts labeled as benign. In the breasts labeled as malignant, there are 21 masses, 26 calcifications, 12 asymmetries and 4 architectural distortions.\footnote{Masses are defined as 3-dimensional space occupying lesion with completely or partially convex-outward borders. Calcifications are tiny specks of calcific deposits. An asymmetry is defined as a unilateral deposit of fibroglandular tissue that does not meet the definition of mass, i.e., it is an area of the fibroglandular tissue that is not seen other breast. Architectural distortion refers to a disruption of the normal random pattern of fibroglandular tissue with no definite mass visible.}\footnote{As one breast had two types of findings, the numbers add up to 39, not 38.} In the breasts labeled as benign, the corresponding numbers of imaging findings are: 87, 102, 36 and 6. 

Our model achieved an AUC of 0.876 
and PRAUC of 0.318. AUCs achieved by individual readers varied from 0.705 to 0.860 (mean: 0.778, std: 0.0435). PRAUCs for readers varied from 0.244 to 0.453 (mean: 0.364, std: 0.0496). Individual ROCs and precision-recall curves, along with their averages are shown in \autoref{fig:human_ai_comparison}(a) and  \autoref{fig:human_ai_comparison}(a*). 

\begin{figure*}[ht]
\centering
    \begin{minipage}[t]{.48\textwidth}
    \centering
    \begin{tabular}{c c}
    \hspace{-1mm}\includegraphics[height=0.416\linewidth]{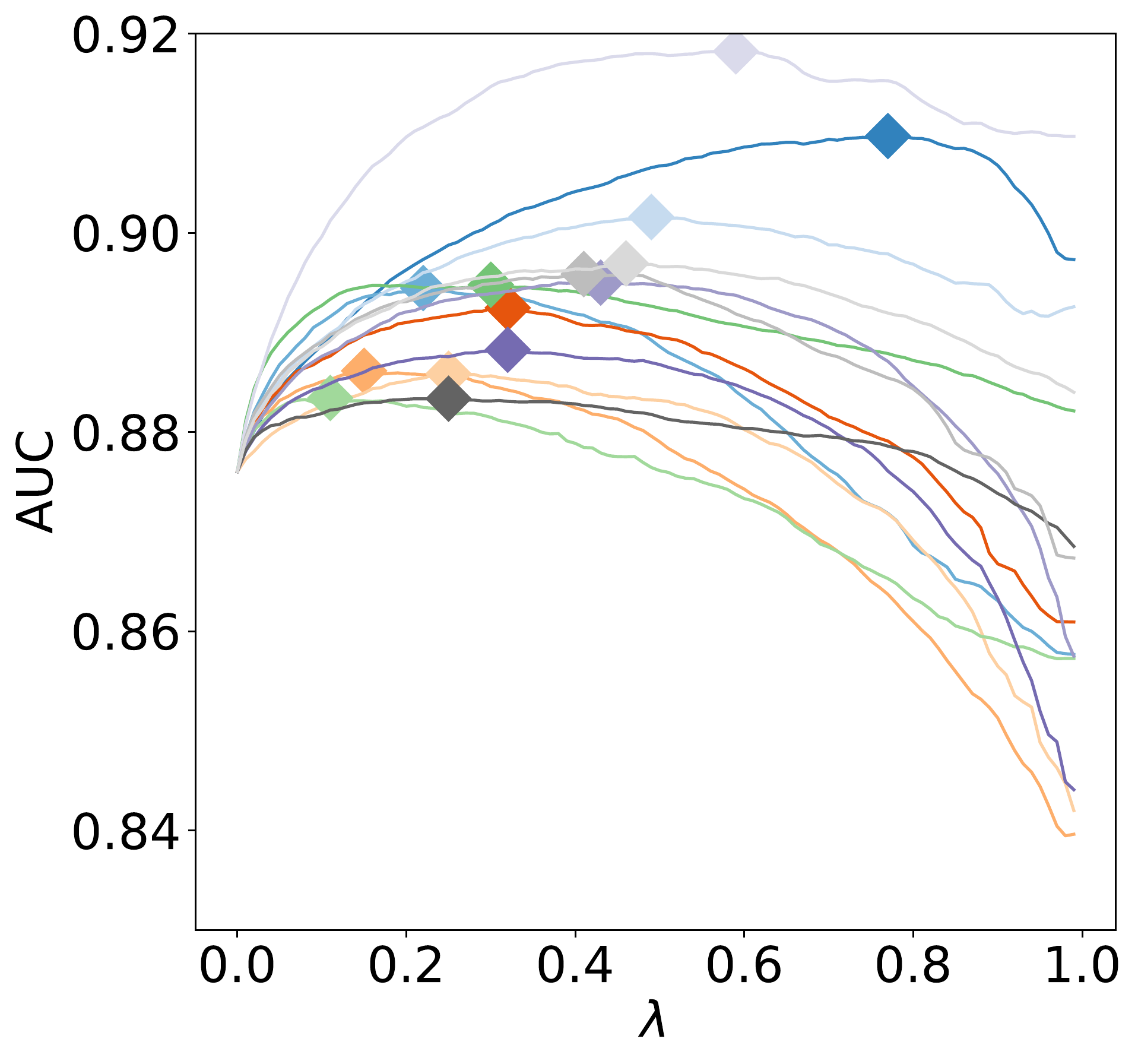} &
    \hspace{-4mm}\includegraphics[height=0.416\linewidth]{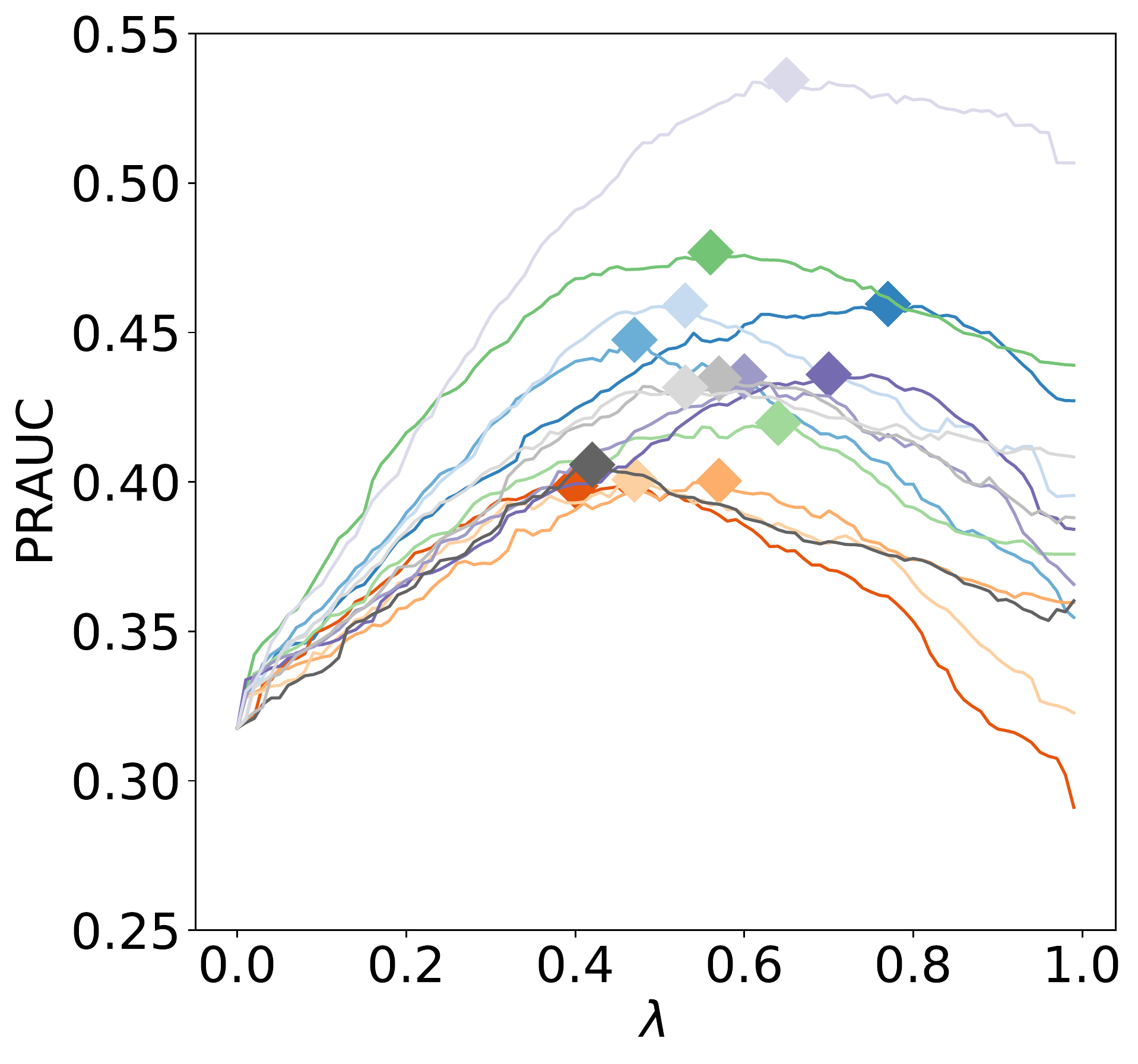}
    \end{tabular}
    \vspace{-4mm}
    \caption{AUC (left) and PRAUC (right) as a function of $\lambda \in [0, 1)$ for hybrids between each reader and our image-and-heatmaps ensemble. Each hybrid achieves the highest AUC/PRAUC for a different $\lambda$ (marked with $\diamondsuit$). 
    }
    \label{fig:weighting}
    \end{minipage}
    \hspace{3mm}
    \begin{minipage}[t]{.48\textwidth}
    \centering
    \begin{tabular}{c c c c}
    \phantom{ab}&\vspace{0.5mm}\hspace{-1mm}\includegraphics[height=0.38\linewidth]{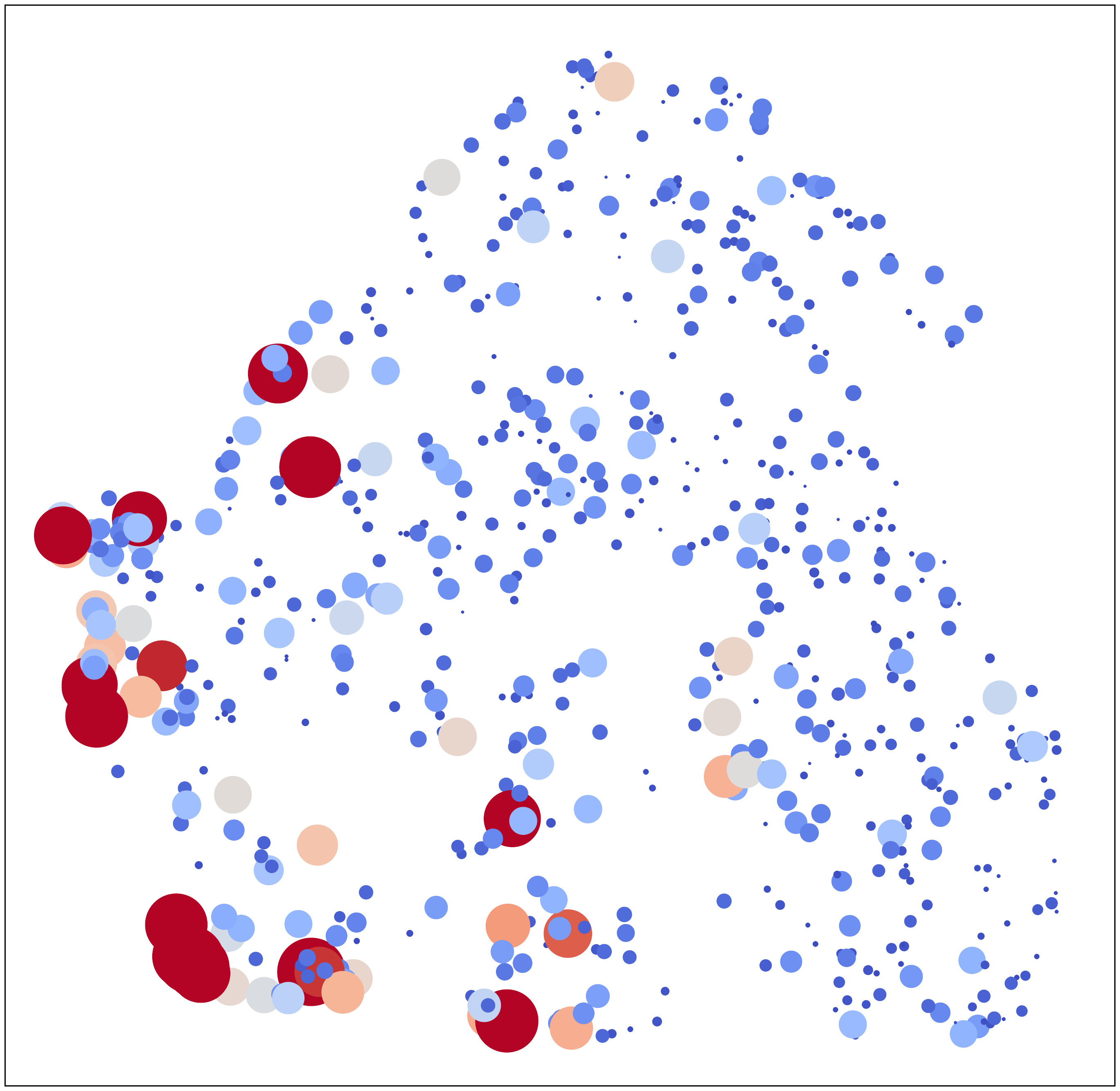} & \vspace{0.5mm}\hspace{-3mm} \includegraphics[height=0.38\linewidth]{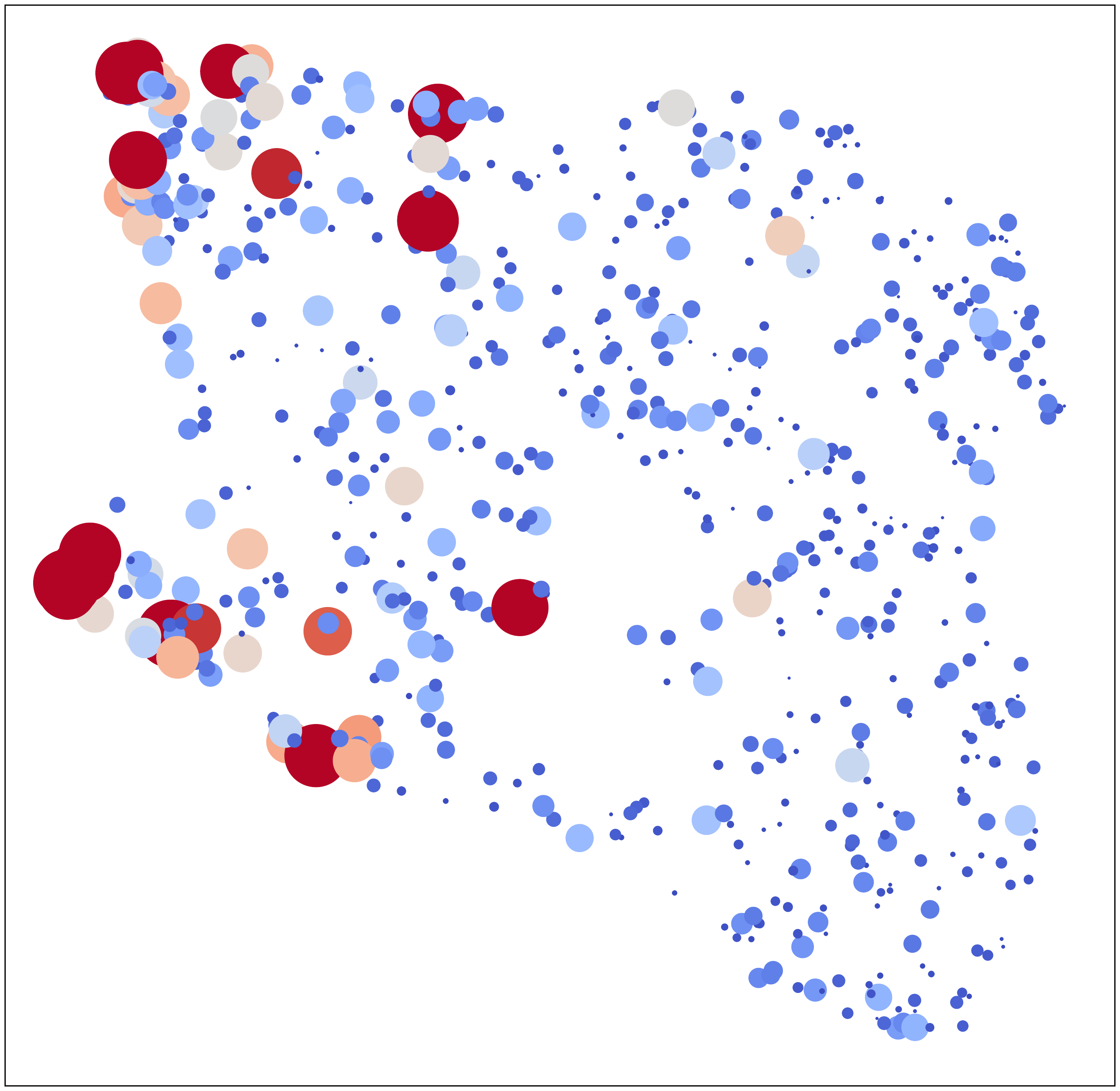} &\hspace{-2mm}
    \raisebox{.11\height}{\includegraphics[height=0.3\linewidth, trim=3mm 0mm 10mm 0mm]{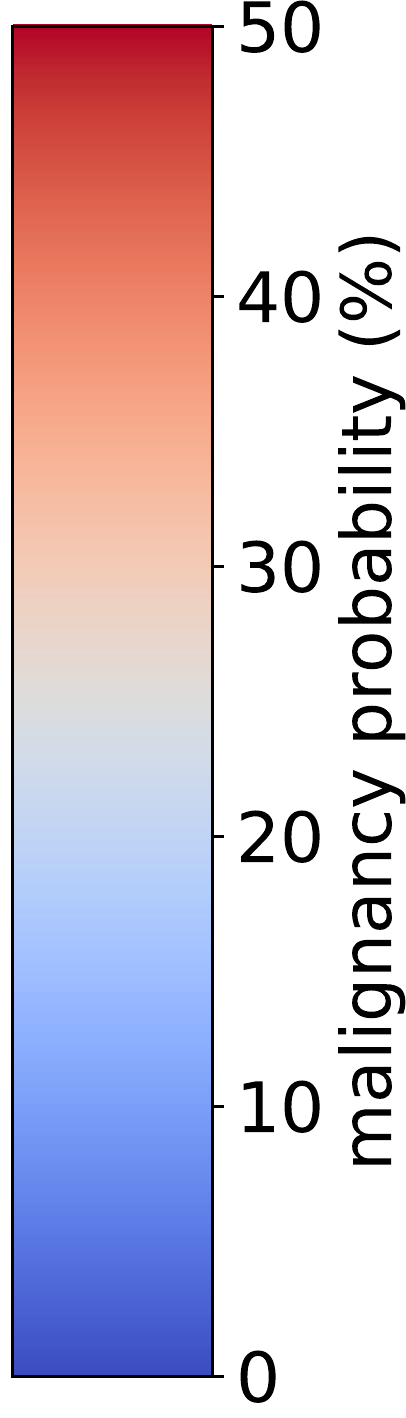}}
    \end{tabular}
    \vspace{-3mm}
    \caption{Exams in the reader study set represented using the concatenated activations from the four image-specific columns (left) and the concatenated activations from the first fully connected layer in both CC and MLO model branches (right).}
    \label{fig:tsne_breast}
    \end{minipage}
\end{figure*}

We also evaluated the accuracy of a human-machine hybrid, whose predictions are a linear combination of predictions of a radiologist and of the model--that is, $\mathbf{y}_\mathrm{hybrid} = \lambda\mathbf{y}_\mathrm{radiologist} + (1 - \lambda)\mathbf{y}_\mathrm{model}$. 
For $\lambda = 0.5$\footnote{We do not have a way to tune $\lambda$ to individual readers, hence we chose $\lambda = 0.5$ as the most natural way of aggregating two sets of predictions when not having prior knowledge of their quality. As \autoref{fig:weighting} shows, an optimal $\lambda$ varies a lot depending on the reader. The stronger reader's performance the smaller the optimal weight on the model. Notably though all readers can be improved by averaging their predictions with the model for both AUC and PRAUC.} (see \autoref{fig:weighting} for the results for $\lambda \in [0, 1)$), hybrids between each reader and the model achieved an average AUC of 0.891 (std: 0.0109) 
and an average PRAUC of 0.431 (std: 0.0332)
(cf. \autoref{fig:human_ai_comparison}(b), \autoref{fig:human_ai_comparison}(b*)). These results suggest our model can be used as a tool to assist radiologists in reading breast cancer screening exams and that it captured different aspects of the task compared to experienced breast radiologists.

\subsection*{Visualization of the representation learned by the classifier}
Additionally, we examined how the network represents the exams internally by visualizing the hidden representations learned by the best single image-and-heatmaps model, for exams in reader study subpopulation. We visualize two sets of activations: concatenated activations from the last layer of each of the four image-specific columns, and concatenated activations from the first fully connected layer in both CC and MLO model branches. Both sets of activations have 1,024 dimensions in total. We embed them into a two-dimensional space using UMAP~\citep{umap} with the Euclidean distance. 

\autoref{fig:tsne_breast} shows the embedded points. Color and size of each point reflect the same information: the warmer and larger the point is, the higher the readers' mean prediction of malignancy is. A score for each exam is computed as an average over predictions for the two breasts. 
We observe that exams classified as more likely to be malignant according to the readers are close to each other for both sets of activations. The fact that previously unseen exams with malignancies were found by the network to be similar further corroborates that our model exhibits strong generalization capabilities.

\section*{Related work}

Prior works approach the task of breast cancer screening exam classification in two paradigms. In one paradigm, only exam-level, breast-level or image-level labels are available. A CNN is first applied to each of the four standard views and the resulting feature vectors are combined to produce a final prediction \cite{high_resolution}. This workflow can be further integrated with multi-task learning where radiological assessments, such as breast density, can be incorporated to model the confidence of the classification \cite{mammo}. Other works formulate the breast cancer exam classification task as weakly supervised localization and produce a class activation map that highlights the locations of suspicious lesions \cite{data_driven}. Such formulations can be paired with multiple-instance learning where each spatial location is treated as a single instance and associated with a score that is correlated with the existence of a malignant finding \cite{multi_instance}.

In the second paradigm, pixel-level labels that indicate the location of benign or malignant findings are also provided to the classifier during training. The pixel-level labels enable training models derived from the R-CNN architecture \cite{breast_cancer_rcnn} or models that divide the mammograms into smaller patches and train patch-level classifiers using the location of malignant findings \cite{kooi2017classifying, shen2017end, teare2017malignancy, multi_scale}. Some of these works directly aggregate outputs from the patch-level classifier to form an image-level prediction. A major limitation of such architectures is that information outside the annotated regions of interest will be neglected. Other works apply the patch-level classifier as a first level of feature extraction on top of which more layers are stacked and the entire model is then optimized jointly. A downside of this kind of architecture is the requirement for the whole model to fit in GPU memory for training, which limits the size of the minibatch used (usually to one), depth of the patch-level model and how densely the patch-level model is applied. Our work is most similar to the latter type of models utilizing pixel-level labels--however our strategy uses a patch-level classifier for producing heatmaps as additional input channels to the breast-level classifier. While we forgo the ability to train the whole model end-to-end, the patch-level classifier can be significantly more powerful and can be densely applied across the original image. As a result, our model has the ability to learn both local features across the entire image as well as macroscopic features such as symmetry between breasts. For a more comprehensive review of prior work, refer to one of the recent reviews \cite{AJR_review, harvey_review}.

A variety of results in terms of AUC for prediction of malignancy have been reported in literature. The most comparable to our work are: \cite{multi_instance} (0.86), \cite{breast_cancer_rcnn} (0.95), \cite{becker2017deep} (0.81), \cite{data_driven} (0.91), \cite{standalone} (0.84) and \cite{ritse} (0.89). Unfortunately, although these results can serve as a rough estimate of model quality, comparing different methods based on these numbers would be misleading. Some authors do not discuss design of their models \cite{becker2017deep, ritse, standalone}, some evaluate their models on very small public datasets \cite{inbreast, DDSM}, insufficient for a meaningful evaluation, while others used private datasets with populations of different distributions (on a spectrum between screening population and biopsied subpopulation), different quality of imaging equipment and even differently defined labels. By making the code and the weights of our model public, we seek to enable more direct comparisons to our work.

\section*{Discussion}

By leveraging a large training set with breast-level and pixel-level labels, we built a neural network which can accurately classify breast cancer screening exams. We attribute this success in large part to the significant amount of computation encapsulated in the patch-level model, which was densely applied to the input images to form heatmaps as additional input channels to a breast-level model. It would be impossible to train this model in a completely end-to-end fashion with currently available hardware. Although our results are promising, we acknowledge that the test set used in our experiments is relatively small and our results require further clinical validation.
We also acknowledge that although our network's performance is stronger than that of the radiologists' on the specific task in our reader study, this is not exactly the task that radiologists perform. Typically, screening mammography is only the first step in a diagnostic pipeline, with the radiologist making a final determination and decision to biopsy only after recall for additional diagnostic mammogram images and possible ultrasound. However, in our study a hybrid model including both a neural network and expert radiologists outperformed either individually, suggesting the use of such a model could improve radiologist sensitivity for breast cancer detection. 

On the other hand, the design of our model is relatively simple. More sophisticated and accurate models are possible. Furthermore, the task we considered in this work, predicting whether the patient had a visible cancer at the time of the screening mammography exam, is the simplest possible among many tasks of interest. In addition to testing the utility of this model in real-time reading of screening mammograms, a clear next step would be predicting the development of breast cancer in the future--before it is even visible to a trained human eye.

\acknow{The authors would like to thank Catriona C. Geras for correcting earlier versions of this manuscript, Michael Cantor for providing us pathology reports, Marc Parente and Eli Bogom-Shanon for help with importing the image data and Mario Videna for supporting our computing environment. We also gratefully acknowledge the support of Nvidia Corporation with the donation of some of the GPUs used in this research. This work was supported in part by grants from the National Institutes of Health (R21CA225175 and P41EB017183).}

\showacknow{}

\bibliography{main}

\begin{thebibliography}{1}

\bibitem{NYU_dataset}
Wu N, et~al. (2019) The {NYU} breast cancer screening dataset v1.0, Technical
  report.
\newblock Available at \url{https://cs.nyu.edu/~kgeras/reports/datav1.0.pdf}.

\bibitem{resnet}
He K, Zhang X, Ren S, Sun J (2016) Deep residual learning for image recognition
  in {\em CVPR}.

\bibitem{adam}
Kingma DP, Ba J (2015) Adam: A method for stochastic optimization in {\em
  ICLR}.

\bibitem{imbalanced_survey}
Branco P, Torgo L, Ribeiro RP (2016) A survey of predictive modeling on
  imbalanced domains.
\newblock {\em ACM Computing Surveys} 49(2).

\bibitem{high_resolution}
Geras KJ, et~al. (2017) High-resolution breast cancer screening with multi-view
  deep convolutional neural networks.
\newblock {\em arXiv:1703.07047}.

\bibitem{6909618}
Oquab M, Bottou L, Laptev I, Sivic J (2014) Learning and transferring mid-level
  image representations using convolutional neural networks in {\em CVPR}.

\bibitem{DBLP:journals/corr/KrauseSHZTDPL15}
Krause J, et~al. (2015) The unreasonable effectiveness of noisy data for
  fine-grained recognition.
\newblock {\em arXiv:1511.06789}.

\bibitem{DBLP:journals/corr/SunSSG17}
Sun C, Shrivastava A, Singh S, Gupta A (2017) Revisiting unreasonable
  effectiveness of data in deep learning era.
\newblock {\em arXiv:1707.02968}.

\bibitem{densenet}
Huang G, Liu Z, van~der Maaten L, Weinberger KQ (2017) Densely connected
  convolutional networks in {\em CVPR}.

\end{thebibliography}


\begin{thebibliography}{10}

\bibitem{RN53}
Siegel RL, Miller KD, Jemal A (2015) Cancer statistics, 2015.
\newblock {\em CA: a cancer journal for clinicians} 65(1).

\bibitem{RN38}
Duffy SW, et~al. (2002) The impact of organized mammography service screening
  on breast carcinoma mortality in seven swedish counties.
\newblock {\em Cancer} 95(3).

\bibitem{RN40}
Kopans DB (2002) Beyond randomized controlled trials: organized mammographic
  screening substantially reduces breast carcinoma mortality.
\newblock {\em Cancer} 94(2).

\bibitem{RN41}
Duffy SW, Tabar L, Smith RA (2002) The mammographic screening trials:
  commentary on the recent work by {O}lsen and {G}otzsche.
\newblock {\em CA Cancer J Clin} 52(2).

\bibitem{RN43}
Kopans DB (2015) An open letter to panels that are deciding guidelines for
  breast cancer screening.
\newblock {\em Breast Cancer Res Treat} 151(1).

\bibitem{with_and_without_CAD}
Lehman CD, et~al. (2015) Diagnostic accuracy of digital screening mammography
  with and without computer-aided detection.
\newblock {\em JAMA Internal Medicine} 175(11).

\bibitem{deep_learning}
LeCun Y, Bengio Y, Hinton G (2015) Deep learning.
\newblock {\em Nature} 521(7553).

\bibitem{convnet}
LeCun Y, et~al. (1989) Handwritten digit recognition with a back-propagation
  network in {\em NIPS}.

\bibitem{alexnet}
Krizhevsky A, Sutskever I, Hinton GE (2012) Imagenet classification with deep
  convolutional neural networks in {\em NIPS}.

\bibitem{vggnet}
Simonyan K, Zisserman A (2014) Very deep convolutional networks for large-scale
  image recognition in {\em ICLR}.

\bibitem{resnet}
He K, Zhang X, Ren S, Sun J (2016) Deep residual learning for image recognition
  in {\em CVPR}.

\bibitem{densenet}
Huang G, Liu Z, van~der Maaten L, Weinberger KQ (2017) Densely connected
  convolutional networks in {\em CVPR}.

\bibitem{NYU_dataset}
Wu N, et~al. (2019) The {NYU} breast cancer screening dataset v1.0, Technical
  report.
\newblock Available at \url{https://cs.nyu.edu/~kgeras/reports/datav1.0.pdf}.

\bibitem{high_resolution}
Geras KJ, et~al. (2017) High-resolution breast cancer screening with multi-view
  deep convolutional neural networks.
\newblock {\em arXiv:1703.07047}.

\bibitem{breast_density}
Wu N, et~al. (2018) Breast density classification with deep convolutional
  neural networks in {\em ICASSP}.

\bibitem{multi_scale}
Lotter W, Sorensen G, Cox D (2017) A multi-scale {CNN} and curriculum learning
  strategy for mammogram classification in {\em DLMIA}.

\bibitem{breast_cancer_rcnn}
Ribli D, Horv\'{a}th A, Unger Z, Pollner P, Csabai I (2018) Detecting and
  classifying lesions in mammograms with deep learning.
\newblock {\em Scientific Reports} 8(1).

\bibitem{imagenet}
Deng J, et~al. (2009) {ImageNet: A Large-Scale Hierarchical Image Database} in
  {\em CVPR}.

\bibitem{ensemble}
Dietterich TG (2000) Ensemble methods in machine learning in {\em Multiple
  classifier systems}.

\bibitem{umap}
McInnes L, Healy J, Saul N, Gro{\ss}berger L (2018) {UMAP:} uniform manifold
  approximation and projection.
\newblock {\em J. Open Source Software} 3(29).

\bibitem{mammo}
Kyono T, Gilbert FJ, van~der Schaar M (2018) {MAMMO:} {A} deep learning
  solution for facilitating radiologist-machine collaboration in breast cancer
  diagnosis.
\newblock {\em arXiv:1811.02661}.

\bibitem{data_driven}
Kim EK, et~al. (2018) Applying data-driven imaging biomarker in mammography for
  breast cancer screening: Preliminary study.
\newblock {\em Scientific Reports} 8.

\bibitem{multi_instance}
Zhu W, Lou Q, Vang YS, Xie X (2017) Deep multi-instance networks with sparse
  label assignment for whole mammogram classification in {\em MICCAI}.

\bibitem{kooi2017classifying}
Kooi T, Karssemeijer N (2017) Classifying symmetrical differences and temporal
  change for the detection of malignant masses in mammography using deep neural
  networks.
\newblock {\em Journal of Medical Imaging} 4(4).

\bibitem{shen2017end}
Shen L (2017) End-to-end training for whole image breast cancer diagnosis using
  an all convolutional design.
\newblock {\em arXiv preprint arXiv:1708.09427}.

\bibitem{teare2017malignancy}
Teare P, Fishman M, Benzaquen O, Toledano E, Elnekave E (2017) Malignancy
  detection on mammography using dual deep convolutional neural networks and
  genetically discovered false color input enhancement.
\newblock {\em Journal of digital imaging} 30(4).

\bibitem{AJR_review}
Gao Y, Geras KJ, Lewin AA, Moy L (2018) New frontiers: An update on cad for
  breast imaging in the age of artificial intelligence.
\newblock {\em American Journal of Roentegonoly}.

\bibitem{harvey_review}
Harvey H, et~al. (2019) The role of deep learning in breast screening.
\newblock {\em Current Breast Cancer Reports}.

\bibitem{becker2017deep}
Becker AS, et~al. (2017) Deep learning in mammography: Diagnostic accuracy of a
  multipurpose image analysis software in the detection of breast cancer.
\newblock {\em Investigative Radiology}.

\bibitem{standalone}
Rodriguez-Ruiz A, et~al. (2019) Stand-alone artificial intelligence for breast
  cancer detection in mammography: Comparison with 101 radiologists.
\newblock {\em Journal of the National Cancer Institute}.

\bibitem{ritse}
Rodr{\'\i}guez-Ruiz A, et~al. (2018) Detection of breast cancer with
  mammography: Effect of an artificial intelligence support system.
\newblock {\em Radiology}.

\bibitem{inbreast}
Moreira IC, et~al. (2012) Inbreast: toward a full-field digital mammographic
  database.
\newblock {\em Academic radiology} 19(2).

\bibitem{DDSM}
Heath M, et~al. (1998) Current status of the digital database for screening
  mammography in {\em Digital Mammography}.

\end{thebibliography}

\end{document}


\newcolumntype{C}[1]{>{\centering\arraybackslash}m{#1}}
\maketitle

\section*{Additional analysis of results}

We detail in this section additional results and analyses to supplement the results in the main paper.

\subsection*{Heatmaps-only model} 

We evaluate the relative benefit of including images and heatmaps as inputs to our model by also considering a model variant where the model only takes as input the patch classification heatmaps, without being shown the original mammogram. The results are shown in \autoref{tab:cancer_pred_input_variant}. 
We see that the heatmaps-only model performs comparably with the image-only model on malignant/not malignant classification, while significantly underperforming on benign/not benign classification. We speculate that this discrepancy arises from the higher prevalence of mammographically occult benign findings. The patch classification models are trained on classifying patches based on pixel-level segmentations, which contain a higher density of label information compared to breast-level labels and thus provide a stronger learning signal, leading to better performance on malignant/not malignant prediction. On the other hand, because benign findings are more likely to be mammographically occult (see Table 2 in \cite{NYU_dataset}), these cases cannot be segmented and hence are not present in the patch dataset--the patch classification model is thus less well-conditioned to those benign findings. Conversely, the image-only model is still shown the benign labels derived from biopsies, and may thus pick up on visual clues suggesting a benign finding despite the cases being considered mammographically occult by radiologists. The superior performance of the heatmaps-only model over the image-only model on malignant/not-malignant classification also suggests that the increased depth of the model and the more strongly supervised nature of the patch classification task outweighs the benefits of training a deep model end-to-end. In addition, we observe a much smaller benefit to ensembling the heatmaps-only models compared to the image-only and image-and-heatmaps models. The intuition behind this observation is that the heatmaps have likely already distilled most of the pertinent information for cancer classification. We speculate that because the heatmaps-only model learn a simpler transformation to target cancer classification, there is lower model diversity and thus a smaller benefit from ensembling. 

Above all, the image-and-heatmaps model still remains the strongest overall model, demonstrating that effectively utilizing both local and global visual information leads to superior performance on the cancer classification problem.

    \begin{table}[ht]
        \centering
        \caption{
            AUCs of model input variants on screening and biopsied populations.
        }
        \begin{tabular}{| l | c | c | c | c |}
        \cline{2-5}
        \multicolumn{1}{c|}{} & \multicolumn{2}{c|}{single} & \multicolumn{2}{c|}{5x ensemble} \\
        \cline{2-5}
        \multicolumn{1}{c|}{} & malignant & benign & malignant & benign \\ 
        \cline{2-5}
        \hline
        \multicolumn{5}{|c|}{\cellcolor{gray!20} {\textbf{screening population}} } \\ \hline
        image-only & 0.827$\pm$0.008 & 0.731$\pm$0.004 & 0.840 & 0.743 \\ \hline
        heatmaps-only & 0.837$\pm$0.010 & 0.674$\pm$0.007 & 0.835 & 0.691 \\ \hline
        image-and-heatmaps & \textbf{0.886}$\pm$\textbf{0.003} & \textbf{0.747}$\pm$\textbf{0.002} & \textbf{0.895} & \textbf{0.756} \\ \hline
    
        \multicolumn{5}{|c|}{\cellcolor{gray!20} {\textbf{biopsied population}} } \\ \hline
        image-only & 0.781$\pm$0.006 & 0.673$\pm$0.003 & 0.791 & 0.682 \\ \hline
        heatmaps-only & 0.805$\pm$0.007 & 0.621$\pm$0.008 & 0.803 & 0.633 \\ \hline
        image-and-heatmaps & \textbf{0.843}$\pm$\textbf{0.004} & \textbf{0.690}$\pm$\textbf{0.002} & \textbf{0.850} & \textbf{0.696} \\ \hline
        \end{tabular}
        \label{tab:cancer_pred_input_variant}
    \end{table}

\subsection*{Correlation between model predictions}
    
\begin{figure*}[htb!]
        \centering
        \begin{tabular}{c c}
        \includegraphics[height=0.3\linewidth]{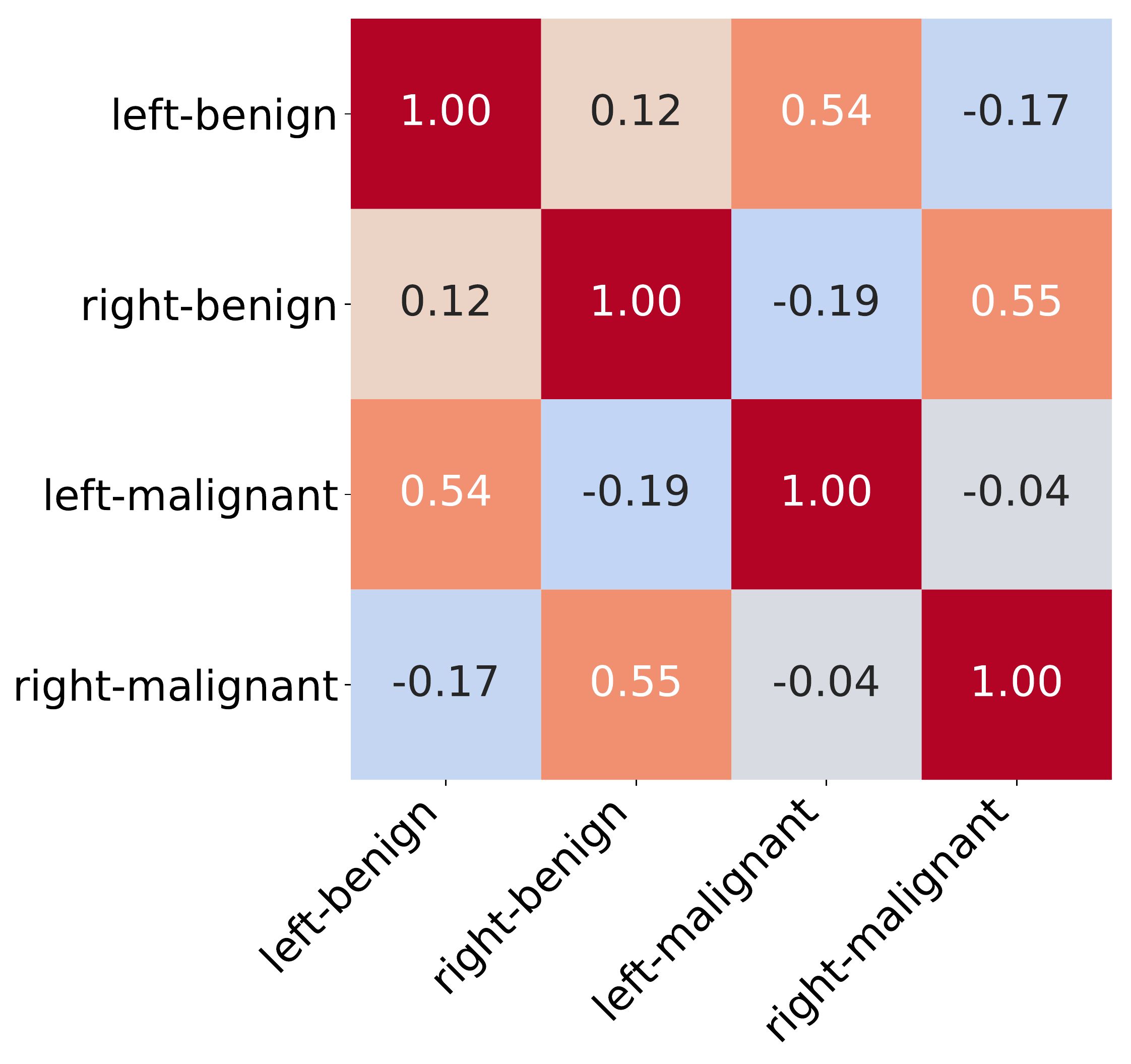}
        &
        \includegraphics[height=0.3\linewidth]{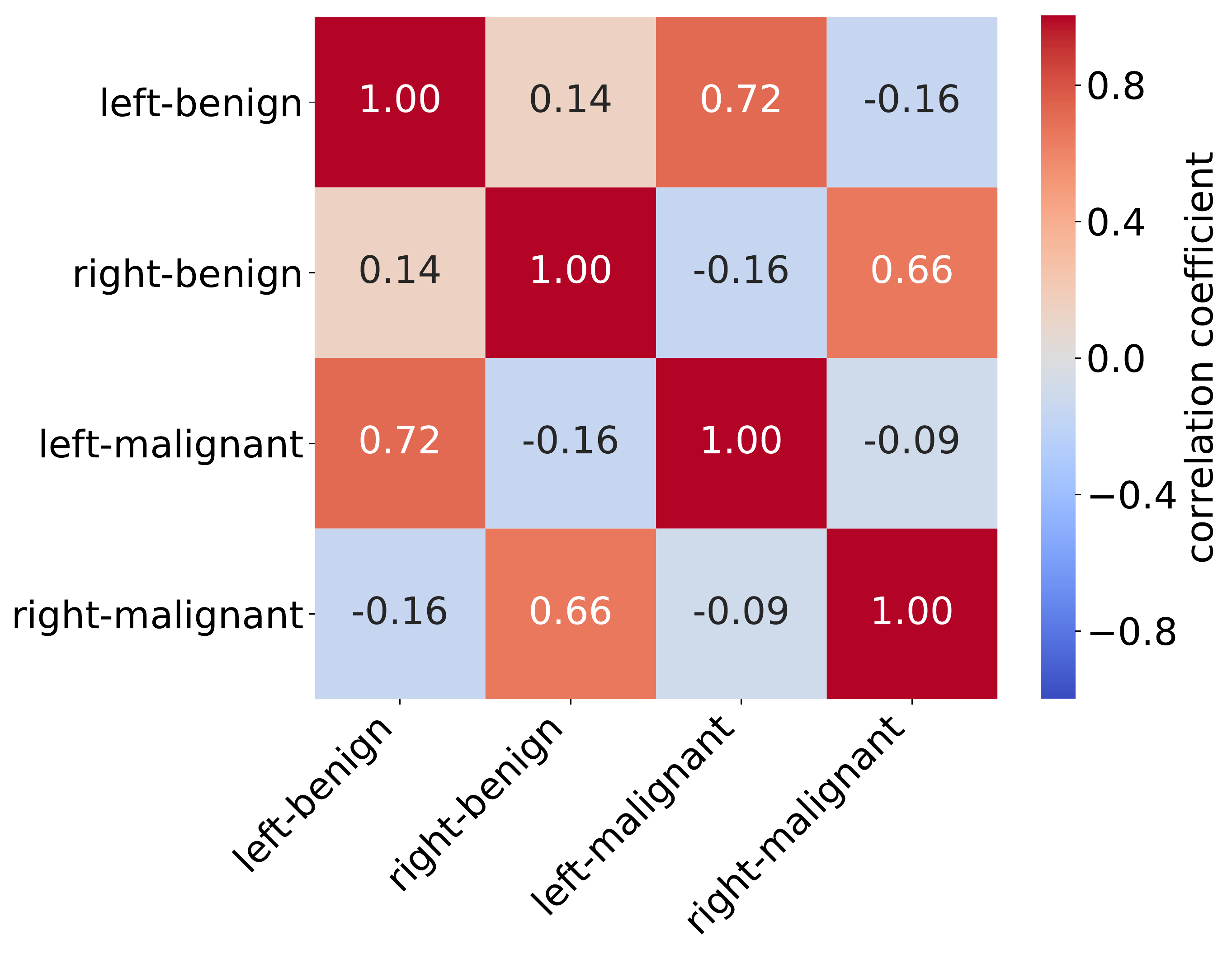}
        \\
        \footnotesize{(a) image-and-heatmaps} & \footnotesize{(b) image-only}
        \end{tabular}
        \vspace{-2mm}
        \caption{
            Correlations of model ensemble predictions across labels.
        }
        \label{fig:pred_correls}
    \end{figure*}

    We visualize in \autoref{fig:pred_correls} the correlations between model predictions for the four different labels (left-benign, left-malignant, right-benign, right-malignant) for a given exam. In both image-only and image-and-heatmaps model ensembles, we observe high correlations between benign and malignant predictions for the same breast, and low correlations for predictions between breasts. Notably, we observe a lower correlation between benign and malignant predictions for the same breast in the image-and-heatmaps model ensemble compared to the image-only model ensemble. This is consistent with other results showing that the image-and-heatmaps models are better able to distinguish between benign and malignant cases, likely due to the additional information from the class-specific heatmaps.

\subsection*{Comparison of CC and MLO model branches}
    \begin{table}[ht]
        \centering
        \caption{
            AUCs of CC and MLO model branches on screening and biopsied populations.
        }
        \begin{tabular}{| l | c | c | c | c |}
        \cline{2-5}
        \multicolumn{1}{c|}{} & \multicolumn{2}{c|}{single} & \multicolumn{2}{c|}{5x ensemble} \\
        \cline{2-5}
        \multicolumn{1}{c|}{} & malignant & benign & malignant & benign \\ 
        \cline{2-5}
        \hline
        \multicolumn{5}{|c|}{\cellcolor{gray!20} {\textbf{screening population}} } \\ \hline
        image-only  & 0.827$\pm$0.008 & 0.731$\pm$0.004 & 0.840 & 0.743 \\ \hline
        image-only (CC) & 0.813$\pm$0.009 & 0.726$\pm$0.004 & 0.830 & 0.739 \\ \hline
        image-only (MLO) & 0.766$\pm$0.012 & 0.691$\pm$0.006 & 0.776 & 0.705 \\ \hline
        image-and-heatmaps  & \textbf{0.886}$\pm$\textbf{0.003} & \textbf{0.747}$\pm$\textbf{0.002} & \textbf{0.895} & \textbf{0.756} \\ \hline
        image-and-heatmaps (CC) & 0.873$\pm$0.006 & 0.740$\pm$0.005 & 0.891 & 0.752 \\ \hline
        image-and-heatmaps (MLO) & 0.834$\pm$0.002 & 0.703$\pm$0.002 & 0.847 & 0.712 \\ \hline
        \multicolumn{5}{|c|}{\cellcolor{gray!20} {\textbf{biopsied population}} } \\ \hline
        image-only  & 0.781$\pm$0.006 & 0.673$\pm$0.003 & 0.791 & 0.682 \\ \hline
        image-only (CC) & 0.774$\pm$0.006 & 0.677$\pm$0.003 & 0.786 & 0.687 \\ \hline
        image-only (MLO) & 0.732$\pm$0.012 & 0.638$\pm$0.002 & 0.743 & 0.649 \\ \hline
        image-and-heatmaps  & \textbf{0.843}$\pm$\textbf{0.004} & \textbf{0.690}$\pm$\textbf{0.002} & \textbf{0.850} & 0.696 \\ \hline
        image-and-heatmaps (CC) & 0.833$\pm$0.004 & \textbf{0.690}$\pm$\textbf{0.004} & 0.847 & \textbf{0.699} \\ \hline
        image-and-heatmaps (MLO) & 0.802$\pm$0.003 & 0.650$\pm$0.002 & 0.813 & 0.656 \\ \hline
        \end{tabular}
        \label{tab:cancer_pred_views}
    \end{table}
    
    The architecture of our model can be decomposed into two separate but symmetric deep neural network models, operating on CC view and MLO view images respectively, which we refer to as the CC and MLO branches of the model. Each branch individually computes predictions for all four labels, and the full model's final prediction is the average of the  predictions of both branches.
    We show in \autoref{tab:cancer_pred_views} the breakdown of the performance of the CC and MLO branches of our model. We observe a fairly consistent trend of the CC model branch outperforming the MLO model branch, across multiple contexts (malignant/not malignant classification, benign/not benign classification, with or without the heatmaps). The superior performance of the CC model branch is consistent with the view of radiologists that findings may be more conspicuous and better visualized in the CC view compared to the MLO view. The predictions of the full model generally outperform using the predictions of either branch individually, except in the case of benign prediction for the biopsied population, where the CC model branch slightly outperforms the averaged prediction.
    
\subsection*{Classifying malignant/benign vs. normal}
    In this section and the next, we further analyze the behavior of our model by decomposing the task of breast cancer classification into two sub-tasks: (i) determining if a breast has any findings, benign or malignant, and (ii) conditional on the presence of a finding, determining whether it is malignant or benign.
    
    First, we evaluated our models on the task of only predicting whether a single breast has either a malignant or a benign finding, or otherwise (negative for both malignant and benign findings). This is equivalent to predicting whether, for a given screening mammogram, a biopsy was subsequently performed. This evaluation is performed over the screening population. Without retraining the model, we took the maximum of malignant and benign predictions as the prediction of a biopsy. We obtained an AUC of 0.767 using the image-and-heatmaps model ensemble, with more results shown in \autoref{tab:cancer_pred_no_malbenvs}. The relatively small margin in performance between the image-only and image-and-heatmaps models indicates that the heatmaps are marginally less useful for the task of determining the presence of any finding at all.
    
    \begin{table}[ht]
        \centering
        \caption{
            AUCs of our models on screening and biopsied populations, on the task of classifying malignant/benign vs normal.
        }
        \begin{tabular}{| l | c | c |}
        \cline{2-3}
        \multicolumn{1}{c|}{} & \multicolumn{1}{c|}{single} & \multicolumn{1}{c|}{5x ensemble} \\
        \cline{2-3}
        \hline
        image-only & 0.740$\pm$0.003 & 0.752 \\ \hline
        image-and-heatmaps & \textbf{0.759}$\pm$\textbf{0.002} & \textbf{0.767} \\ \hline
        \end{tabular}
        \label{tab:cancer_pred_no_malbenvs}
    \end{table}
    
\subsection*{Classifying malignant vs. benign}
    Next, we investigated the ability of our models to distinguish between malignant and benign findings on exams where we know there is a mammographically visible finding--this isolates the discriminative capability of our model between types of findings.
    We constructed the \textit{one-class biopsied subpopulation}, with a subset of 384 breasts from our test set, comprising only of breasts labeled either only malignant or only benign. We evaluate our models on the ability to predict whether a finding in a given breast was malignant or benign.
    To adapt the predictions of our model to this binary classification task, we normalized the prediction probabilities for the two classes to sum to one and calculated AUCs based on these normalized predictions. The image-only ensemble attained an AUC of 0.738 while the image-and-heatmaps ensemble attained an AUC of 0.803. This, along with the results above, provides evidence that the heatmaps help primarily in distinguishing between benign and malignant findings.
    
\subsection*{Additional reader study analysis}

We supplement the results of our reader study shown in the main paper with additional analysis in this section.
 
\subsubsection*{Reader ensemble and hybrid ensemble} 
\begin{figure}[h!]
\centering
    \begin{tabular}{c c}
         \includegraphics[width=0.4\linewidth]{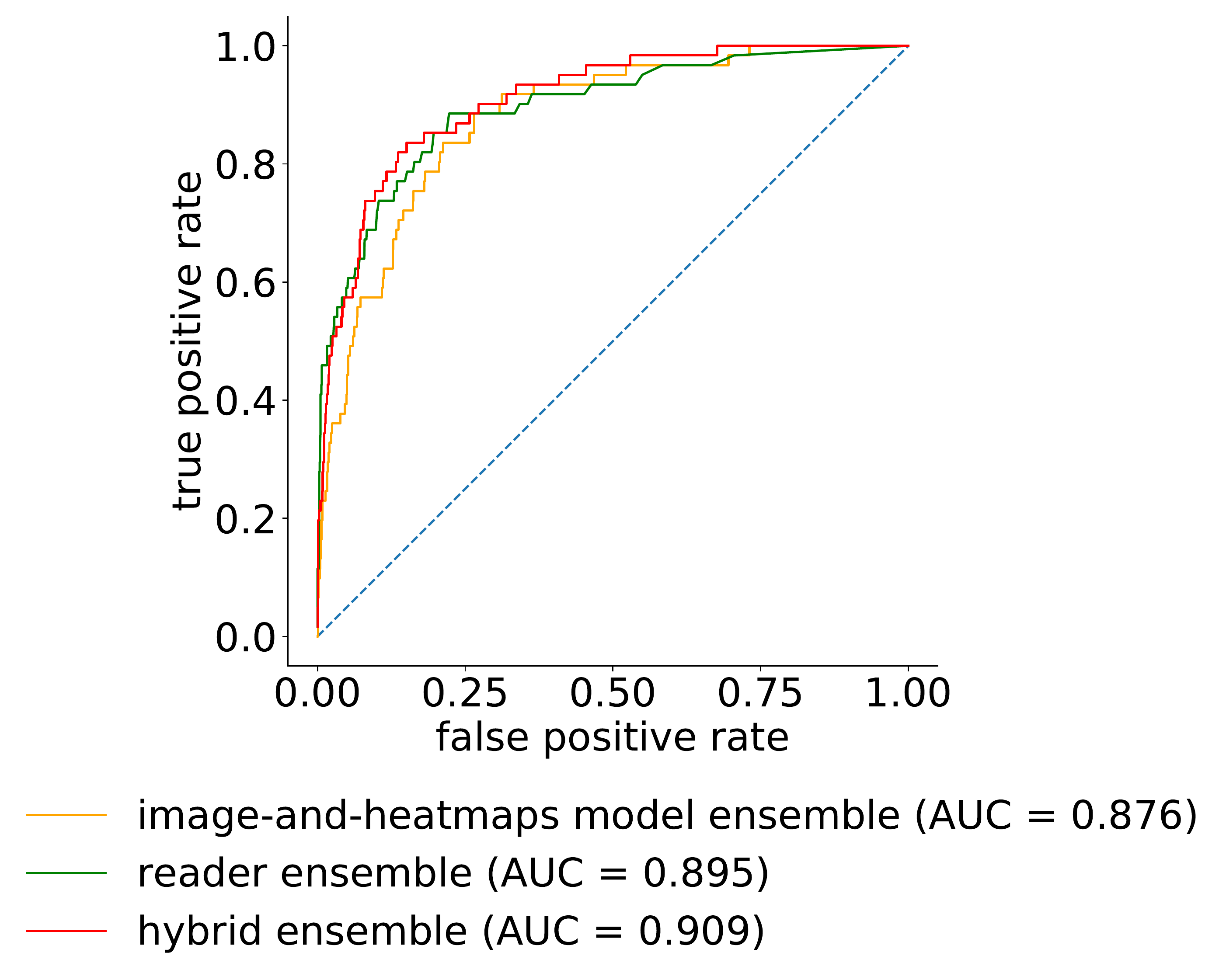} & \includegraphics[width=0.4\linewidth]{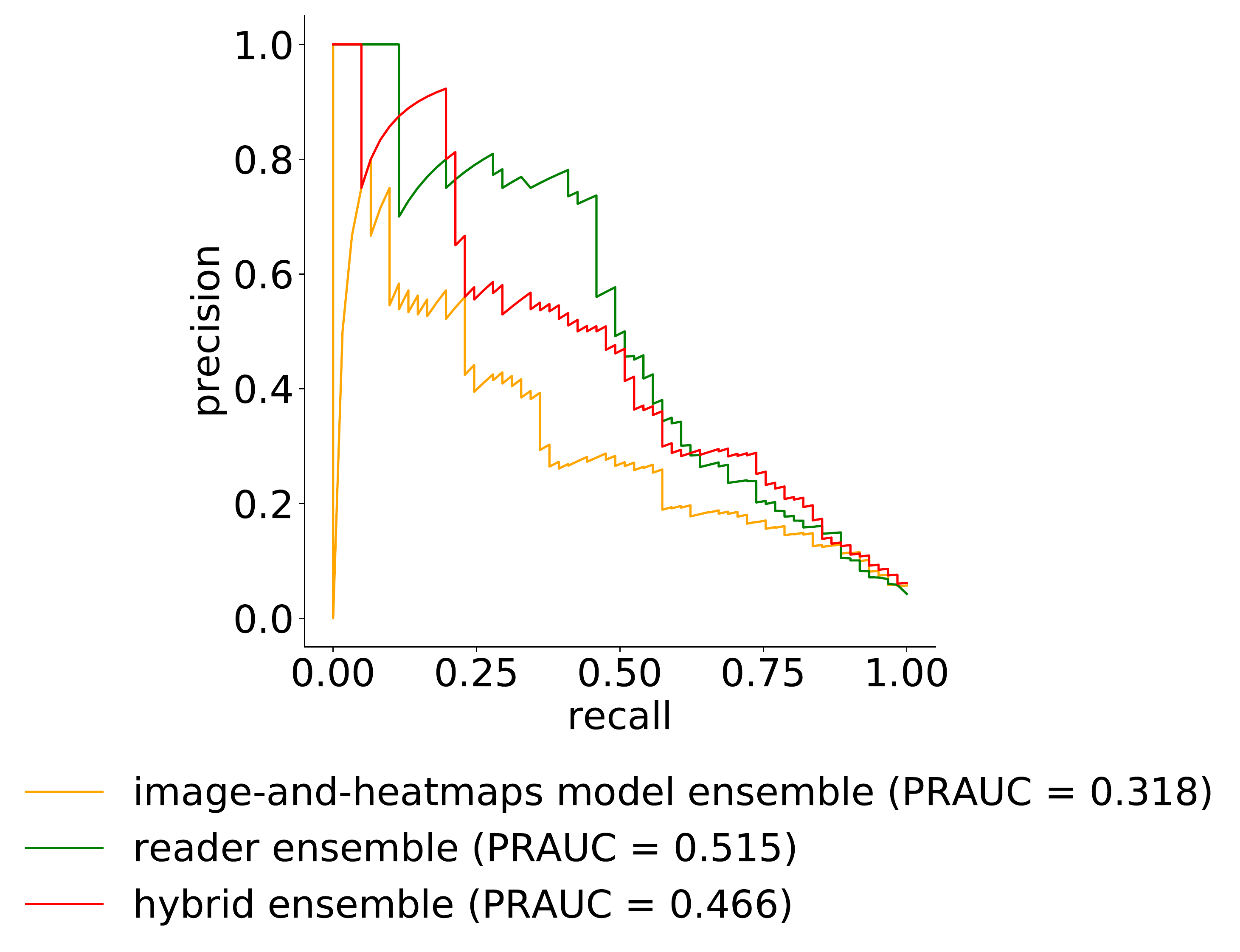}
    \end{tabular}
    \caption{ROC curves and precision-recall curves of reader ensemble, hybrid ensemble and our image-and-heatmaps model ensemble.}
    \label{fig:reader_ensemble}
\end{figure}

We evaluate a \textit{reader ensemble} by averaging the predictions across our 14 readers. We also evaluate a \textit{hybrid ensemble} by averaging the predictions of the ensemble of readers with our image-and-heatmaps model ensemble, where we equally weight both sets of predictions. \autoref{fig:reader_ensemble} shows the ROC curves and precision-recall curves of these two ensembles compared to our model ensemble alone. We observe that the hybrid ensemble outperforms the reader ensemble based on AUC, but underperforms based on PRAUC. This suggests that although the combination of our model a single radiologist tends to lead to improved accuracy, the benefit that our model could provide to a group of radiologists is far more limited. 

\subsubsection*{Representation learned by image-only model} 
\begin{figure}[!htb]
    \centering
    \begin{tabular}{c c c c}
    \phantom{abcdefghihihi}& \hspace{-2mm}\includegraphics[height=0.32\linewidth]{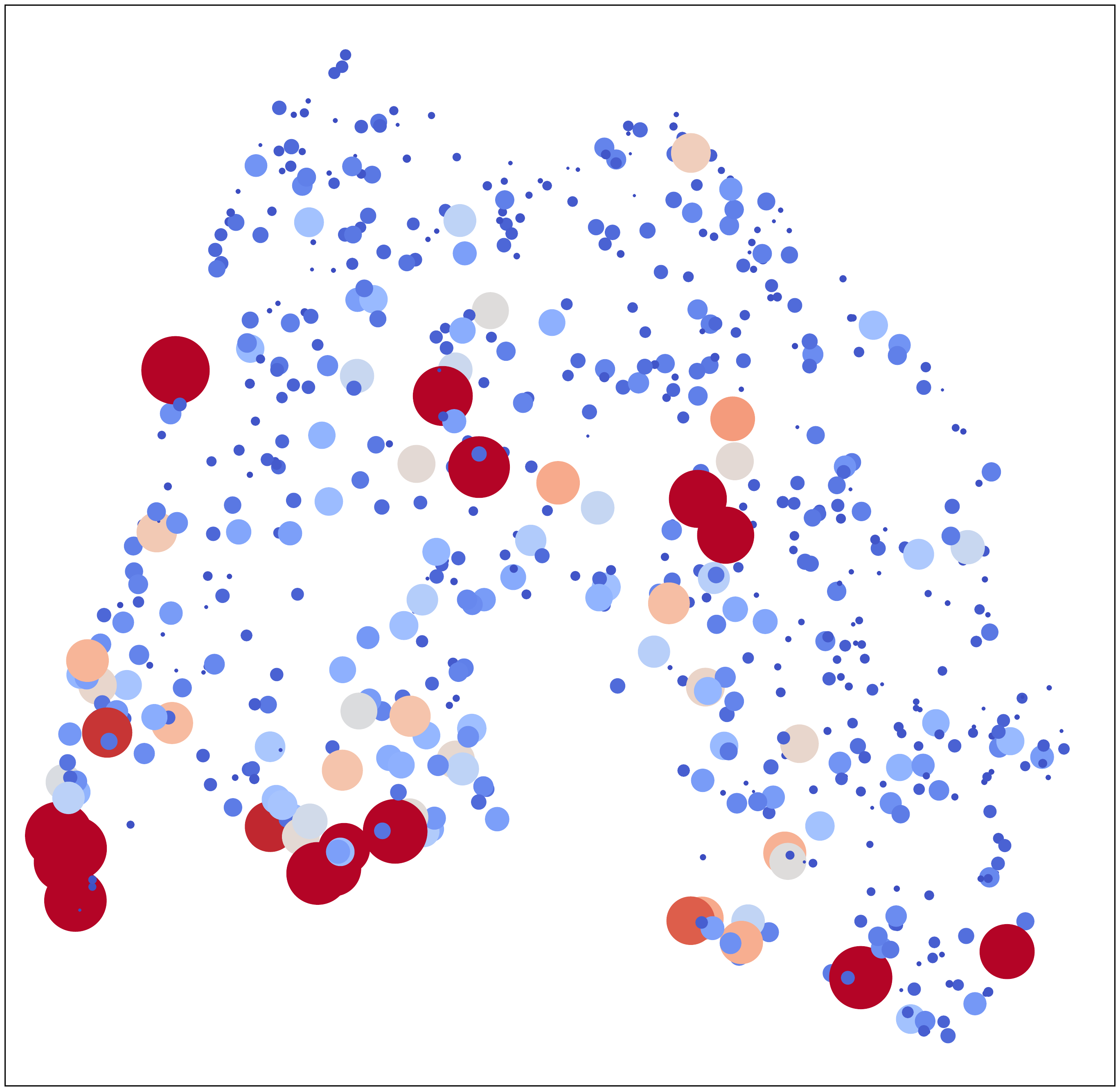}&\hspace{-4mm} \includegraphics[height=0.32\linewidth]{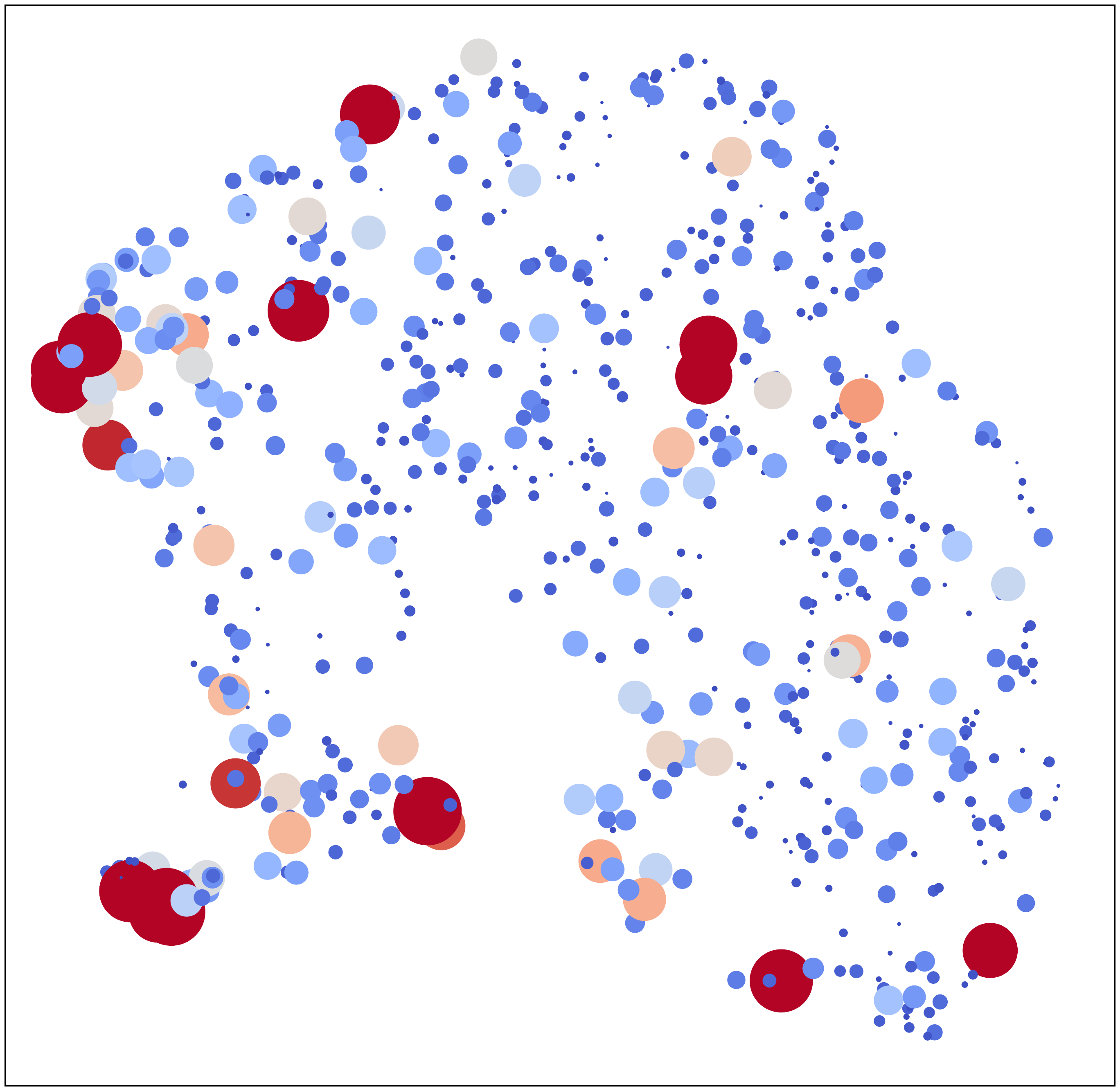} & 
    \raisebox{.16\height}{\includegraphics[height=0.22\linewidth, trim=3mm 0mm 10mm 0mm]{figures/tsne_reader_study_viewsplit_colorbar.pdf}}
    \end{tabular}
    \vspace{-2mm}
    \caption{Exams in the reader study set represented using the concatenated activations from the four image-specific columns (left) and the concatenated activations from the first fully connected layer in both CC and MLO model branches (right). The above activations are learned by the best image-only model. }
    \label{fig:tsne_breast}
\end{figure}

We visualize the hidden representations learned by the best image-only model in addition to the best image-and-heatmaps model, computed on the same reader study subpopulation (cf. \autoref{fig:tsne_breast}).
Compared to the distribution of representations learned by the image-and-heatmaps model, exams classified as more likely to be malignant according to readers were spread more randomly over the space but group together in several clusters. This pattern is apparent in both sets of activations, and suggests that the learned representational space that the mammograms are projected to are better conditioned for malignancy classification for the image-and-heatmaps model compared to the image-only model.

\subsubsection*{Error analysis}
\begin{figure}[!htb]
    \centering
    \begin{minipage}[t]{.31\textwidth}
    \centering
    \begin{tabular}{c c c}
         \hspace{-5.5mm}\includegraphics[width=0.37\linewidth, height=0.51\linewidth]{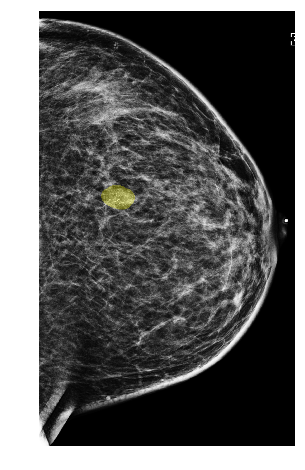} &  \hspace{-6.0mm}\includegraphics[width=0.37\linewidth, height=0.51\linewidth]{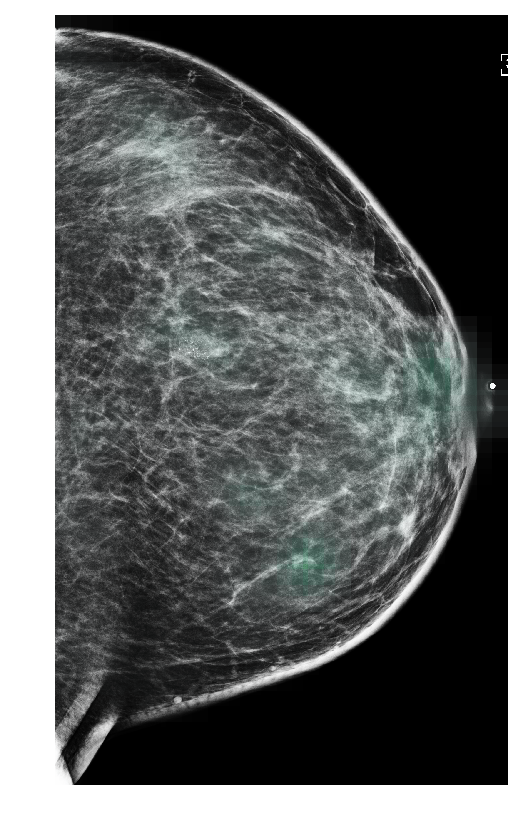}& \hspace{-5.5mm}\includegraphics[width=0.37\linewidth, height=0.51\linewidth]{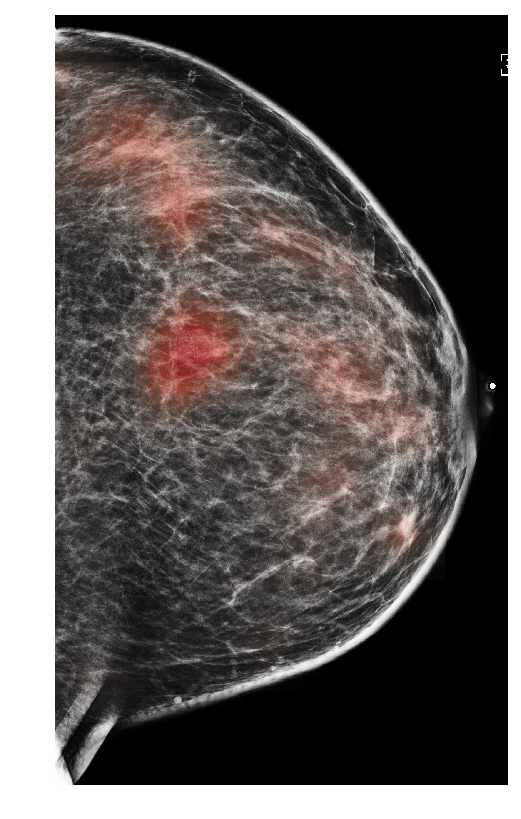}  \\
         \hspace{-5.5mm}\includegraphics[width=0.37\linewidth, height=0.51\linewidth]{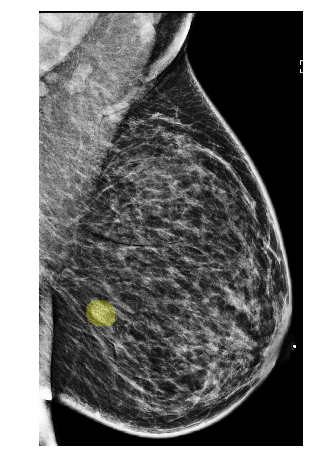} & \hspace{-6.0mm}\includegraphics[width=0.37\linewidth, height=0.51\linewidth]{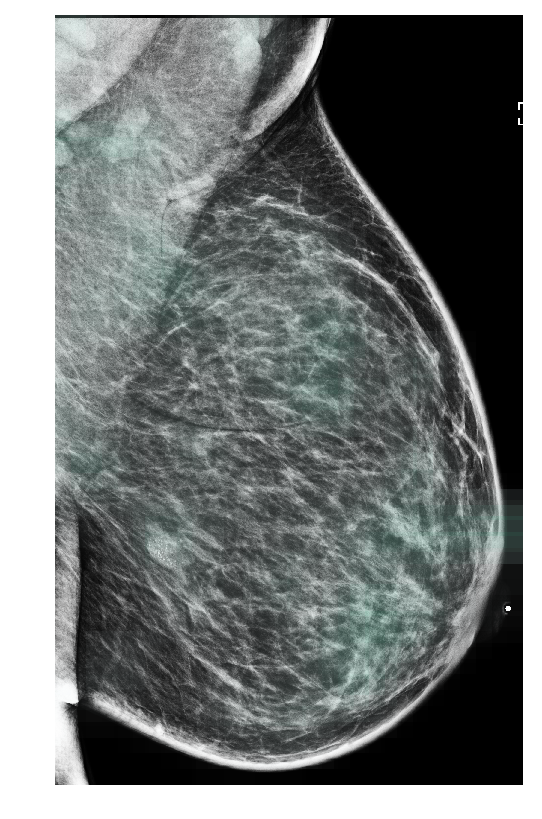} & \hspace{-5.5mm}\includegraphics[width=0.37\linewidth, height=0.51\linewidth]{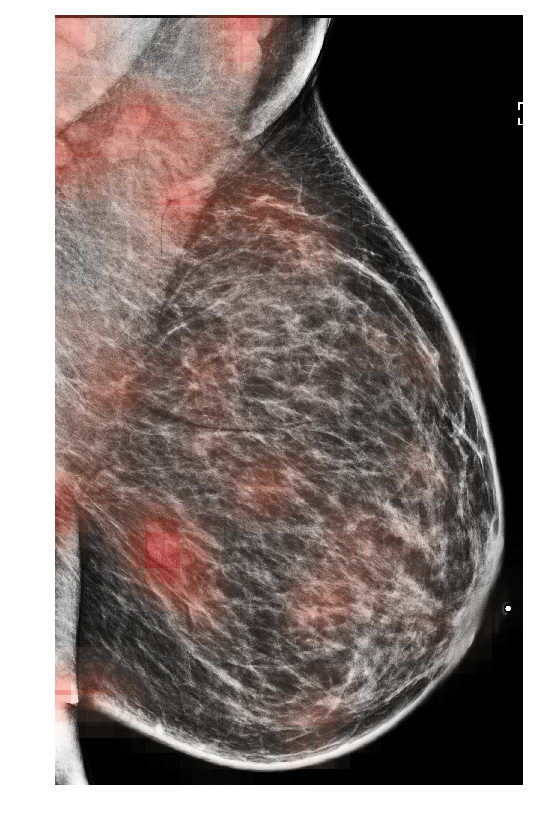} 
    \end{tabular}
    \vspace{-4mm}
    \caption{A biopsy-proven high-risk benign finding marked in yellow on both CC (top row) and MLO views (the second row) in the patient's right breast. The heatmaps overlying the images (green for benign and red for malignant) are shown after the images with segmentation. The malignant score for this breast given by the model is 0.997 while the benign score is 0.909. The `malignant' heatmap highlighted the marked area but the `benign' heatmap did not, for both CC and MLO views. The mean malignant score given by the 14 readers is 0.699, with 12 readers giving scores over 0.6. 
    }
    \label{fig:error_analysis_case_1}
    \end{minipage}
    \hspace{3mm}
    \begin{minipage}[t]{.31\textwidth}
    \centering
    \begin{tabular}{c c c}
         \hspace{-5.5mm}\includegraphics[width=0.37\linewidth, height=0.51\linewidth]{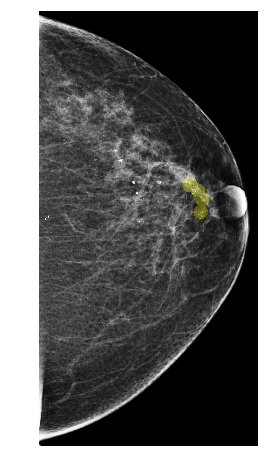} & \hspace{-6.0mm}\includegraphics[width=0.37\linewidth, height=0.51\linewidth]{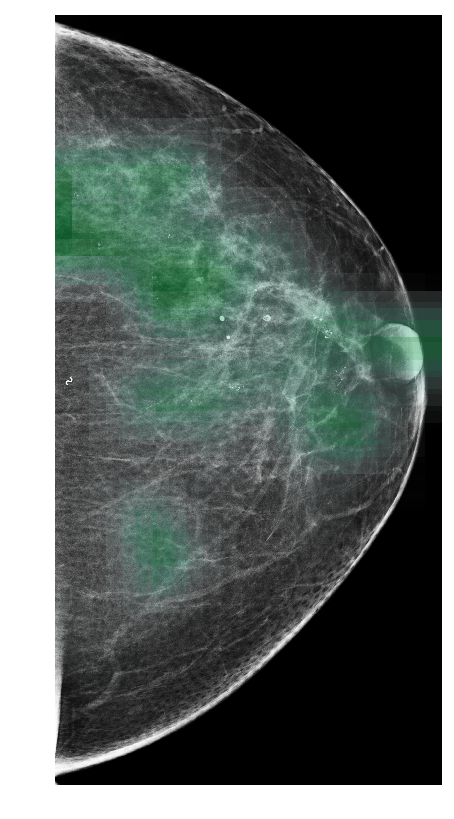}& \hspace{-5.5mm}\includegraphics[width=0.37\linewidth, height=0.51\linewidth]{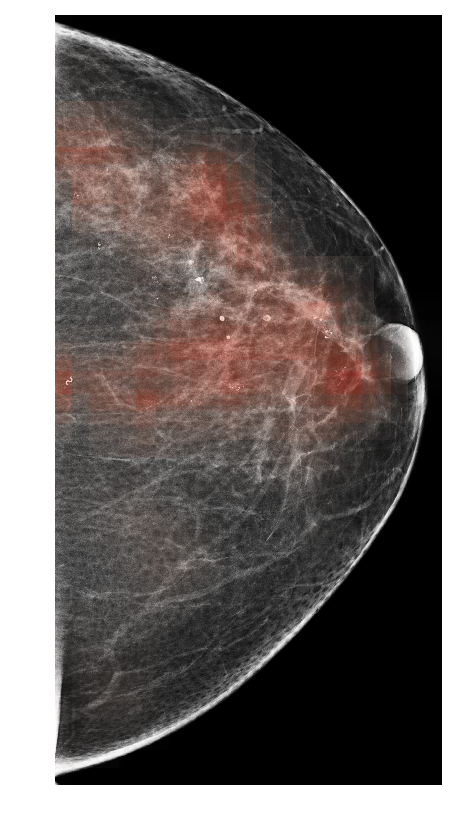}  \\
         \hspace{-5.5mm}\includegraphics[width=0.37\linewidth, height=0.51\linewidth]{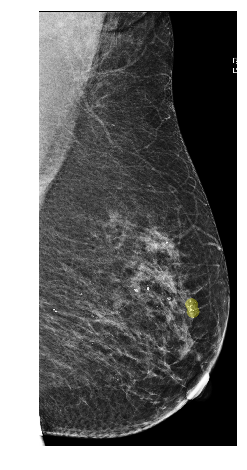} & \hspace{-6.0mm}\includegraphics[width=0.37\linewidth, height=0.51\linewidth]{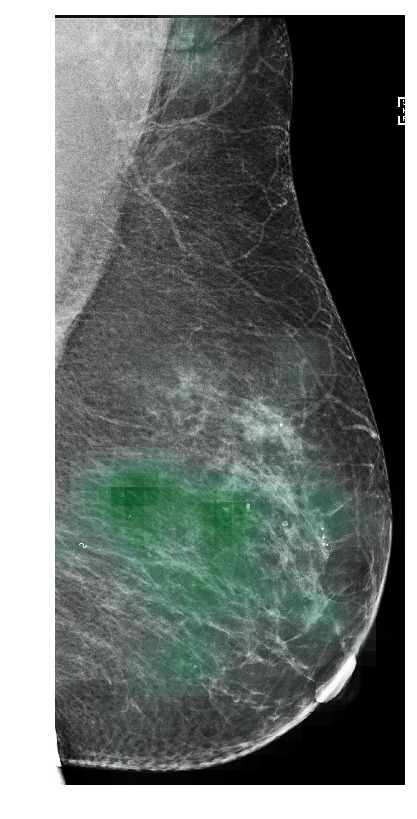}& \hspace{-5.5mm}\includegraphics[width=0.37\linewidth, height=0.51\linewidth]{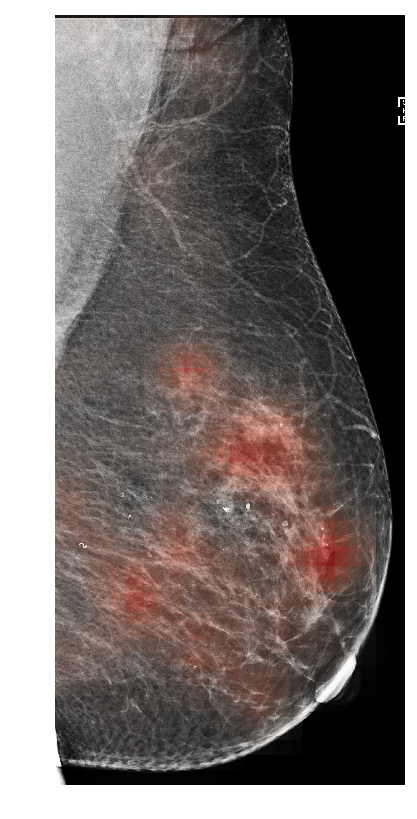} 
    \end{tabular}
    \vspace{-4mm}
    \caption{A region with biopsy-proven high-risk benign findings marked in yellow on both the CC view and the MLO view in the patient's left breast. Images and heatmaps are shown with the same layout as in \autoref{fig:error_analysis_case_1}. Compared with the case in \autoref{fig:error_analysis_case_1}, the gap between the malignant and benign scores given by the model for this breast is larger--the malignant score is 0.709 and the benign score is 0.433. The highest malignant scores given by the 14 readers is 0.25 and the mean is only 0.03. 
    }
    \label{fig:error_analysis_case_2}
    \end{minipage}
    \hspace{3mm}
    \begin{minipage}[t]{.31\textwidth}
    \centering
    \begin{tabular}{c c c}
        \hspace{-5.5mm} \includegraphics[width=0.37\linewidth, height=0.51\linewidth]{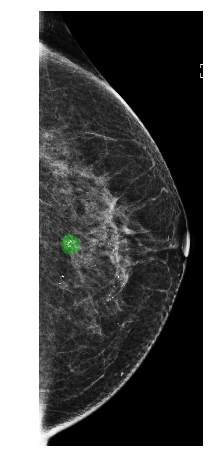} &  \hspace{-6.0mm}\includegraphics[width=0.37\linewidth, height=0.51\linewidth]{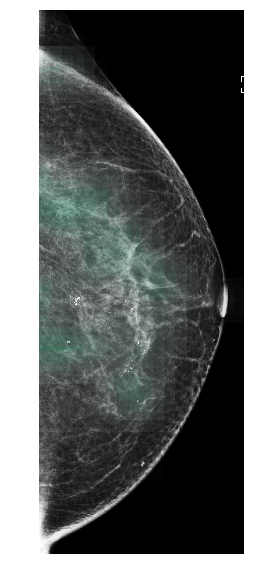}& \hspace{-5.5mm}\includegraphics[width=0.37\linewidth, height=0.51\linewidth]{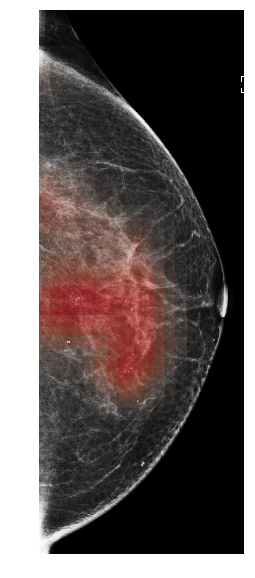}  \\
         \hspace{-5.5mm}\includegraphics[width=0.37\linewidth, height=0.51\linewidth]{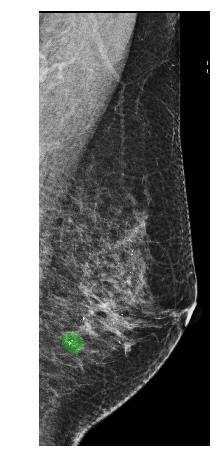} & \hspace{-6.0mm}\includegraphics[width=0.37\linewidth, height=0.51\linewidth]{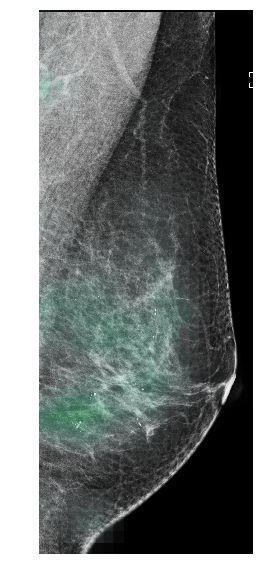} & \hspace{-5.5mm}\includegraphics[width=0.37\linewidth, height=0.51\linewidth]{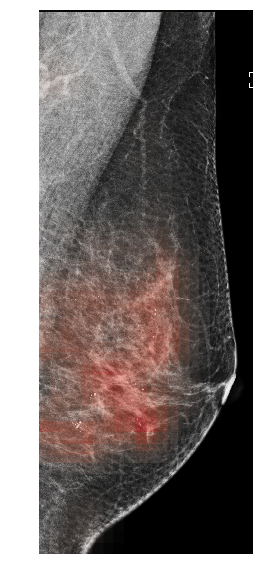} 
    \end{tabular}
    \vspace{-4mm}
    \caption{A biopsy-proven benign finding marked in green on both the CC view and the MLO view in the patient's left breast. Images and heatmaps are shown with the same layout as in \autoref{fig:error_analysis_case_1}. The malignant score for this breast given by the model is 0.735 while the benign score is 0.549. Readers were highly confident that this case was benign and their mean malignant score is 0.05, with the highest score being only 0.2.     }
    \label{fig:error_analysis_case_3}
    \end{minipage}
\end{figure}

\begin{figure}[!htb]
    \centering
    \begin{minipage}[t]{.31\textwidth}
    \centering
    \begin{tabular}{c c c}
         \hspace{-5.5mm}\includegraphics[width=0.37\linewidth, height=0.51\linewidth]{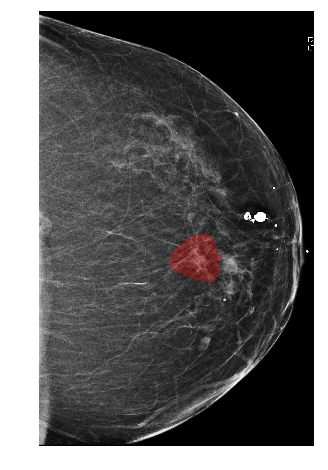} &  \hspace{-6.0mm}\includegraphics[width=0.37\linewidth, height=0.51\linewidth]{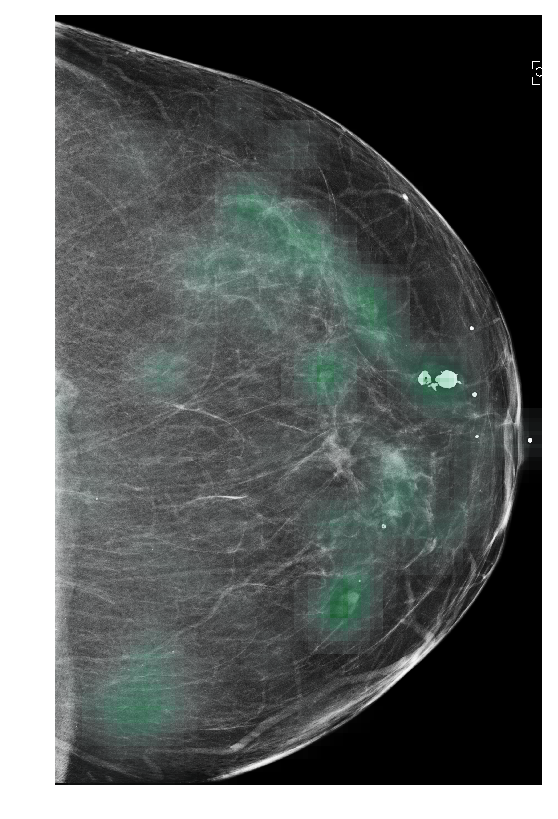}& \hspace{-5.5mm}\includegraphics[width=0.37\linewidth, height=0.51\linewidth]{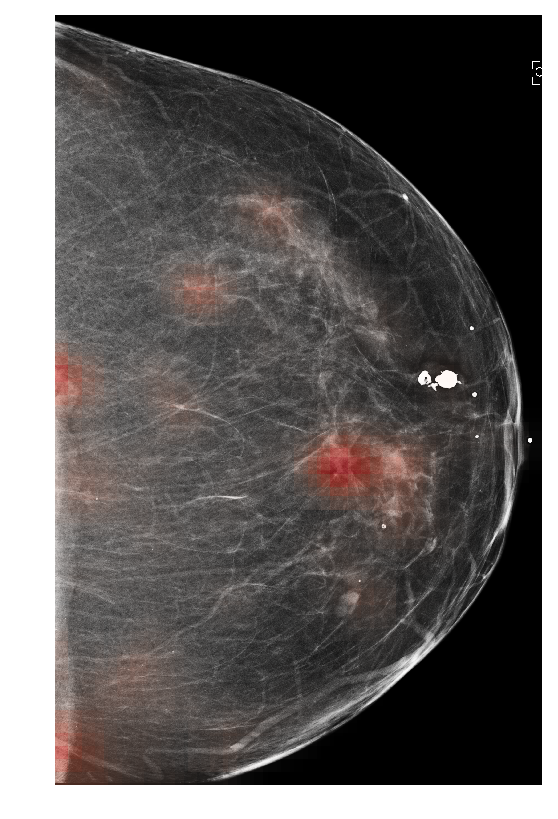}  \\
         \hspace{-5.5mm}\includegraphics[width=0.37\linewidth, height=0.51\linewidth]{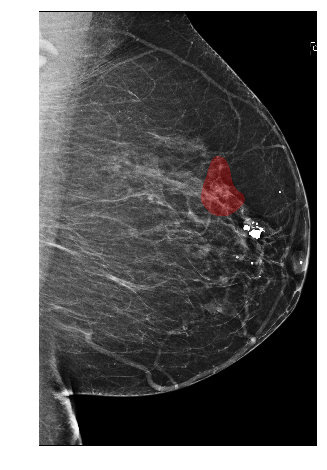} & \hspace{-6.0mm}\includegraphics[width=0.37\linewidth, height=0.51\linewidth]{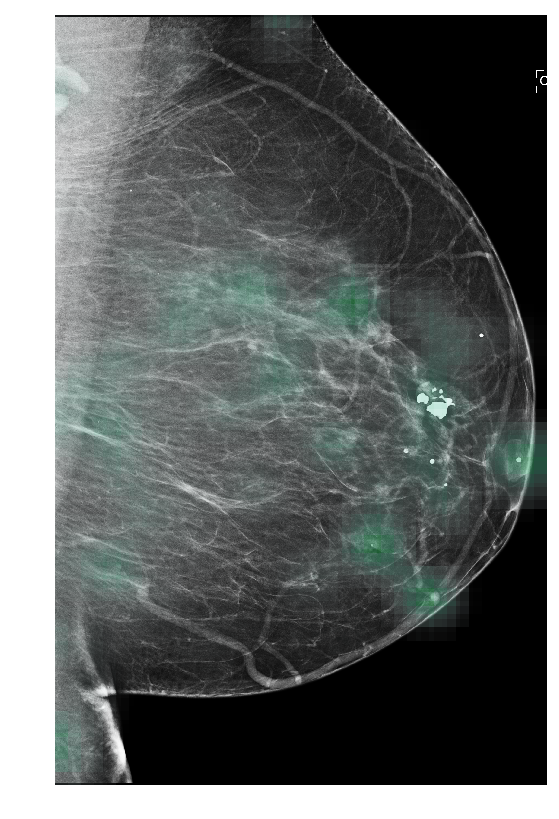} & \hspace{-5.5mm}\includegraphics[width=0.37\linewidth, height=0.51\linewidth]{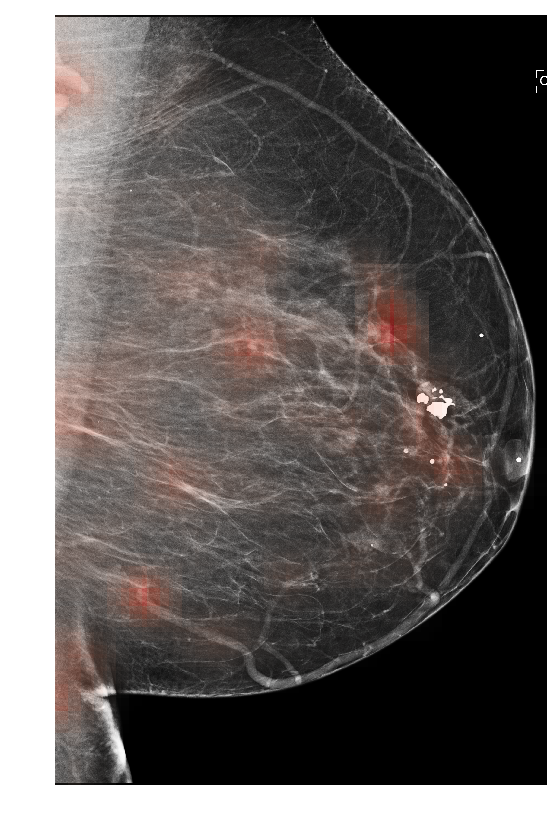} 
    \end{tabular}
    \vspace{-4mm}
    \caption{A biopsy-proven malignant finding marked in red on both the CC view and the MLO view in the patient's right breast. Images and heatmaps are shown with the same layout as in \autoref{fig:error_analysis_case_1}. The malignant score for this breast given by the model is 0.210 while the benign score is 0.621. Doctors' mean malignant score is 0.459 and eight readers among the 14 provided a score higher than 0.5. 
    }
    \label{fig:error_analysis_case_4}
    \end{minipage}
    \hspace{3mm}
    \begin{minipage}[t]{.31\textwidth}
    \centering
    \begin{tabular}{c c c}
         \hspace{-5.5mm}\includegraphics[width=0.37\linewidth, height=0.51\linewidth]{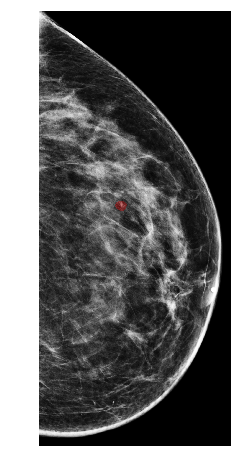} & \hspace{-6.0mm}\includegraphics[width=0.37\linewidth, height=0.51\linewidth]{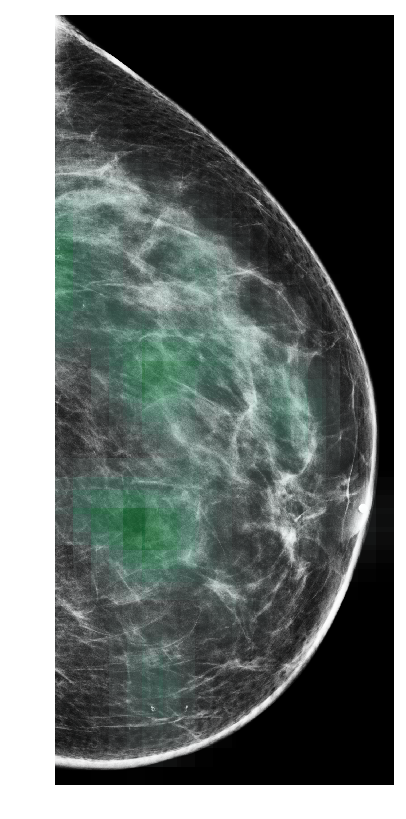}& \hspace{-5.5mm}\includegraphics[width=0.37\linewidth, height=0.51\linewidth]{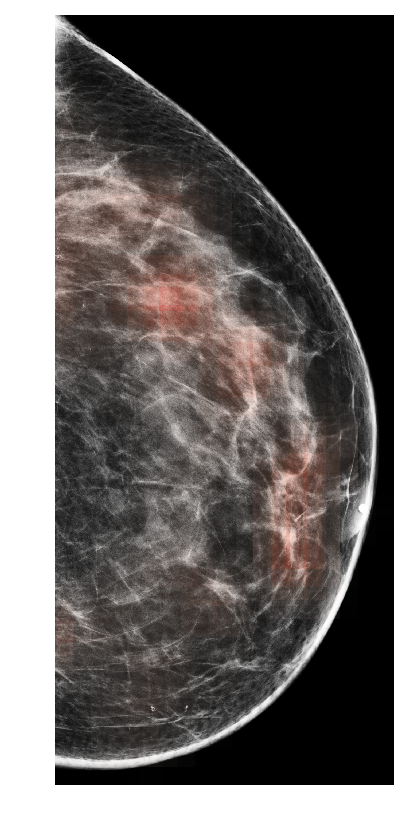}  \\
         \hspace{-5.5mm}\includegraphics[width=0.37\linewidth, height=0.51\linewidth]{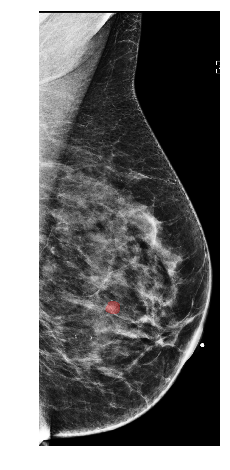} & \hspace{-6.0mm}\includegraphics[width=0.37\linewidth, height=0.51\linewidth]{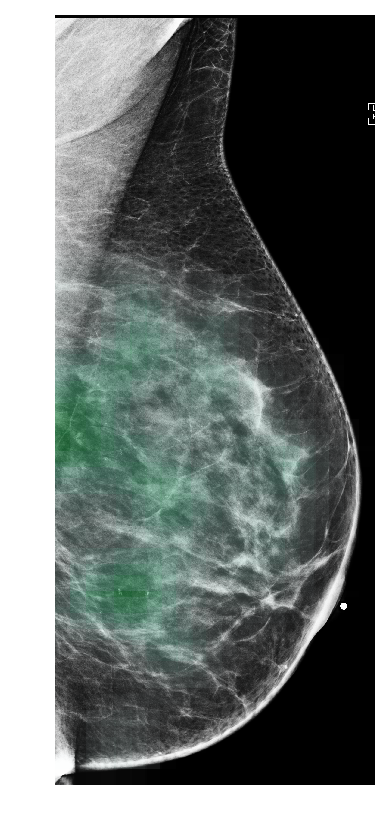}& \hspace{-5.5mm}\includegraphics[width=0.37\linewidth, height=0.51\linewidth]{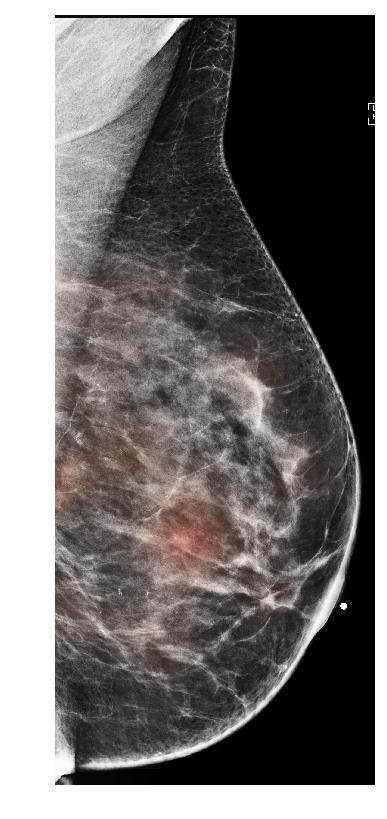} 
    \end{tabular}
    \vspace{-4mm}
    \caption{A biopsy-proven malignant finding marked in red on both the CC view and the MLO view in the patient's left breast. And the breast was also labeled as benign according to related pathology reports. Images and heatmaps are shown with the same layout as in \autoref{fig:error_analysis_case_1}. The malignant score given by the model is 0.068 and benign score is 0.433. The mean malignant score given by the 14 readers is 0.176, with the highest being only 0.30.     }
    \label{fig:error_analysis_case_5}
    \end{minipage}
    \hspace{3mm}
    \begin{minipage}[t]{.31\textwidth}
    \centering
    \begin{tabular}{c c c}
         \hspace{-5.5mm}\includegraphics[width=0.37\linewidth, height=0.51\linewidth]{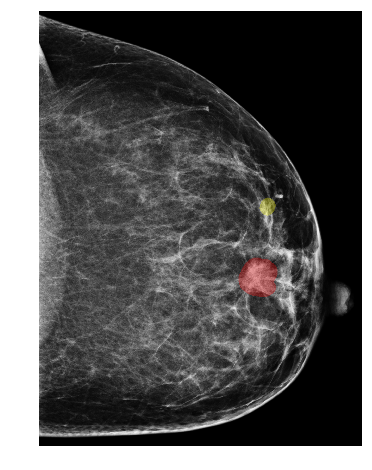} &  \hspace{-6.0mm}\includegraphics[width=0.37\linewidth, height=0.51\linewidth]{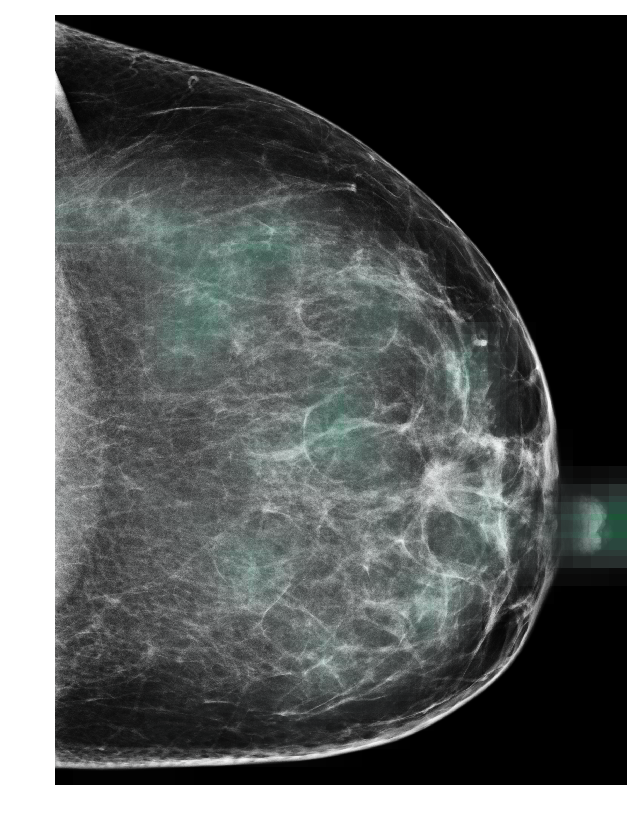}& \hspace{-5.5mm}\includegraphics[width=0.37\linewidth, height=0.51\linewidth]{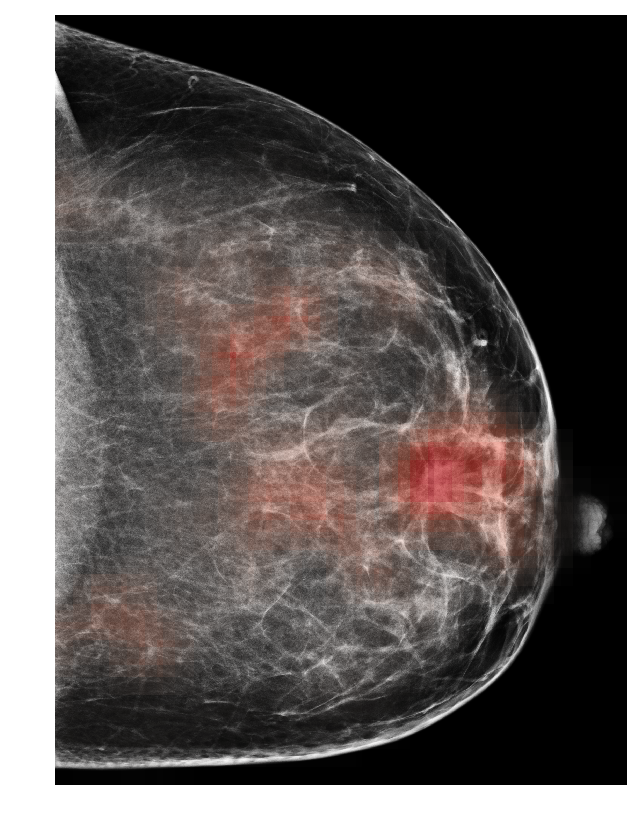}  \\
         \hspace{-5.5mm}\includegraphics[width=0.37\linewidth, height=0.51\linewidth]{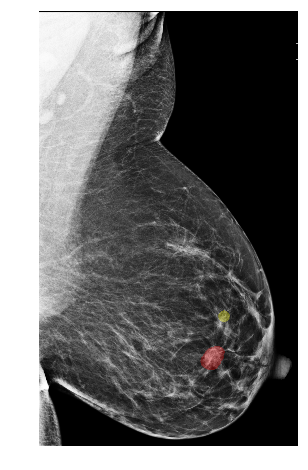} & \hspace{-6.0mm}\includegraphics[width=0.37\linewidth, height=0.51\linewidth]{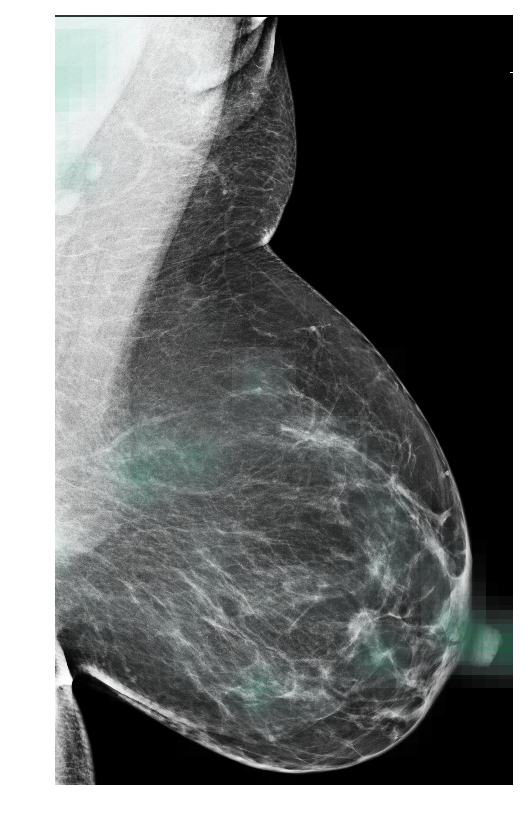} & \hspace{-5.5mm}\includegraphics[width=0.37\linewidth, height=0.51\linewidth]{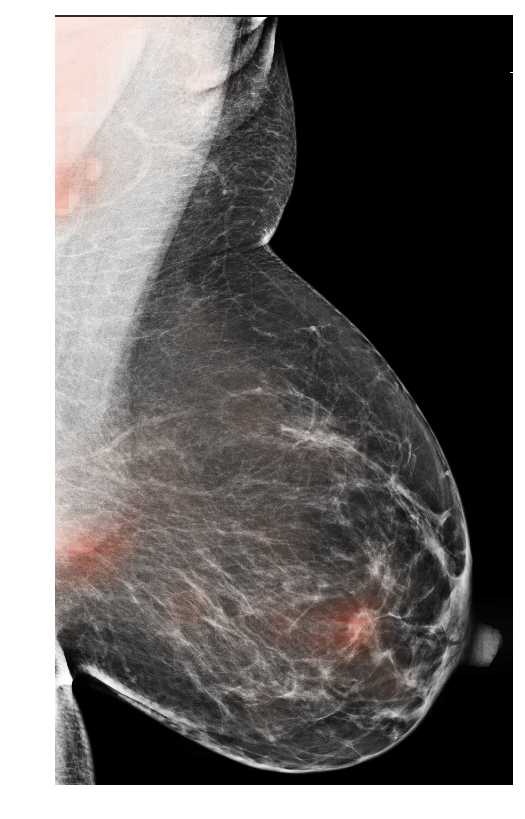} 
    \end{tabular}
    \vspace{-4mm}
    \caption{A biopsy-proven high-risk benign finding marked in yellow and a malignant finding marked in red on both the CC view and the MLO view in the patient's right breast. Images and heatmaps are shown with the same layout as in \autoref{fig:error_analysis_case_1}. The malignant score for this breast given by the model is only 0.044 while the benign score is 0.162. Doctors were confident that it was highly suspicious--their mean malignant score is 0.698 and 10 of them provided a probability estimate over 0.5. 
    }
    \label{fig:error_analysis_case_6}
    \end{minipage}
\end{figure}

\begin{figure}[!htb]
    \centering
    \begin{minipage}[t]{.31\textwidth}
    \centering
    \begin{tabular}{c c c}
         \hspace{-5.5mm}\includegraphics[width=0.37\linewidth, height=0.51\linewidth]{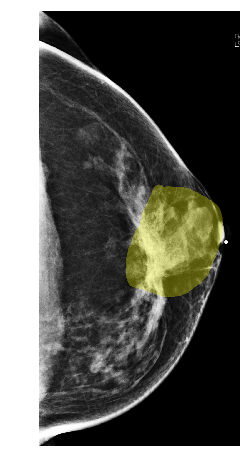} &  \hspace{-6.0mm}\includegraphics[width=0.37\linewidth, height=0.51\linewidth]{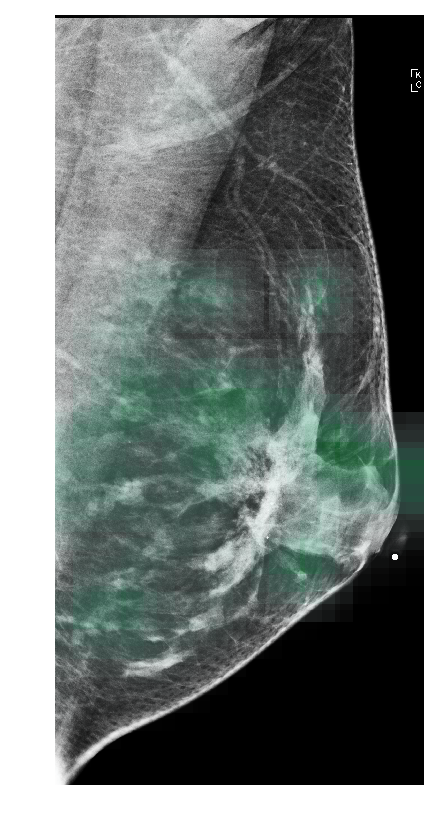}& \hspace{-5.5mm}\includegraphics[width=0.37\linewidth, height=0.51\linewidth]{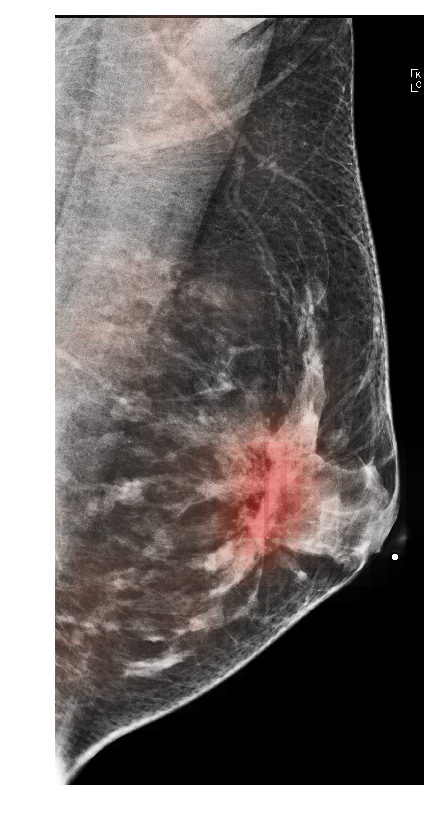}  \\
         \hspace{-5.5mm}\includegraphics[width=0.37\linewidth, height=0.51\linewidth]{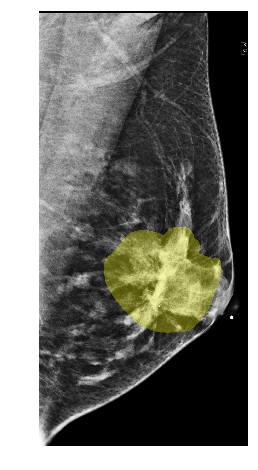} & \hspace{-6.0mm}\includegraphics[width=0.37\linewidth, height=0.51\linewidth]{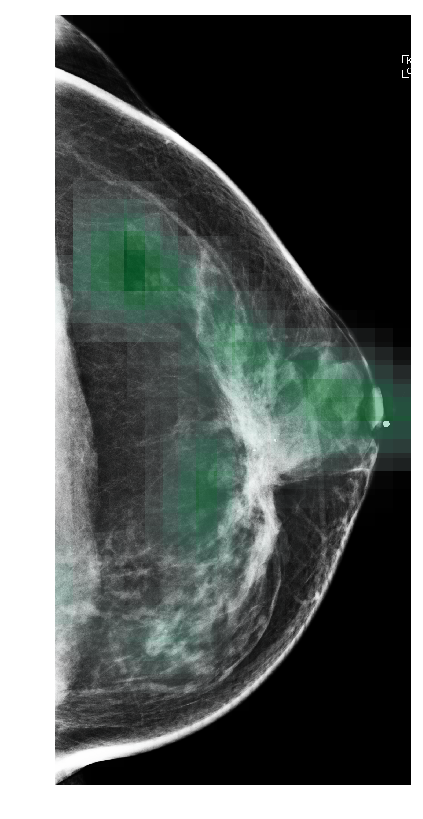} & \hspace{-5.5mm}\includegraphics[width=0.37\linewidth, height=0.51\linewidth]{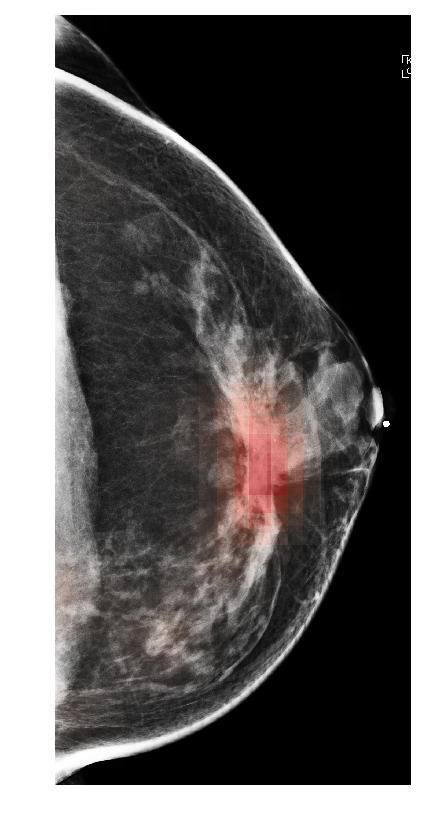} 
    \end{tabular}
    \vspace{-4mm}
    \caption{A biopsy-proven high-risk benign finding marked in yellow on both the CC view and the MLO view in the patient's left breast. Images and heatmaps are shown with the same layout as in \autoref{fig:error_analysis_case_1}. The malignant score for this breast given by the model is 0.124 and the benign score is 0.530. Readers' mean malignant score is 0.763. 
    }
    \label{fig:error_analysis_case_7}
    \end{minipage}
    \hspace{3mm}
    \begin{minipage}[t]{.31\textwidth}
    \centering
    \begin{tabular}{c c c}
         \hspace{-5.5mm}\includegraphics[width=0.37\linewidth, height=0.51\linewidth]{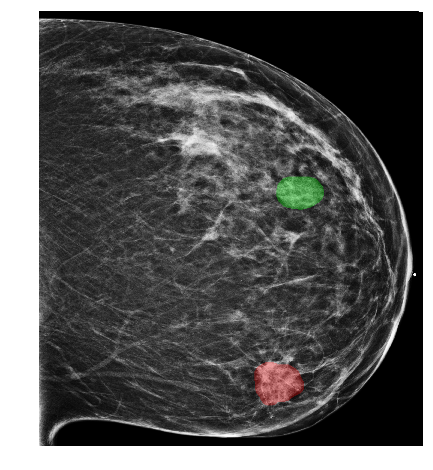} & \hspace{-6.0mm}\includegraphics[width=0.37\linewidth, height=0.51\linewidth]{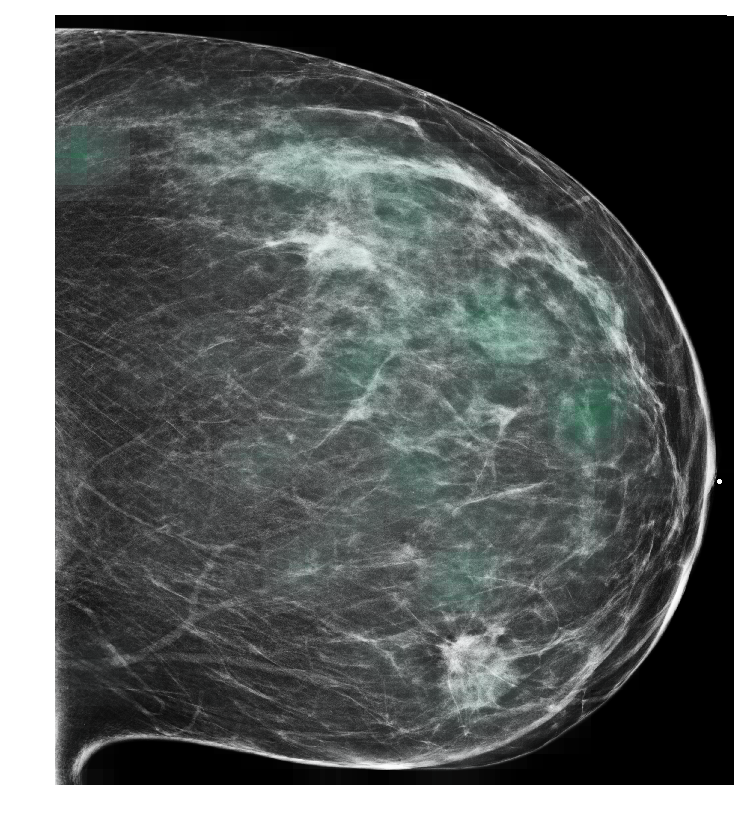}& \hspace{-5.5mm}\includegraphics[width=0.37\linewidth, height=0.51\linewidth]{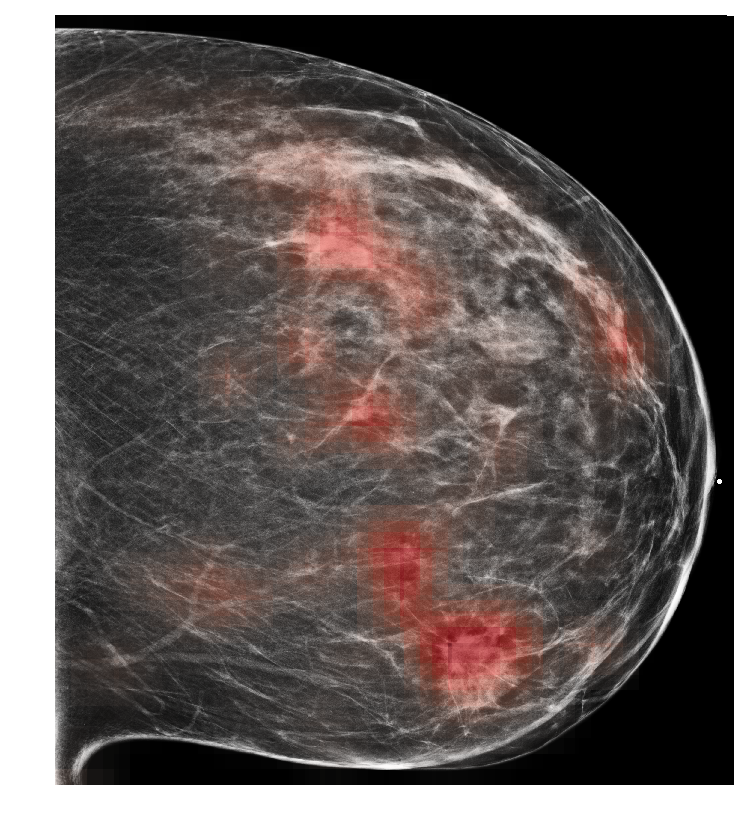}  \\
         \hspace{-5.5mm}\includegraphics[width=0.37\linewidth, height=0.51\linewidth]{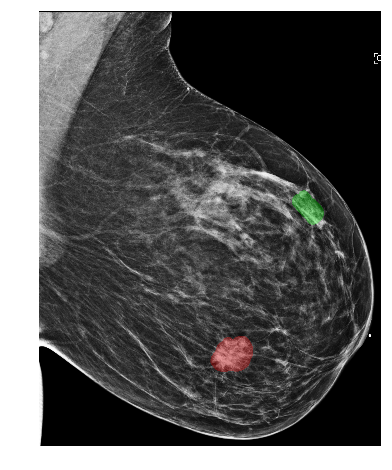} & \hspace{-6.0mm}\includegraphics[width=0.37\linewidth, height=0.51\linewidth]{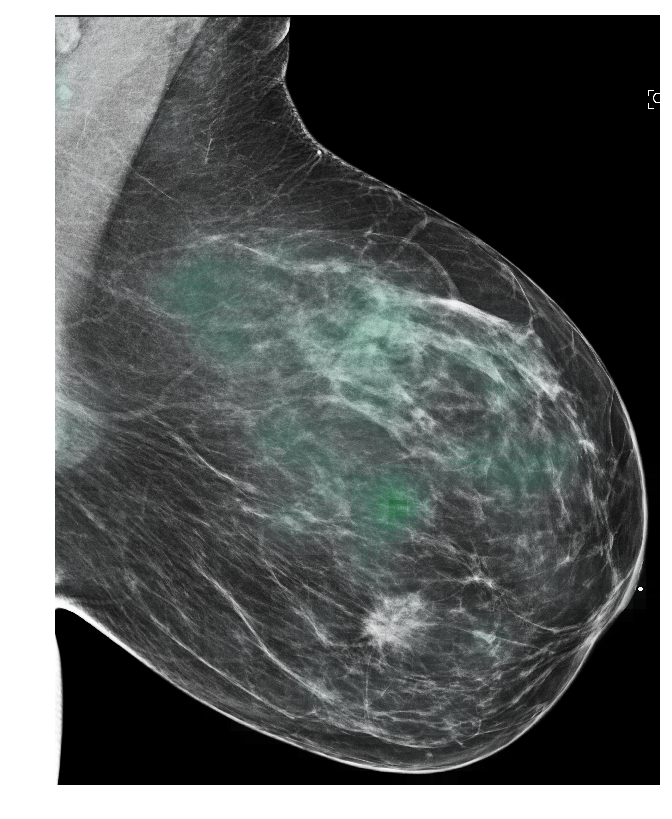}& \hspace{-5.5mm}\includegraphics[width=0.37\linewidth, height=0.51\linewidth]{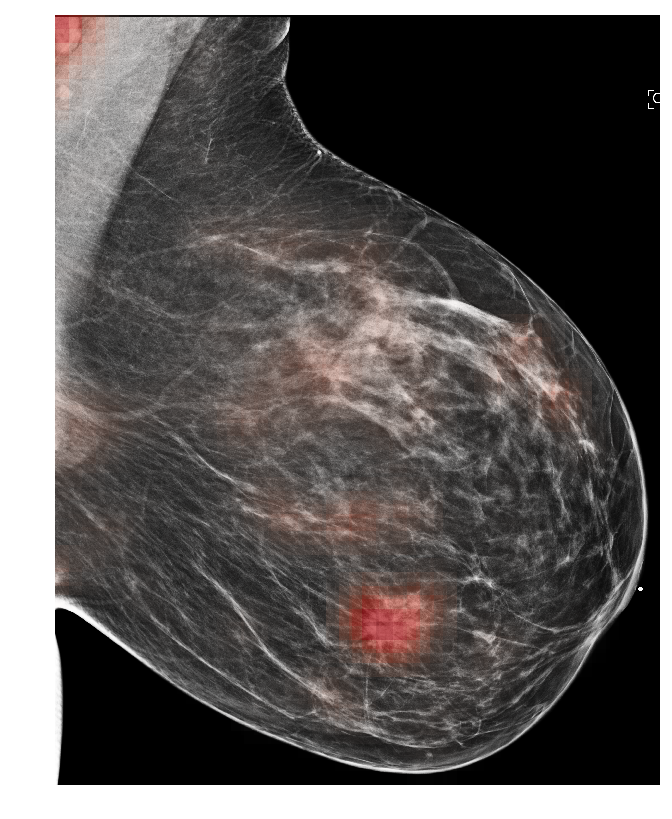} 
    \end{tabular}
    \vspace{-4mm}
    \caption{A biopsy-proven malignant finding marked in red and a benign finding marked in green on both the CC view and the MLO view in the patient's right breast. Images and heatmaps are shown with the same layout as in \autoref{fig:error_analysis_case_1}. The malignant score given by the model is 0.702 and the benign score is 0.682. Mean malignant score given by readers is 0.978 and ten of them gave a score over 0.9. 
    }
    \label{fig:error_analysis_case_8}
    \end{minipage}
    \hspace{3mm}
    \begin{minipage}[t]{.31\textwidth}
    \centering
    \begin{tabular}{c c c}
         \hspace{-5.5mm}\includegraphics[width=0.37\linewidth, height=0.51\linewidth]{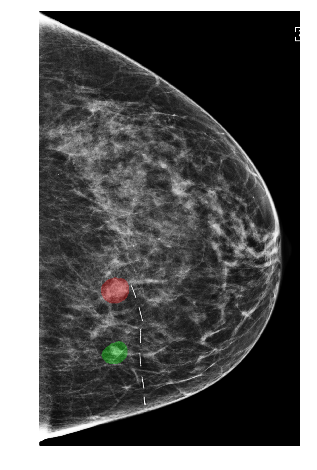} &  \hspace{-6.0mm}\includegraphics[width=0.37\linewidth, height=0.51\linewidth]{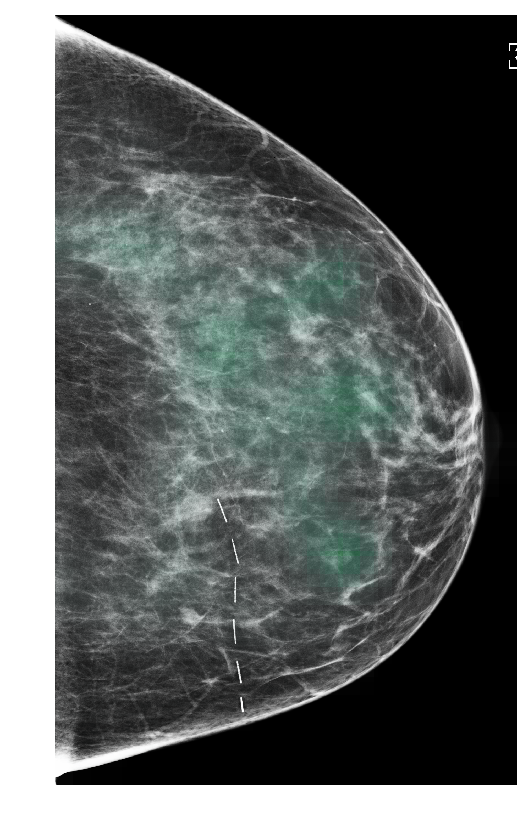}& \hspace{-5.5mm}\includegraphics[width=0.37\linewidth, height=0.51\linewidth]{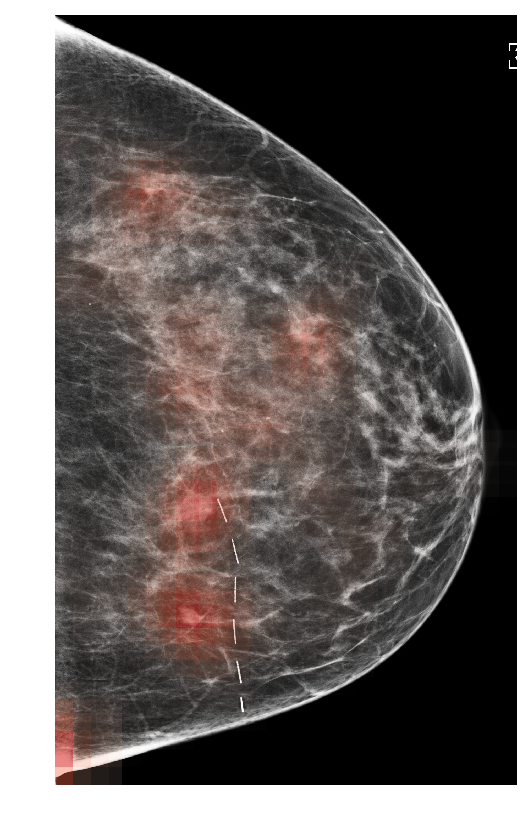}  \\
         \hspace{-5.5mm}\includegraphics[width=0.37\linewidth, height=0.51\linewidth]{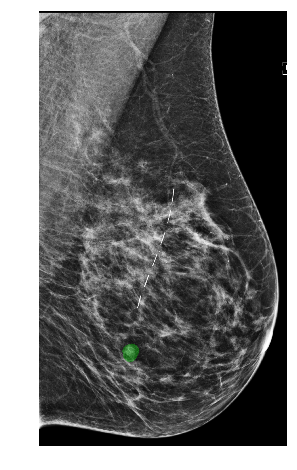} & \hspace{-6.0mm}\includegraphics[width=0.37\linewidth, height=0.51\linewidth]{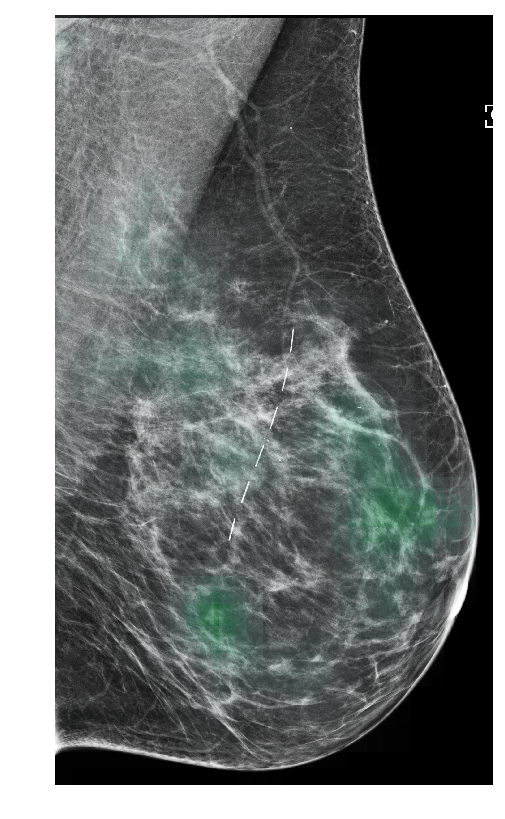} & \hspace{-5.5mm}\includegraphics[width=0.37\linewidth, height=0.51\linewidth]{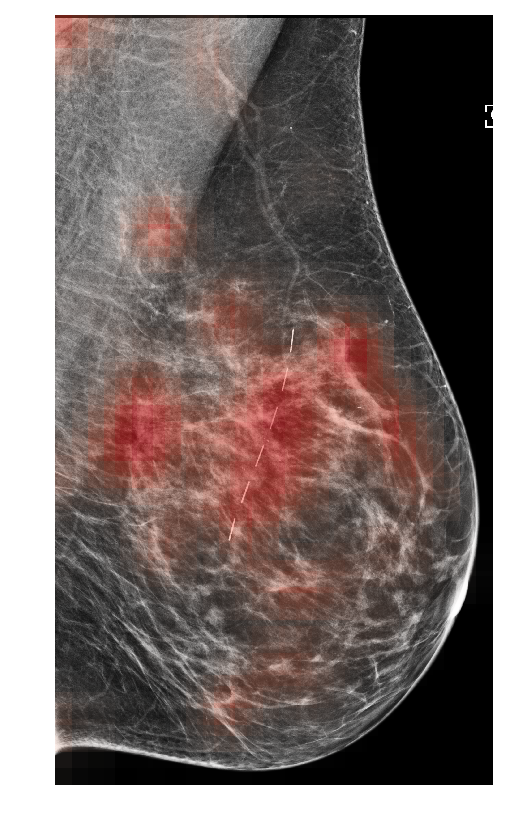} 
    \end{tabular}
    \vspace{-4mm}
    \caption{A biopsy-proven benign finding marked in green on both CC view (top row) and MLO view (the second row) and a malignant finding marked in red only on CC view, in the patient's right breast. Images and heatmaps are shown with the same layout as in \autoref{fig:error_analysis_case_1}. The malignant score for this breast given by the model is only 0.590 while the benign score is 0.557. The readers' mean malignant score is 0.071, with the highest score being only 0.3. }
    \label{fig:error_analysis_case_9}
    \end{minipage}
\end{figure}

To further understand the performance of our image-and-heatmaps model ensemble (referred to as \textit{the model} for the remainder of this section), especially its medical relevance, we conducted the following detailed analysis on nine breasts comprised of examples which were classified incorrectly but confidently by the model. These includes false positive cases for malignancy (cf. \autoref{fig:error_analysis_case_1}, \autoref{fig:error_analysis_case_2}, \autoref{fig:error_analysis_case_3}) and false negative cases for malignancy (cf. \autoref{fig:error_analysis_case_4}, \autoref{fig:error_analysis_case_5}, \autoref{fig:error_analysis_case_6}), as well as examples where the model strongly disagreed with the average predictions of the 14 readers (cf. \autoref{fig:error_analysis_case_7}, \autoref{fig:error_analysis_case_8}, \autoref{fig:error_analysis_case_9}). Examples are shown with annotated lesions, both heatmaps, and a brief summary of the model's and readers' predictions.

The case in \autoref{fig:error_analysis_case_1} succinctly illustrates the ambiguity in medical imaging. Both the model and readers were highly confident in predicting cancer, but the result of the biopsy was a high-risk benign finding.
Further evidence for ambiguity is found in \autoref{fig:error_analysis_case_5}, where both the model and readers predicted that the calcifications were benign. Although the findings appear relatively benign on screening mammography, radiologists often recommend a biopsy of low suspicion calcifications (>2-10\% chance of malignancy) due to the known wide variation in the appearance of malignant calcifications and the opportunity to identify an early, more treatable cancer.
Above all, some of the model's false negative and false positive cases can be explained as evidence for the inherent ambiguity in imaging, which highlights that screening mammography may not be sufficient to determine the correct diagnosis for certain findings.

In \autoref{fig:error_analysis_case_9}, the readers' scores for malignancy are consistently low while the score given by our model is 0.590. In fact, the small mass marked in green on the image turned out to be benign, while on the diagnostic mammogram, the area marked in red looked more suspicious and turned out to be a cancer. This case illustrates the strength of our model--when multiple suspicious findings are present and some are more obvious and easier to determine, human readers may be fatigued from reading a series of mammograms and could be more prone to error by not fully considering each suspicious finding. 
\autoref{fig:error_analysis_case_3} is another good example from this perspective when a finding presents differently on two views. Upon a retrospective review by a radiologist, the smaller faint calcifications on the MLO view indeed appear suspicious, whereas readers may have focused on the CC view that look benign during the reader study. 

Our model still lacks the ability to summarize information about changes from multiple images and views. Some cases with incorrect predictions could be summarized from this perspective, such as \autoref{fig:error_analysis_case_2} and \autoref{fig:error_analysis_case_6}. In \autoref{fig:error_analysis_case_2}, while the radiologist confidently thought that the case would be benign, since even the distribution of the calcification was particularly suspicious on one image, this did not hold up on additional images. In \autoref{fig:error_analysis_case_6}, the model may have missed the malignant finding because it only appeared highly suspicious on the MLO view, but experienced doctors still caught it with high confidence. However, in \autoref{fig:error_analysis_case_7}, our model shows its potential in utilizing both global and local information. According to radiologists, the case looks very suspicious, especially on MLO view, where there is a white mass with irregular margins (termed
architectural distortion). However, the pathology was benign. The model indeed provided a low score for a malignant finding and a high score for a benign finding. While the `malignant' heatmap appeared more correlated with the area under the yellow mask for both views, the `benign' heatmap was widely distributed with high magnitude.      

\autoref{fig:error_analysis_case_4} is an example of a false negative for the model. Although we obverse an overlap between the region highlighted by the the `malignant' heatmap and the lesion, the model's prediction for malignant findings is low while the its prediction for benign findings is higher than 0.5. However, there is an asymmetry with architectural distortion on both views--an imaging feature that has a high probability of malignancy.  Hence, radiologists assigned a high malignancy score. 

Another case where readers were more confident than the model is \autoref{fig:error_analysis_case_8}. The mass in the right breast appears suspicious because it has an architectural distortion. In addition, its location at the bottom from the MLO view (inferior breast), and its medial location from the CC view makes it highly suspicious. In this scenario, radiologists incorporated information from the global view of the mammogram in making their assessment about the likelihood of malignancy for this case. 

\section*{Network architecture and training}

We detail in this section the model architecture, training procedure and hyperparameters associated with training our deep convolutional neural network for cancer classification.

    \subsection*{Breast-level cancer classification model}
    We use a single classification model to generate predictions for each of the four labels of an exam, corresponding to the presence of findings in either breast (left-benign, left-malignant, right-benign, right-malignant).
    The model takes as input a set of four single-channel images, corresponding to the four standard mammographic views (R-CC, L-CC, R-MLO, L-MLO). We use an input resolution of $2677\times1942$ pixels for CC views, and $2974\times1748$ pixels for MLO views, based on the optimal window procedure described in \cite{NYU_dataset}. 
    When additionally using the heatmaps produced by the auxiliary network learning from patch-level labels, we concatenate them as extra input channels to the corresponding views, resulting in three channels in total: the image, the `benign' patch classification heatmap, and the `malignant' patch classification heatmap. 
    
    The model is composed of four view-specific columns, each based on the ResNet architecture \cite{resnet} that computes a fixed-dimension hidden representation for each view. Weights are shared between the L-CC and R-CC columns, and L-MLO and R-MLO columns regardless of model variant. The output of the model is four separate binary probability estimates--one for each of the four labels.

    We initialized the weights of the view-specific columns by pretraining with BI-RADS labels (see section below), and randomly initialized the rest. We trained the whole model using stochastic gradient descent with the Adam optimization algorithm \cite{adam}, using a learning rate of $10^{-5}$ and a minibatch of size $4$. Our loss function was cross-entropy averaged across all four labels. We applied L2-regularization to our model weights with a coefficient of $10^{-4.5}$. 
    As only a small fraction of the exams in our training set contained images of biopsied breasts, learning with all data in the training set would be extremely slow as the model would only be shown a relatively small number of positive examples per epoch. To alleviate this issue, we adopted the following two strategies. 
    First, while we trained the cancer classification model on data from all screening exams, within each training epoch, the model was shown all exams with biopsies in the training set (4,844 exams) but only a random subset of an equal number of exams without biopsies (also 4,844 exams) \cite{imbalanced_survey}. Secondly, as mentioned above, we initialized the ResNet weights of the cancer classification model from a model trained on BI-RADS classification, a task for which we have labels for all exams.
    
\begin{figure}[h]
    \centering
    \begin{minipage}{\textwidth}
        \centering
        \includegraphics[width=1.0\linewidth]{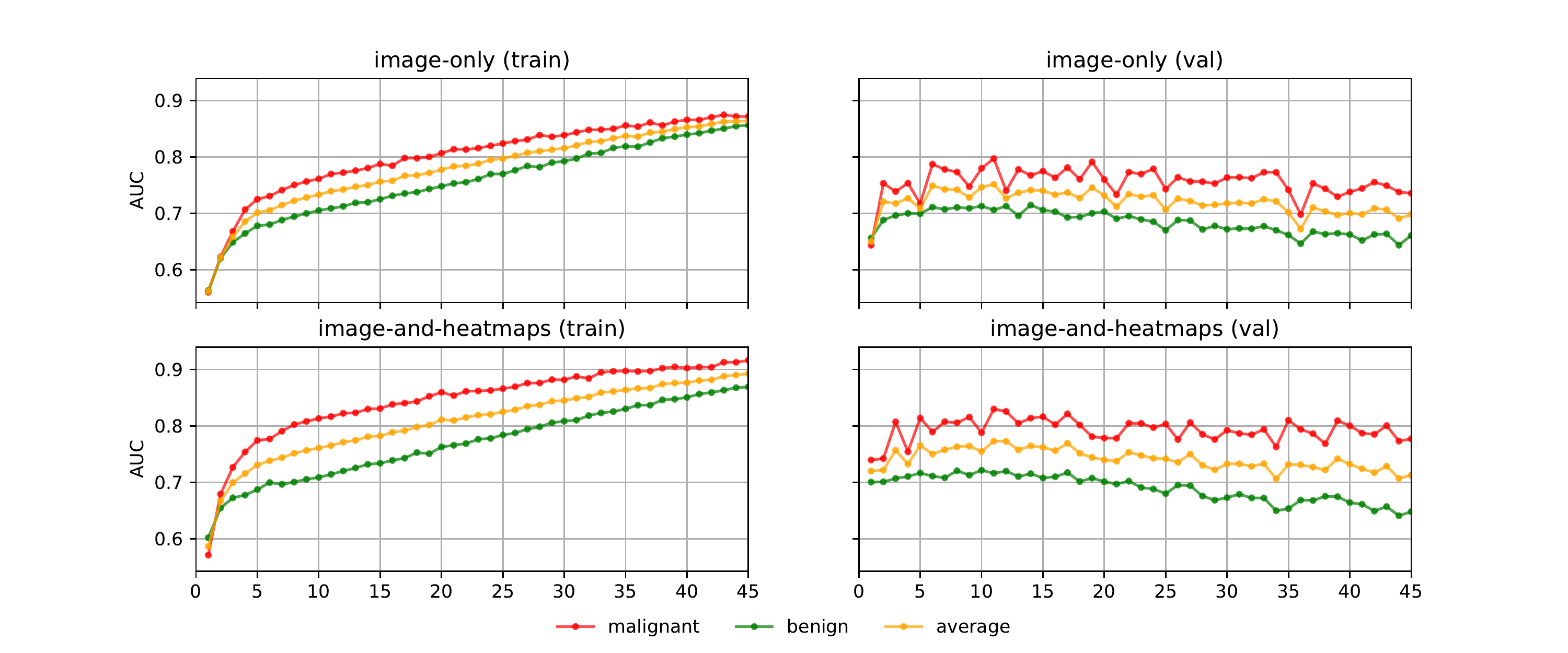}
        \caption{Training curves for image-only and image-and-heatmaps models. AUCs are averaged across prediction heads and target labels. Training AUCs are computed on subsampled data with an equal number of biopsied and randomly subsampled non-biopsied examples, while validation AUCs are computed on the full validation set.}
        \label{fig:training_curves}
    \end{minipage}
    \end{figure}

    We early-stopped the training when the average of the validation AUCs over the four labels
    computed on the validation set did not improve for 20 epochs. We then selected the version of the model with the best validation AUC as our final model candidate. We show the training and validation curve for one image-only model and one image-and-heatmaps model in \autoref{fig:training_curves}. 
    For the training curve, we computed the AUC of each prediction 
    and corresponding label (e.g. left breast/CC/benign) and average across the breast sides and CC/MLO branches.
    The AUC is computed on a training data subsample that has an equal number of biopsied and non-biopsied examples. We do the same for the validation curve, except we compute the AUC on the full validation data set. Because of the difference in distributions, the training and validation AUC curves are not directly comparable--we refer the reader to the discussion in the main paper on how differences in the proportion of biopsied examples can significantly influence AUC calculations.
    We observe that the image-and-heatmap model attains higher training and validation AUC for malignancy prediction compared to the image-only model, whereas the AUCs for benign prediction are not significantly different between the image-and-heatmaps and image-only models.
    
    The full image-only model has 6,132,592 trainable parameters, while the image-and-heatmaps model has 6,135,728 trainable parameters. The only difference between both architectures is the size of the kernel in the first convolutional layer to accommodate the difference in the number of input channels. On an Nvidia V100 GPU, an image-only model takes about 12 hours to train to the best validation performance, while an image-and-heatmaps model takes about 24 hours. A significant amount of training overhead is associated with the time to load and augment the high resolution mammography images.

\section*{Data augmentation for model training}
    
\begin{figure}[h]
        \centering
        \begin{tabular}{c c c c}
        \includegraphics[height=0.3\linewidth]{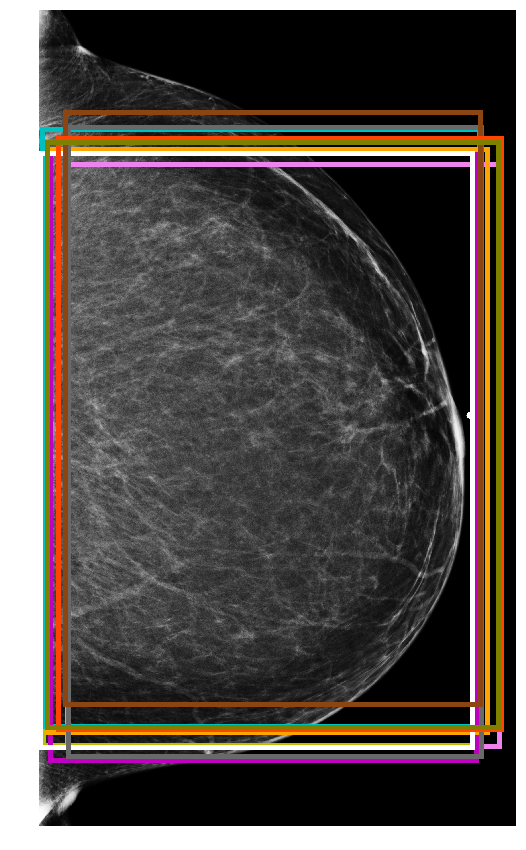} &
        \includegraphics[height=0.3\linewidth]{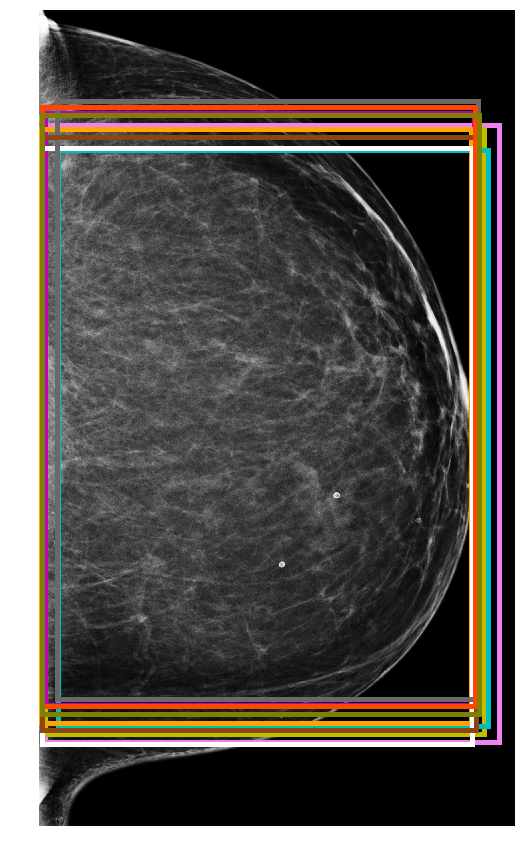} &
        \includegraphics[height=0.3\linewidth]{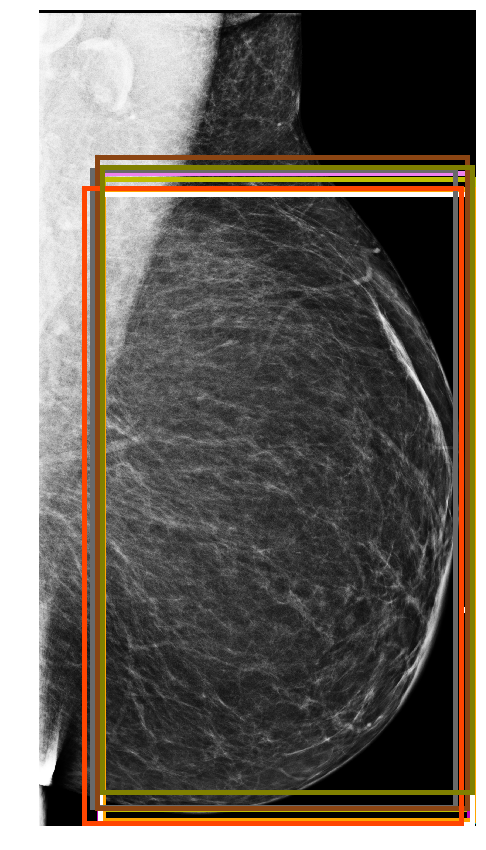} &
        \includegraphics[height=0.3\linewidth]{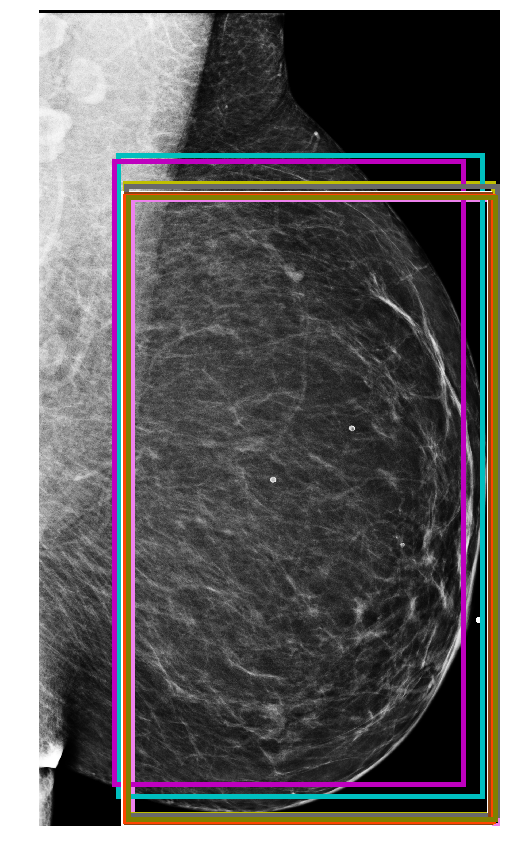}
        \end{tabular}
        \caption{
        Example of drawing 10 augmentation windows with random noise in the location and size of the windows. 
             }
        \label{fig:augmentation_cropping_noise}
    \end{figure}

    Data augmentation is often applied in training deep neural networks to increase the diversity of the training data samples to improve the robustness of the trained model. We apply size augmentation (slightly modifying the crop window size, and resizing using bicubic interpolation to fit the desired size for the model) and location augmentation (adding noise around the chosen optimal center of the window). Examples can be found in \autoref{fig:augmentation_cropping_noise}. We limited the maximum value for both size and location augmentation to 100 pixels in any direction. If the image was too small to apply augmentation, we additionally pad the images to allow enough room. 
    
    At test time, we similarly apply data augmentation, and average predictions over 10 random augmentations to compute the prediction for a given sample. No data augmentation is used during validation.
    
\subsection*{Model variants}
    \begin{figure*}[h!]
        \centering
        \begin{tabular}{c c c c}
        \includegraphics[width=0.23\linewidth,trim={2.3cm 2.4cm 2.7cm 3.8cm}, clip]{figures/model_2.pdf}&
        \includegraphics[width=0.23\linewidth,trim={2.3cm 2.4cm 2.7cm 3.8cm}, clip]{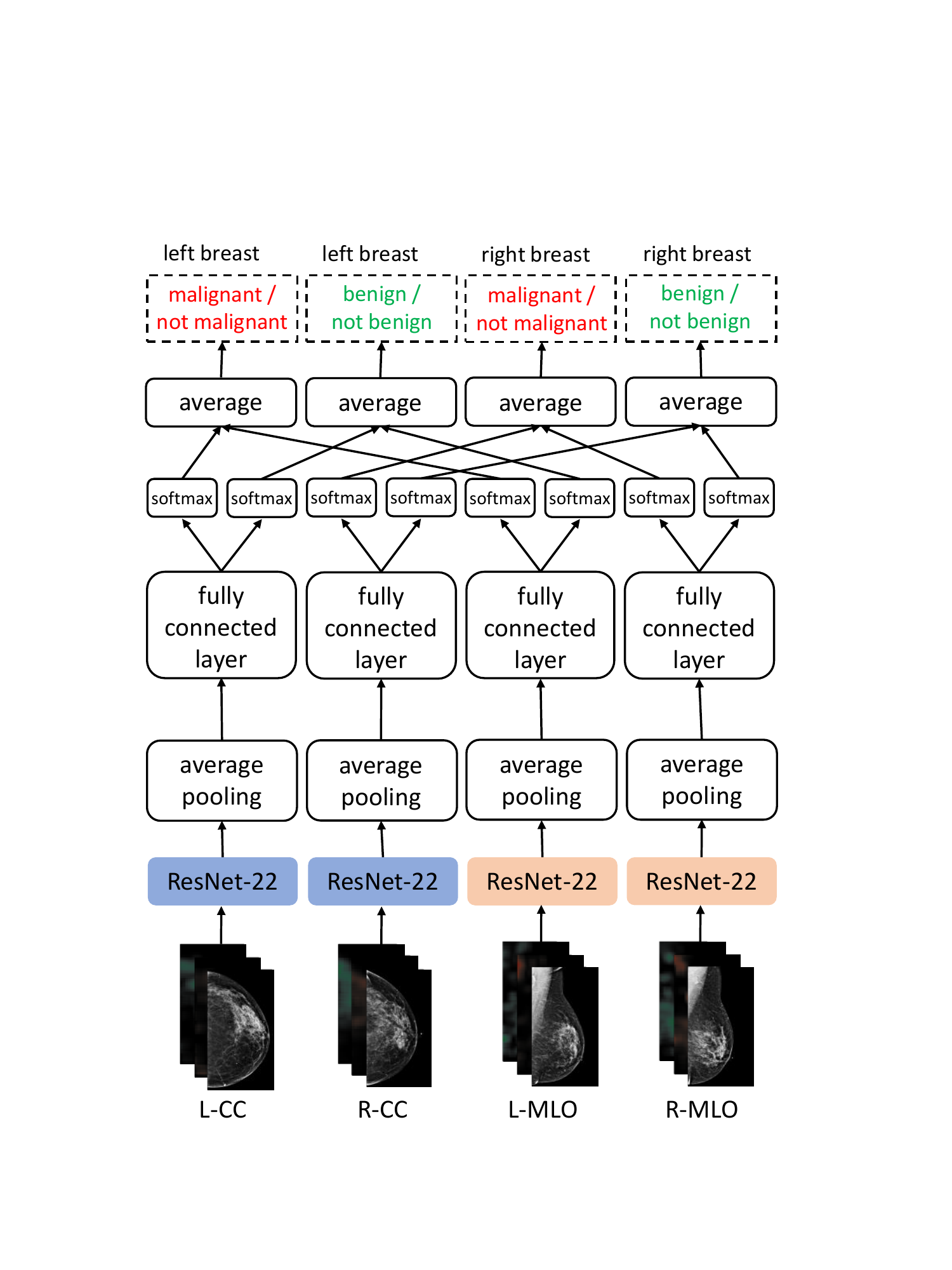}&
        \includegraphics[width=0.23\linewidth,trim={2.3cm 2.4cm 2.7cm 3.8cm}, clip]{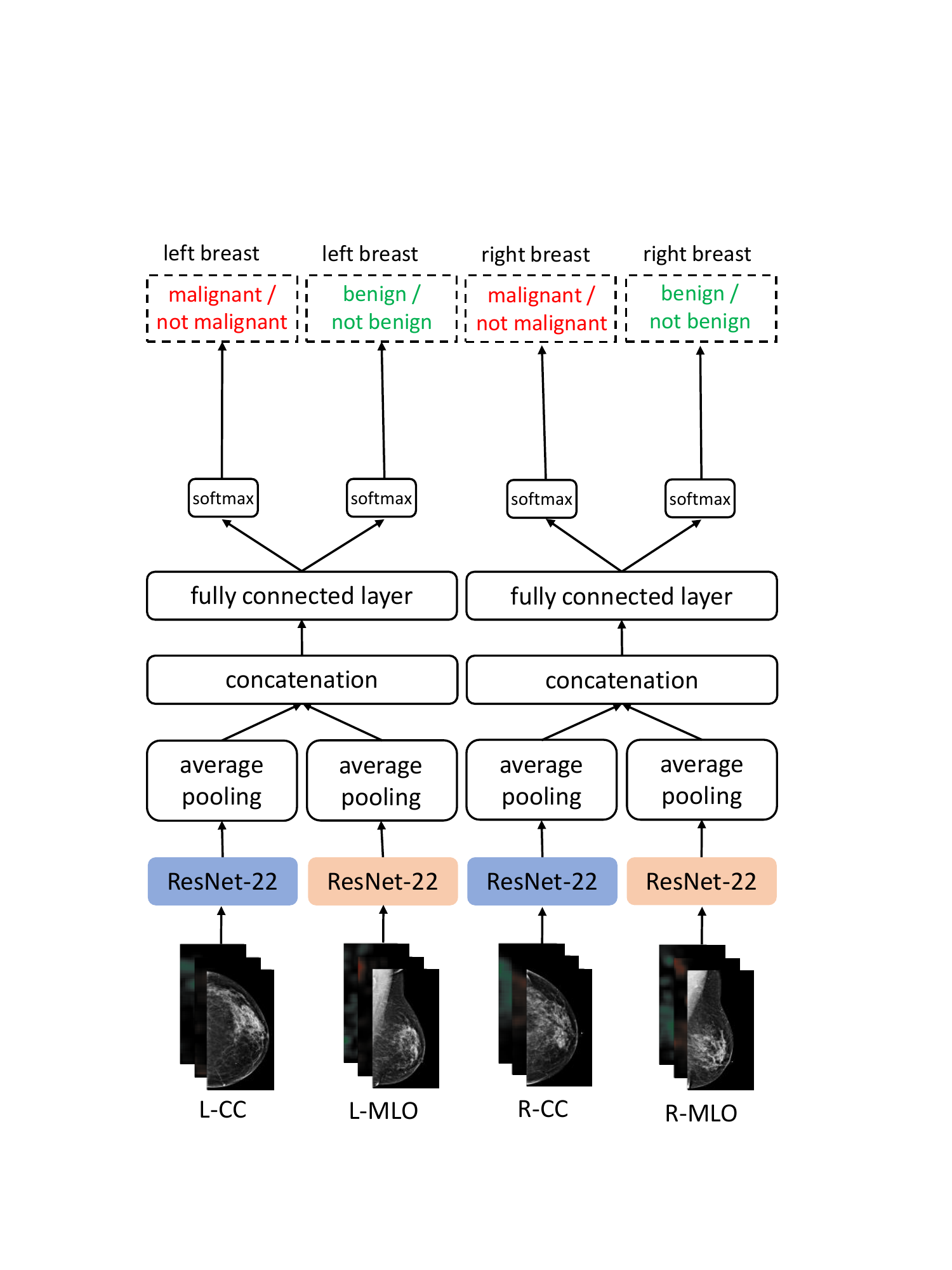}&
        \includegraphics[width=0.23\linewidth,trim={2.3cm 2.4cm 2.7cm 3.8cm}, clip]{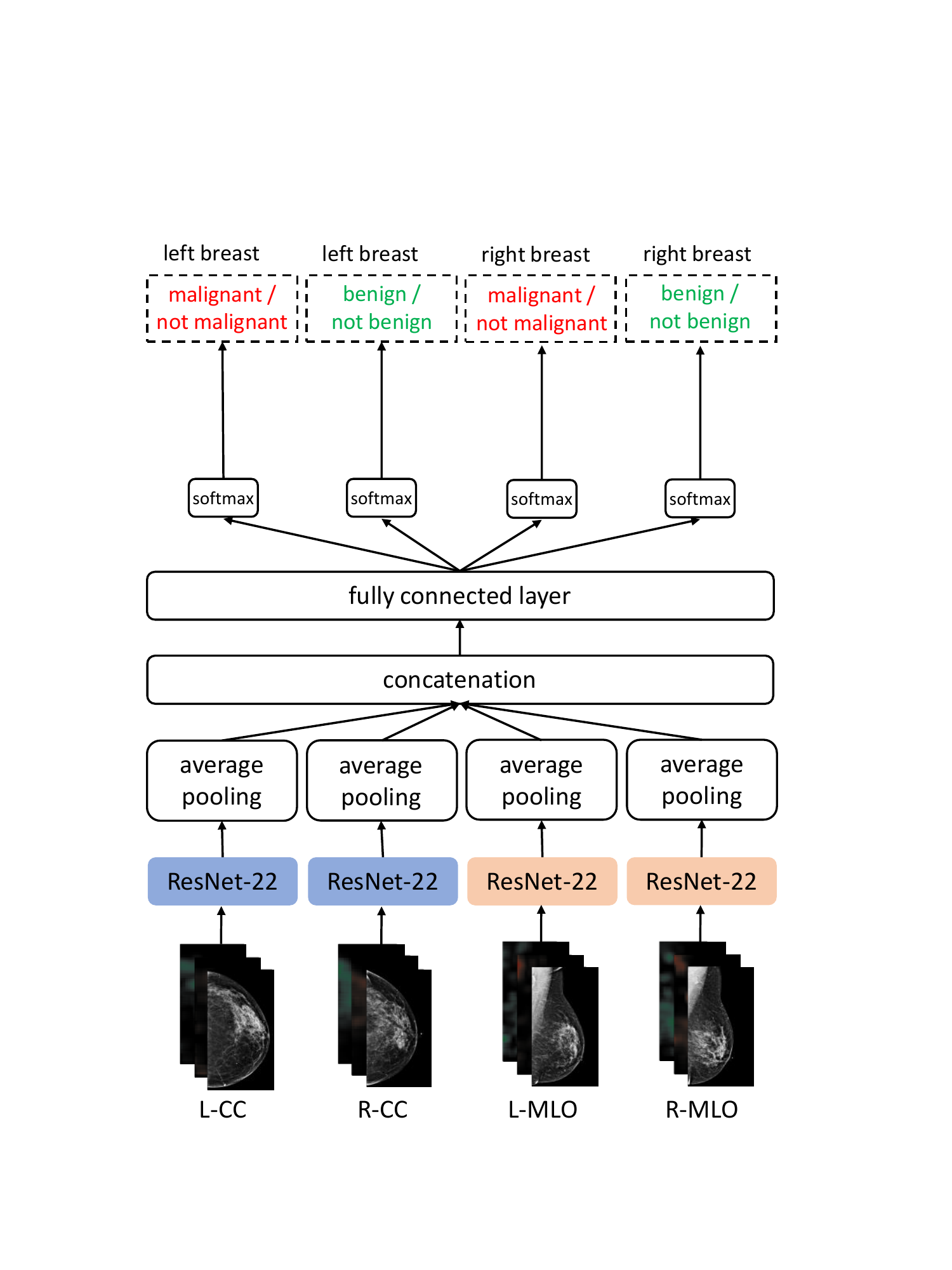}
        \\
        \footnotesize{(a) view-wise} & \footnotesize{(b) image-wise} & \footnotesize{(c) breast-wise} & \footnotesize{(d) joint}
        \end{tabular}
        \vspace{-2mm}
        \caption{
            Four model variants for incorporating information across the four screening mammography views in an exam. All variants are constrained to have a total of 1,024 hidden activations between fully connected layers. 
            The `view-wise' model, which is the primary model used in our experiments, contains separate model branches for CC and MLO views--we average the predictions across both branches. The `image-wise' model has a model branch for each image, and we similarly average the predictions. The `breast-wise' model has separate branches per breast (left and right). The `joint' model only has a single branch, operating on the concatenated representations of all four images.
            }
        \label{fig:architectures}
    \end{figure*}

    \begin{table}[ht]
        \centering
        \caption{
            AUC of model variants on screening and biopsied populations.
        }
        \begin{tabular}{| l | c | c | c | c |}
        \cline{2-5}
        \multicolumn{1}{c|}{} & \multicolumn{2}{c|}{single} & \multicolumn{2}{c|}{5x ensemble} \\
        \cline{2-5}
        \multicolumn{1}{c|}{} & malignant & benign & malignant & benign \\ 
        \cline{2-5}
        \hline
        \multicolumn{5}{|c|}{\cellcolor{gray!20} {\textbf{screening population}} } \\ \hline
        image-only (view-wise) & 0.827$\pm$0.008 & 0.731$\pm$0.004 & 0.840 & 0.743 \\ \hline
        image-only (image-wise) & 0.830$\pm$0.006 & 0.759$\pm$0.002 & 0.841 & 0.766 \\ \hline
        image-only (breast-wise) & 0.821$\pm$0.012 & 0.757$\pm$0.002 & 0.836 & 0.768 \\ \hline
        image-only (joint) & 0.822$\pm$0.008 & 0.737$\pm$0.004 & 0.831 & 0.746 \\ \hline
        image-and-heatmaps (view-wise) & \textbf{0.886}$\pm$\textbf{0.003} & 0.747$\pm$0.002 & \textbf{0.895} & 0.756 \\ \hline
        image-and-heatmaps (image-wise) & 0.875$\pm$0.001 & \textbf{0.765}$\pm$\textbf{0.003} & 0.885 & 0.774 \\ \hline
        image-and-heatmaps (breast-wise) & 0.876$\pm$0.004 & 0.764$\pm$0.004 & 0.889 & \textbf{0.779} \\ \hline
        image-and-heatmaps (joint) & 0.860$\pm$0.008 & 0.745$\pm$0.002 & 0.876 & 0.763 \\ \hline
        \multicolumn{5}{|c|}{\cellcolor{gray!20} {\textbf{biopsied population}} } \\ \hline
        image-only (view-wise) & 0.781$\pm$0.006 & 0.673$\pm$0.003 & 0.791 & 0.682 \\ \hline
        image-only (image-wise) & 0.740$\pm$0.007 & 0.638$\pm$0.001 & 0.749 & 0.642 \\ \hline
        image-only (breast-wise) & 0.726$\pm$0.009 & 0.639$\pm$0.002 & 0.738 & 0.645 \\ \hline
        image-only (joint) & 0.780$\pm$0.006 & 0.682$\pm$0.001 & 0.787 & 0.688 \\ \hline
        image-and-heatmaps (view-wise) & \textbf{0.843}$\pm$\textbf{0.004} & 0.690$\pm$0.002 & \textbf{0.850} & 0.696 \\ \hline
        image-and-heatmaps (image-wise) & 0.812$\pm$0.001 & 0.653$\pm$0.003 & 0.821 & 0.658 \\ \hline
        image-and-heatmaps (breast-wise) & 0.805$\pm$0.004 & 0.652$\pm$0.004 & 0.818 & 0.661 \\ \hline
        image-and-heatmaps (joint) & 0.817$\pm$0.008 & \textbf{0.696}$\pm$\textbf{0.005} & 0.830 & \textbf{0.709} \\ \hline
        \end{tabular}
        \label{tab:cancer_pred_variant}
    \end{table}

    Based on the four view-specific hidden representations, we considered four model variants for incorporating the information from all four views in producing our output predictions. The full architectures of the four variants are shown in \autoref{fig:architectures}. The `view-wise' model concatenates L-CC and R-CC representations, and L-MLO and R-MLO representations, and uses separate CC and MLO prediction heads to generate predictions for all four labels. This is the model used in the main paper, chosen based on validation performance on the screening population. The `image-wise' model has separate prediction heads for each of the four views, predicting only the malignant or benign labels for the corresponding breast. The `side-wise' model concatenates L-CC and L-MLO representations, and R-CC and R-MLO representations, and has separate prediction heads for each breast. Lastly, the `joint' model concatenates the representations of all four views and jointly predicts malignant and benign findings for both breasts. Regardless of architecture, each model consists of two fully connected layers that produce four probability estimates--one for each of the four labels.

    We show results across different model variants in \autoref{tab:cancer_pred_variant}, evaluated on the screening population. Overall, all four model variants achieve high and relatively similar AUCs. The `view-wise' image-and-heatmaps ensemble, which is also architecturally most similar to the BI-RADS model used in the pretraining stage, performs the best in predicting malignant/not malignant, attaining an AUC of 0.895 on the screening population and 0.850 on the biopsied population. However, some of the other model variants do outperform the `view-wise' ensemble for benign/not-benign prediction. Among the image-only models, the four model variants perform roughly comparably, though still consistently underperforming the image-and-heatmaps models. We emphasize that the `view-wise' model was chosen as the model shown in the main paper based on the average of malignant/not malignant and benign/not benign AUCs on the validation set, and not based on test set results.
    
    Constructing an ensemble of the four model variants for the image-and-heatmaps model, with five randomly initialized models per variant,\footnote{Only the weights in the fully connected layers are randomly initialized--we use the same set of pretrained BI-RADS weights to initialize ResNet columns in all experiments, excluding the experiments with models without BI-RADS pretraining.} results in an AUC of 0.778 on benign/not benign prediction, and 0.899 on malignant/not malignant prediction on the screening population. Although this performance is superior to any individual model variant, running such a large ensemble of 20 separate models would be prohibitively expensive in practice.

\subsection*{Single-view ResNet}
    
\begin{figure}[h]
    \centering
    \begin{minipage}{\textwidth}
        \centering
        \includegraphics[width=0.8\linewidth]{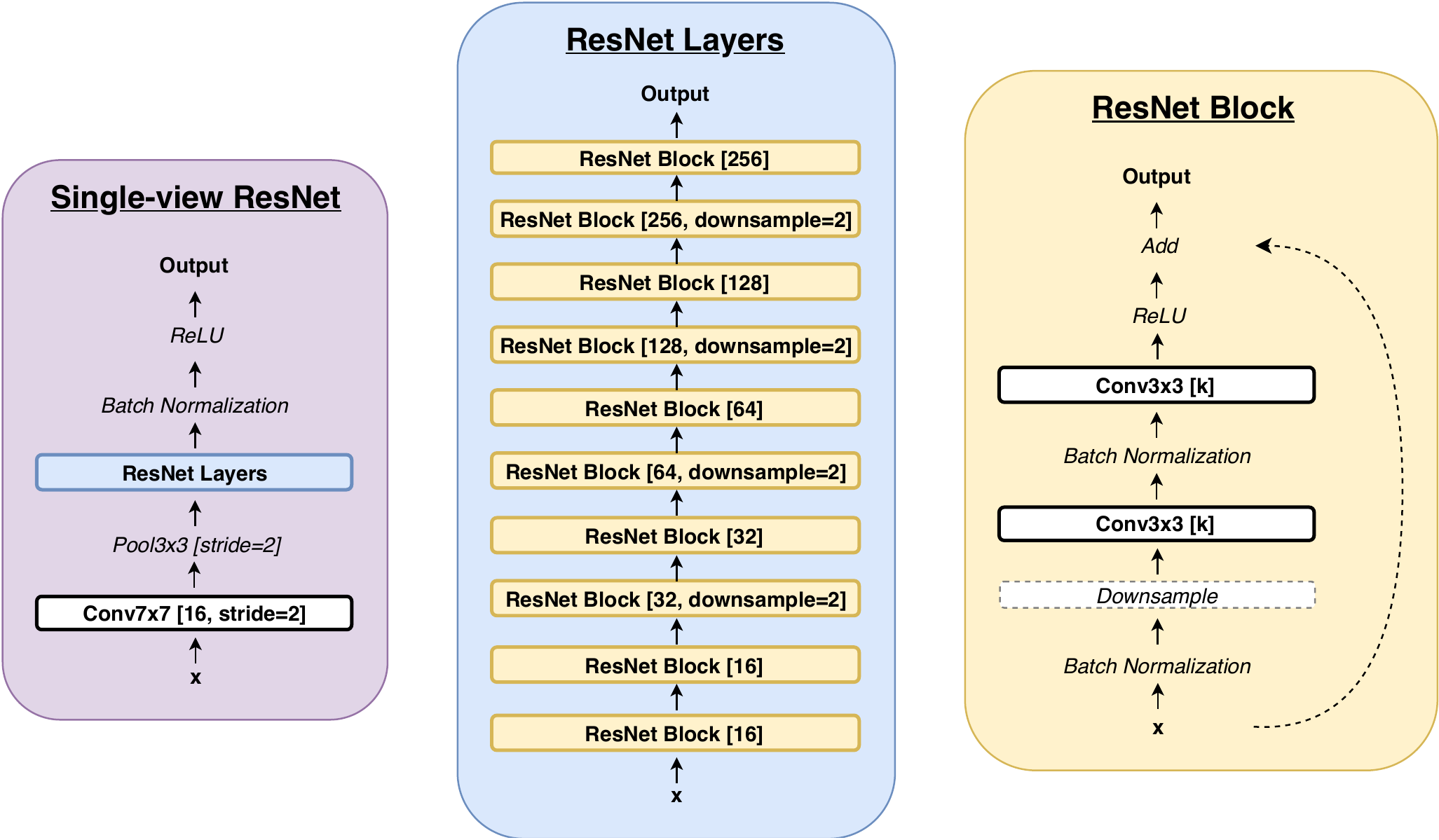}
        \caption{Architecture of single-view ResNet. The numbers in square brackets indicate the number of output channels, unless otherwise specified. Where no downsampling factor is specified for a ResNet block, the downsampling layer reduces to a 1x1 convolution. \textbf{Left}: Overview of the single-view ResNet, which consists of a set of ResNet layers. \textbf{Center}: ResNet layers consist of a sequence of ResNet blocks with different downsampling and output channels. \textbf{Right}: ResNet blocks consist of two 3x3 convolutional layers, with interleaving downsampling and batch normalization operations, and a residual connection between input and output.}
        \label{fig:single_view_resnet}
    \end{minipage}
    \end{figure}

    The overall model consists of four separate ResNet \cite{resnet} models corresponding to each of the four views. In this section, we describe the structure of these ResNets. The full architecture of each ResNet is shown in \autoref{fig:single_view_resnet}. We tied the weights for the L-CC and R-CC ResNets, as well as the L-MLO and R-MLO ResNets. Likewise, we flipped the L-CC and L-MLO images before feeding them to the model, so all breast images are rightward-oriented, allowing the shared ResNet weights to operate on similarly oriented images.

    The output of each ResNet is a $H\times W \times 256$-dimensional tensor where $H$ and $W$ are downsampled from the original input size, with $H$=42 and $W$=31 for the CC view, and $H$=47 and $W$=28 for MLO view. We average-pool across the spatial dimensions to obtain a 256-dimension hidden representation vector for each view. For reference, we show the dimensions of the hidden activations after each major layer of the ResNet in \autoref{tab:resnet_dimensions}.
    
    \begin{table}[ht]
    \centering
    \caption{
        Dimensions hidden of activation after each layer in the ResNet, shown as $D \times H\times W$.
    }
    \begin{tabular}{| c | c | c| }
        \cline{2-3}
        \multicolumn{1}{c|}{} & \textbf{CC view} & \textbf{MLO view} \\ \hline
        Conv7x7 & 1339$\times$971$\times$16 & 1487$\times$874$\times$16 \\ \hline
        ResBlock 0 & 670$\times$486$\times$16 & 744$\times$437$\times$16 \\ \hline
        ResBlock 1 & 335$\times$243$\times$32 & 372$\times$219$\times$32 \\ \hline
        ResBlock 2 & 168$\times$122$\times$64 & 186$\times$110$\times$64 \\ \hline
        ResBlock 3 & 84$\times$61$\times$128 & 93$\times$55$\times$128 \\ \hline
        ResBlock 4 & 42$\times$31$\times$256 & 47$\times$28$\times$256 \\ \hline
    \end{tabular}
    \label{tab:resnet_dimensions}
\end{table}

\subsection*{Pretraining on BI-RADS classification}
    
    Because of the small number of labeled biopsied examples we have available, we apply transfer learning to improve the robustness and performance of our models. Transfer learning involves reusing parts of a model pretrained on another task as a starting point for training the target model, taking advantage of the learned representations from the pretraining task. 
    
    For our model, we apply transfer learning from a network pretrained on a BI-RADS classification task, as in \cite{high_resolution}, which corresponds to predicting a radiologist's assessment of a patient's risk of developing breast cancer based on screening mammography. 
    The three BI-RADS classes we consider are: BI-RADS Category 0 (``incomplete''), BI-RADS Category 1 (``normal'') and BI-RADS Category 2 (``benign''). The algorithm used to extract these labels is explained in \cite{NYU_dataset}. Although these labels are potentially much noisier than biopsy outcomes (being assessments of clinicians based on screening mammograms and not informed by a biopsy), compared to the 4,844 exams with biopsy-proven cancer labels in the training set, we have over 99,528 training examples with BI-RADS 0 and BI-RADS 2 labels. As shown in \cite{6909618}, a few thousand training exams may be insufficient to learn millions of parameters in CNN architectures--instead, convolutional layers can be pretrained as a ``generic extractor of mid-level image representation" and thereafter reused. On the other hand, although the BI-RADS labels are noisy, neural networks can reach reasonable levels of performance even when trained with noisy labels, as shown in \cite{DBLP:journals/corr/KrauseSHZTDPL15} and \cite{DBLP:journals/corr/SunSSG17}, 
    and the information learned can then be transferred to the cancer classification model. In fact, our experiments show that pretraining on BI-RADS classification contributes significantly to the performance of our model.

    \begin{figure}[h]
        \centering
        \begin{tabular}{c}
        \includegraphics[width=0.4\linewidth,trim={2.35cm 2.5cm 2.7cm 3.1cm}, clip]{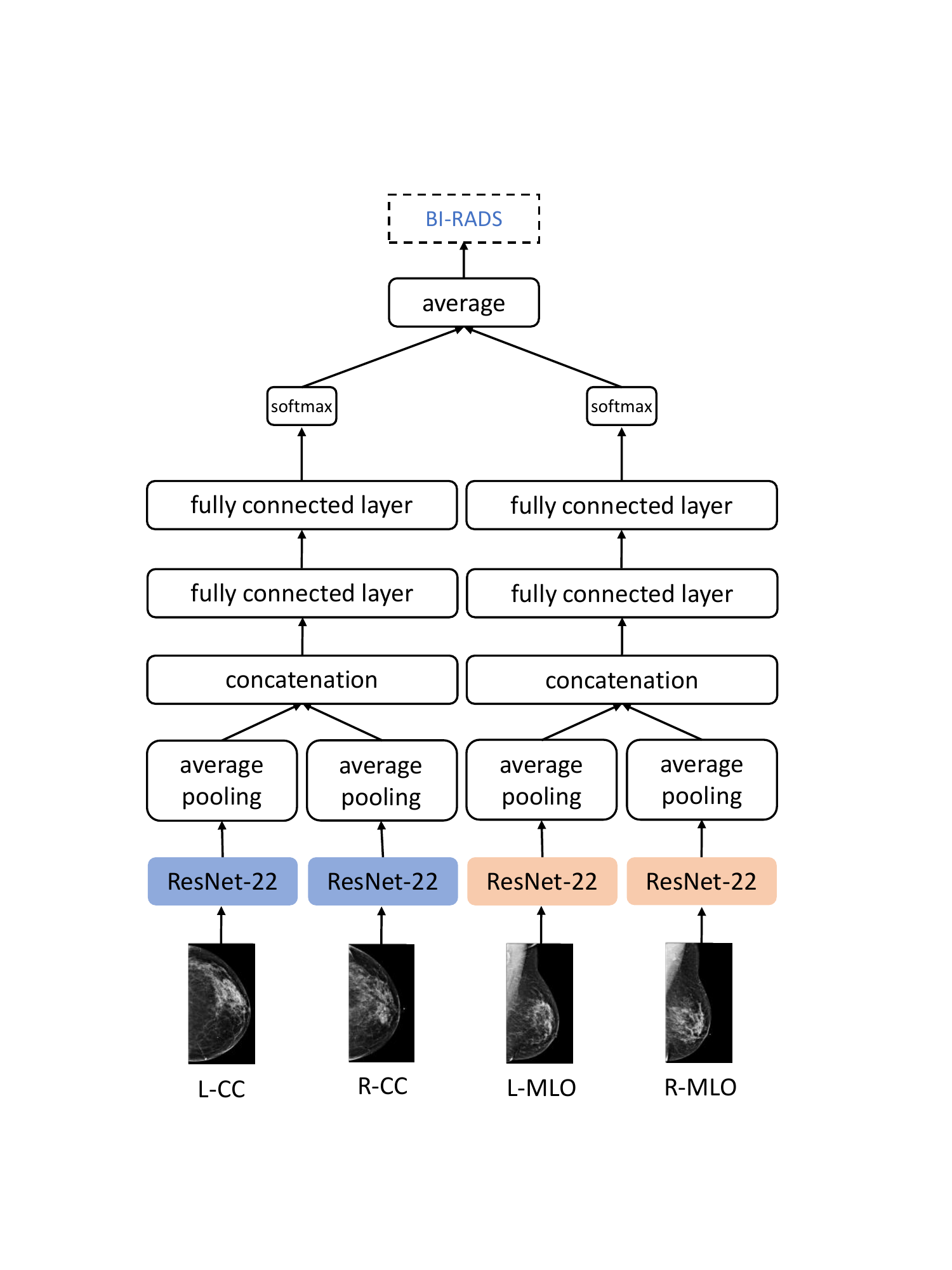}
        \end{tabular}
        \caption{BI-RADS classification model architecture. The architecture is largely similar to the `view-wise' cancer classification model variant, except that the output is a set of probability estimates over the three output classes. The model consists of four ResNet columns, with weights shared within CC and MLO branches of the model.}
        \label{fig:birads_model_figure}
    \end{figure}

    The model we use for BI-RADS classification is shown in \autoref{fig:birads_model_figure}. It is largely similar to the `view-wise' model architecture for cancer classification described in the \textit{Model variants} section above, except that the output layer outputs probability estimates over three classes for a single label. Although the BI-RADS classification task is a three-class classification task, we measured the performance of the model by averaging AUCs of 0-vs-other, 1-vs-other and 2-vs-other predictions on the validation set.
    
    The rest of the training details (e.g. ResNet architecture, optimizer hyperparameters) are identical to those of the cancer classification model, except that the model was trained with a minibatch size of 24 instead of 4. We early-stopped training based on validation AUCs after no improvements for 20 epochs, and initialized the ResNet weights for the cancer classification model using the learned weights in the BI-RADS model. Where we used heatmaps as additional input channels, we duplicated the weights on the bottommost convolutional kernel such that the model can operate on inputs with three channels--the rest of the model is left unchanged. In our experimental results, we used a BI-RADS model trained for 111 epochs (326 hours or approximately 14 days on four Nvidia V100 GPUs), which obtained an averaged validation AUC of 0.748.
    
    We emphasize here that we used the same train-validation-test splits for pretraining our BI-RADS classification model as in training our cancer classification model, so no data leakage across splits was possible.

\subsubsection*{Cancer classification model without BI-RADS pretraining}
    
    In this section, we evaluate the benefit of the BI-RADS pretraining by comparing the performance of our models to cancer classification models trained without using weights from a pretrained BI-RADS model. Specifically, we train a set of cancer classification models by starting from entirely randomly initialized model weights.
    
    The results are shown in \autoref{tab:cancer_pred_no_pretraining}. In every case, we see an improvement in performance from using weights of a model pretrained on BI-RAD classification, compared to randomly initializing the model weights and training from scratch. The improvement in performance from using pretrained weights tends to be larger for the image-only model compared to image-and-heatmaps models. We hypothesize that this is because the heatmaps already contain significant information pertaining to cancer classification, and hence the model can likely more quickly learn to make use of the heatmaps for cancer classification. In contrast, the image-only models rely entirely on the ResNets to effectively encode visual information for cancer classification, and therefore using the weights of a model pretrained for BI-RADS classification contributes significantly to the model performance.
   
    \begin{table}[ht]
        \centering
        \caption{
            AUCs of our models on screening and biopsied populations, with and without BI-RADS pretraining.
        }
        \begin{tabular}{| l | c | c | c | c |}
        \cline{2-5}
        \multicolumn{1}{c|}{} & \multicolumn{2}{c|}{single} & \multicolumn{2}{c|}{5x ensemble} \\
        \cline{2-5}
        \multicolumn{1}{c|}{} & malignant & benign & malignant & benign \\ 
        \cline{2-5}
        \hline
        \multicolumn{5}{|c|}{\cellcolor{gray!20} {\textbf{screening population}} } \\ \hline
        image-only (pretrained) & 0.827$\pm$0.008 & 0.731$\pm$0.004 & 0.840 & 0.743 \\ \hline
        image-only (random) & 0.687$\pm$0.009 & 0.657$\pm$0.006 & 0.703 & 0.669 \\ \hline
        image-and-heatmaps (pretrained) & \textbf{0.886}$\pm$\textbf{0.003} & \textbf{0.747}$\pm$\textbf{0.002} & \textbf{0.895} & \textbf{0.756} \\ \hline
        image-and-heatmaps (random) & 0.856$\pm$0.007 & 0.701$\pm$0.004 & 0.868 & 0.708 \\ \hline
        \multicolumn{5}{|c|}{\cellcolor{gray!20} {\textbf{biopsied population}} } \\ \hline
        image-only (pretrained) & 0.781$\pm$0.006 & 0.673$\pm$0.003 & 0.791 & 0.682 \\ \hline
        image-only (random) & 0.693$\pm$0.006 & 0.564$\pm$0.006 & 0.709 & 0.571 \\ \hline
        image-and-heatmaps (pretrained) & \textbf{0.843}$\pm$\textbf{0.004} & \textbf{0.690}$\pm$\textbf{0.002} & \textbf{0.850} & \textbf{0.696} \\ \hline
        image-and-heatmaps (random) & 0.828$\pm$0.008 & 0.633$\pm$0.006 & 0.841 & 0.640 \\ \hline
        \end{tabular}
        \label{tab:cancer_pred_no_pretraining}
    \end{table}

\section*{Details of the auxiliary patch-level classifier}

\begin{figure}[h]
\resizebox{.98\textwidth}{!}{
\begin{tabular}{C{0.5\textwidth}  C{0.5\textwidth}}
    \centering
    \begin{tabular}{c c}
    \includegraphics[width=0.22\textwidth]{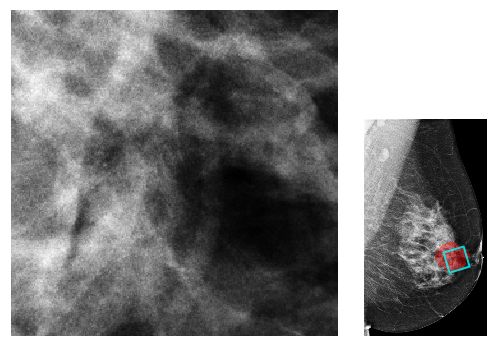} &
    \includegraphics[width=0.22\textwidth]{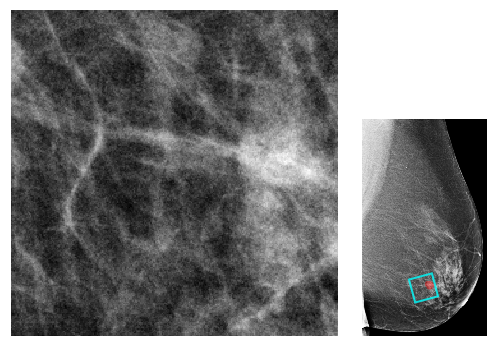} \\
    \includegraphics[width=0.22\textwidth]{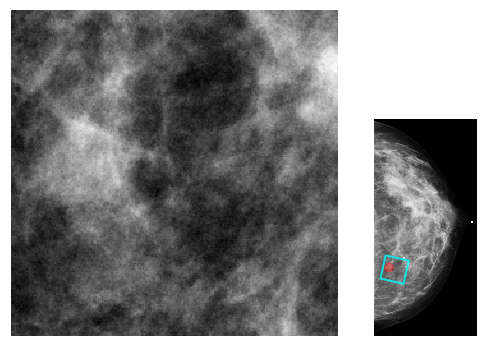} &
    \includegraphics[width=0.22\textwidth]{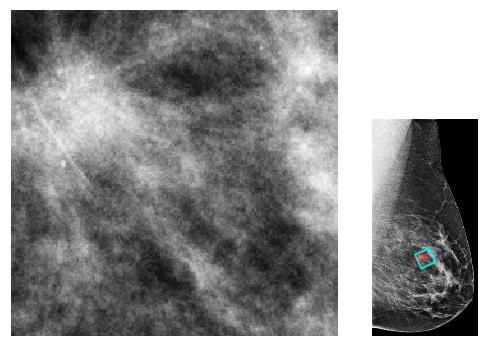}
    \end{tabular}
    &
    \begin{tabular}{c c}
    \includegraphics[width=0.22\textwidth]{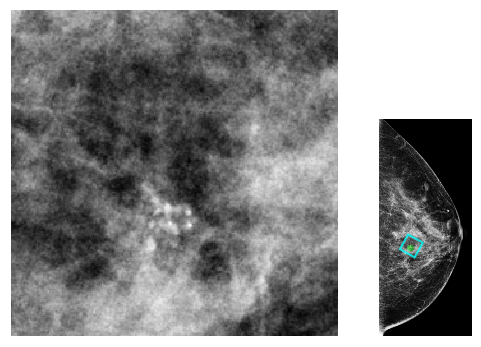} &
    \includegraphics[width=0.22\textwidth]{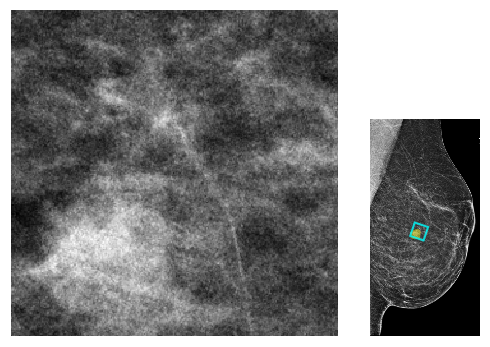} \\
    \includegraphics[width=0.22\textwidth]{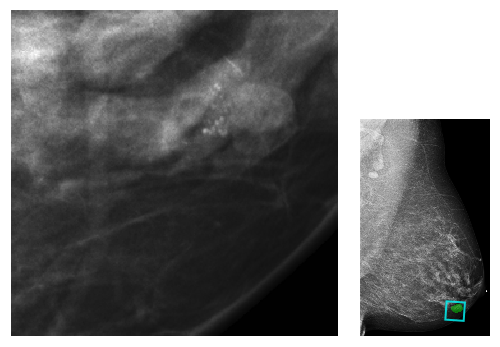} &
    \includegraphics[width=0.22\textwidth]{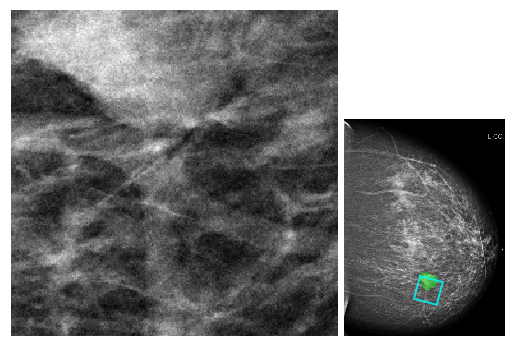}
    \end{tabular}\\
    (a) Malignant. Examples of patches overlapping only with biopsied malignant findings (marked with red). &
    (b) Benign. Examples of patches overlapping only with biopsied benign findings (marked with yellow or green).\\
    \begin{tabular}{c c}
    \includegraphics[width=0.22\textwidth]{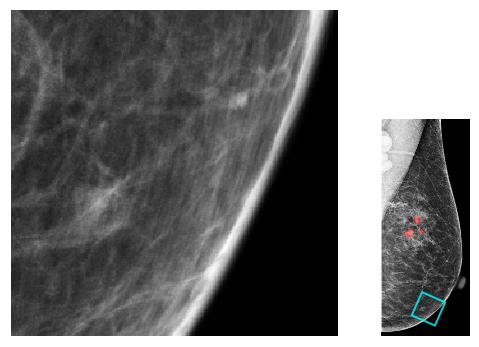} &
    \includegraphics[width=0.22\textwidth]{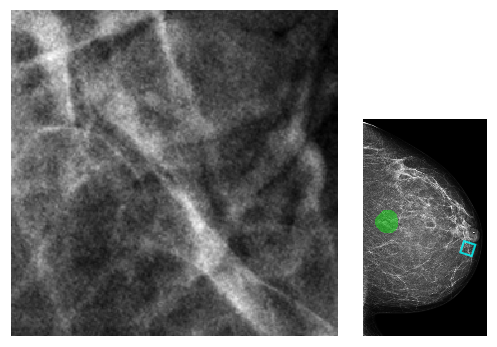} \\
    \includegraphics[width=0.22\textwidth]{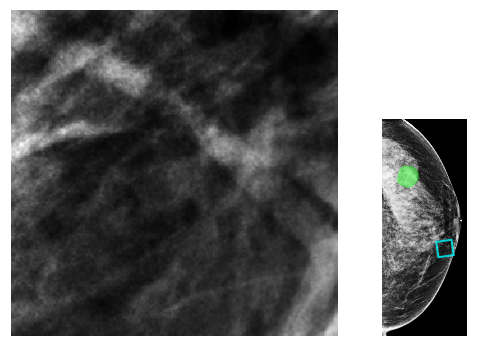} &
    \includegraphics[width=0.22\textwidth]{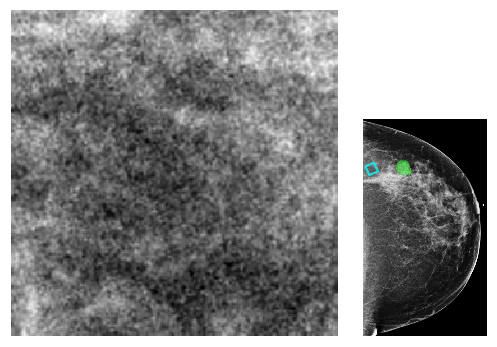}
    \end{tabular} &
    \begin{tabular}{c c}
    \includegraphics[width=0.22\textwidth]{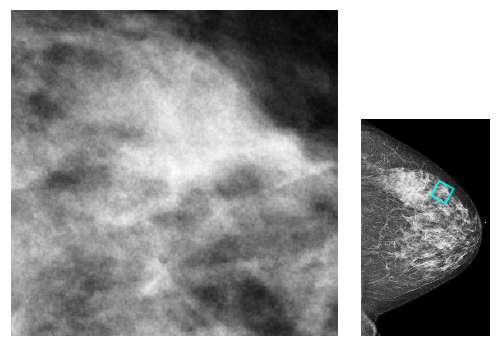} &
    \includegraphics[width=0.22\textwidth]{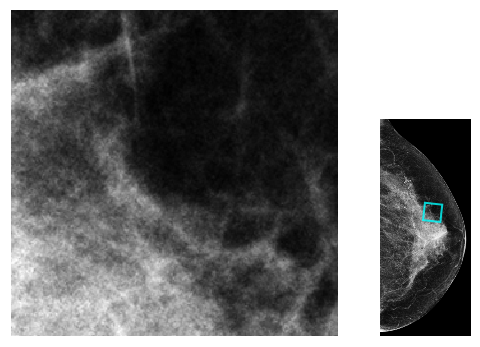} \\
    \includegraphics[width=0.22\textwidth]{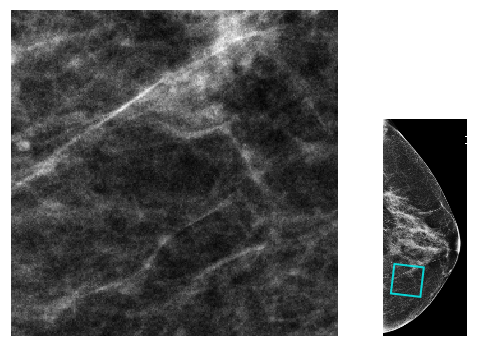} &
    \includegraphics[width=0.22\textwidth]{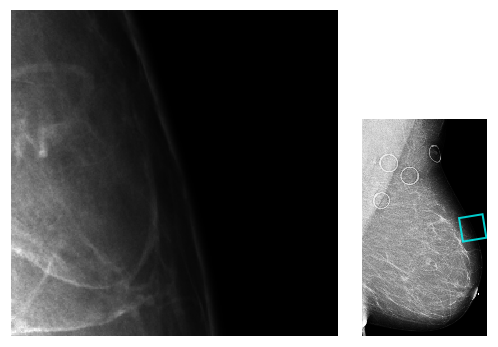}
    \end{tabular}\\
    (c) Outside. Examples of patches from images with biopsied findings but without an overlap with any biopsied findings. & (d) Negative. Examples of patches from images without any biopsied findings.
\end{tabular}
}
    \caption{Examples of patches sampled according to the procedure described in the \textit{Sampling the patches} section. From (a) to (d), four patches are shown with the images of their origin, for four classes: malignant, benign, outside and negative. 
    Patches are shown on the left, while the images of origin (with indicated biopsied findings if any were present) are on the right. 
    The blue squares indicate the location of the patches in the original images.
    The meaning of the colored regions on the images is described in a greater detail in \cite{NYU_dataset}. }
    \label{fig:example_patches}
\end{figure}

We used a dataset of 5,000,000 patches to train the auxiliary patch-level classification network to classify patches into one of four classes: (i) patches overlapping only with area segmented by annotations in red, indicating malignant findings (malignant); (ii) patches overlapping only with area segmented by annotations in green or yellow, indicating benign findings (benign); (iii) patches from segmented images but not overlapping with any marked area (outside); (iv) patches from images in exams labeled as negative for both benign and malignant (negative). As described in \cite{NYU_dataset}, 
the findings were manually indicated on the images by radiologists on a pixel-level, based on results from pathology. Images which are mammographically occult, i.e., the lesions that were biopsied were not visible on the image, were not taken into consideration while generating this training set. 

\subsubsection*{Sampling the patches} 
Patches in the dataset were generated from all available mammography exams in the training set--the same as those used to train the breast-level model. Before extracting the patches, images were all cropped according to the algorithms described in \cite{NYU_dataset}. As was the case for the breast-level model, we flipped the L-CC and L-MLO images so that all breast images were rightward-oriented. Each patch was cropped as a square from a full-size image. To sample a patch, we first sampled a location for the center of the patch, then sampled its size from a uniform distribution between 128 pixels and 384 pixels and finally sampled an angle by which the crop window was rotated, also from a uniform distribution from -30 to 30 degrees. A sample was rejected if it contained any pixels outside of the full-size image or only contained zero-valued pixels (i.e. containing only background and no breast tissue). Once extracted, the patches were resized to 256$\times$256 pixels. Examples are shown in \autoref{fig:example_patches}.

\subsubsection*{Training and architecture}
We used a DenseNet-121 architecture \cite{densenet} for our patch-level auxiliary classifier, with four dense blocks with 6, 12, 24, 16 dense layers respectively. The entire network has approximately seven million parameters. 
We initialized the weights of the model with the weights of a DenseNet-121 trained on ImageNet.

The number of images with visible biopsied findings is small (0.85\%) in comparison to the total number of images. Furthermore, the fraction of the total image area associated with visible biopsied findings is also small (0.87\%, averaging above images with segmentation). To accommodate this, in each training epoch, we randomly sampled 10,000 patches: 20 from the malignant class, 35 from the benign class, 5,000 from the outside class and 4,945 from the negative class. This ratio of malignant, benign, outside and negative patches was chosen to reflect the ratio of $\mathbf{area_{m}}$, $\mathbf{area_{b}}$, $\mathbf{area_{o}}$ and $\mathbf{area_{n}}$, which are the sums of the respective fractions of total areas over our segmented training data set. $\mathbf{area_{m}}$ and $\mathbf{area_{b}}$ denote the total sum of the area under biopsied malignant and benign findings respectively over the entire $6,758$ images with segmentation in the training set. 
Accordingly, the sum of the remaining area is denoted by $\mathbf{area_{o}}$. 7,000 images without any segmentation were randomly sampled and $\mathbf{area_{n}}$ denotes the sum of the size of all those images. In order to address the extreme class imbalance, we used weighted cross-entropy as the training loss, wherein the class weights were set as the inverse of the above patch ratio so that losses on incorrect predictions of malignant and benign patches were appropriately upweighted. Weighted cross-entropy loss has the following form:

\begin{equation*}
\mathcal{L}(\mathbf{x})= \sum_{c} \mathbf{w}_{c} \log \hat{p}_{c}( \mathbf{x}),
\label{eq:auxiliary_objective}
\end{equation*}
where $\mathbf{x}$ is the input image, $c$ is the class label assumed to be in $\{\text{malignant (m)},  \text{benign (b)},  \text{outside (o)}, \text{negative (n)}\}$ and $\hat{p}_{c}(\mathbf{x})$ is the probability of class $c$ predicted by the network. The coefficient $\mathbf{w}_c$ is computed as
$$\mathbf{w}_c = \frac{\Pi_{k \ne c}N_k }{\sum_{j \in {m, b, o, n}} \Pi_{k\ne j}N_k }, $$
where $N_k$ is the number of patches for class $k$. 

We trained the network using the Adam optimization algorithm \cite{adam}, with a batch size of 100 and a learning rate of $10^{-5}$. 

\subsubsection*{Patch classification heatmap generation}
The patch-level auxiliary classifier is applied to the full resolution images in a sliding window fashion to create two class-specific heatmaps, corresponding to the malignant and benign predictions of the patch-level classifier. Since mammography images vary in sizes (before cropping is applied to use them as an input for the model), to slide the classifier over the image, we used Algorithm~\ref{alg:stride_setting} to compute the appropriate values of the strides for vertical and horizontal dimensions. We applied strides of approximately 70 pixels wide across both dimensions. We used non-augmented $256\times 256$ patches as inputs to the patch classifier. For each patch, we projected the respective predicted class probabilities to the original $256\times 256$ input area of each patch, and we averaged the predicted probabilities for pixels in overlapping patches. Ultimately, we generate two heatmaps for each image--one for prediction of malignancy and one for prediction of benign findings. Both are passed as additional input channels to the breast-level model. The predicted probabilities for outside and negative patch classes are not used here.

\begin{algorithm*}[ht]
\caption{Strides setting}
\begin{algorithmic}[1]
    \Function{strides\_setting}{\texttt{image\_size}}
    \State \texttt{prefixed\_stride} = 70
    \State \texttt{patch\_size} = 256
    \State \texttt{sliding\_steps} = (\texttt{image\_size} - \texttt{patch\_size}) // \texttt{prefixed\_stride}
    \State \texttt{pixel\_remaining} = (\texttt{image\_size} - \texttt{patch\_size}) \% \texttt{prefixed\_stride} 
    \If{\texttt{pixel\_remaining} == 0} 
    \State \texttt{stride\_list} = [\texttt{prefixed\_stride}] * \texttt{sliding\_steps}
    \Else
    \State \texttt{sliding\_steps} += 1
    \State \texttt{pixel\_overlap} = \texttt{prefixed\_stride} - \texttt{pixel\_remaining}
    \State \texttt{stride\_avg} = \texttt{prefixed\_stride} - \texttt{pixel\_overlap} // \texttt{sliding\_steps}
    \State \texttt{stride\_list} = [\texttt{stride\_avg}] * \texttt{sliding\_steps}
    \State randomly choose number of \texttt{pixel\_overlap} \% \texttt{sliding\_steps} items from \texttt{stride\_list} and decrement 1 for each.
    \EndIf
    \Return \texttt{stride\_list}
    \EndFunction

\end{algorithmic}
\label{alg:stride_setting}
\end{algorithm*}

\begin{figure}[htb!]
    \centering
    \begin{tabular}{C{0.15cm} c c c}
    & \begin{tabular}{c c c}
        \hspace{-5mm}image & \hspace{4mm}malignant & \hspace{3.5mm}benign 
    \end{tabular}&\begin{tabular}{c c c}
        \hspace{-3mm}image & \hspace{3.5mm}malignant & \hspace{4mm}benign
    \end{tabular} &\begin{tabular}{c c c}
        \hspace{-3mm}image & \hspace{3.5mm}malignant & \hspace{4mm}benign
    \end{tabular} \\
    
    \hspace{-6mm}\begin{tabular}{c}
    \rotatebox[origin=c]{270}{R-CC}
    \end{tabular}&
    \begin{tabular}{c c c}
    \hspace{-8mm}
    \includegraphics[height = 0.15\textwidth, width = 0.10\textwidth, trim={0mm 0mm 0mm 0mm}]{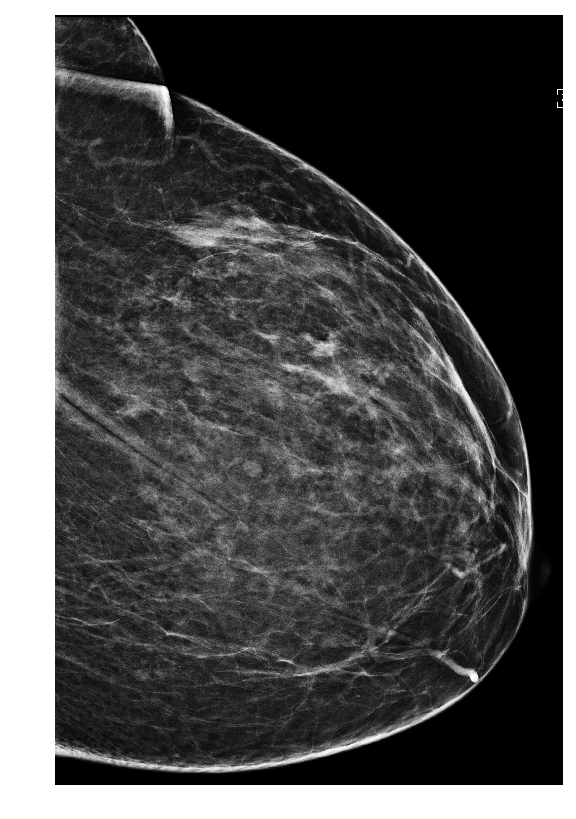}\hspace{-4mm}&
    \hspace{-5mm}\includegraphics[height = 0.15\textwidth, width = 0.10\textwidth, trim={0mm 0mm 0mm 0mm}]{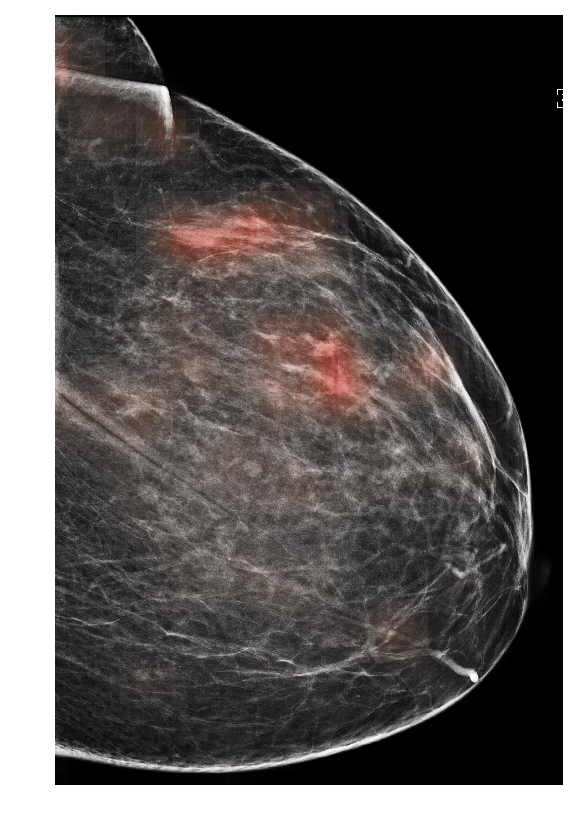}\hspace{-4mm}& \hspace{-5mm}\includegraphics[height = 0.15\textwidth, width = 0.10\textwidth, trim={0mm 0mm 0mm 0mm}]{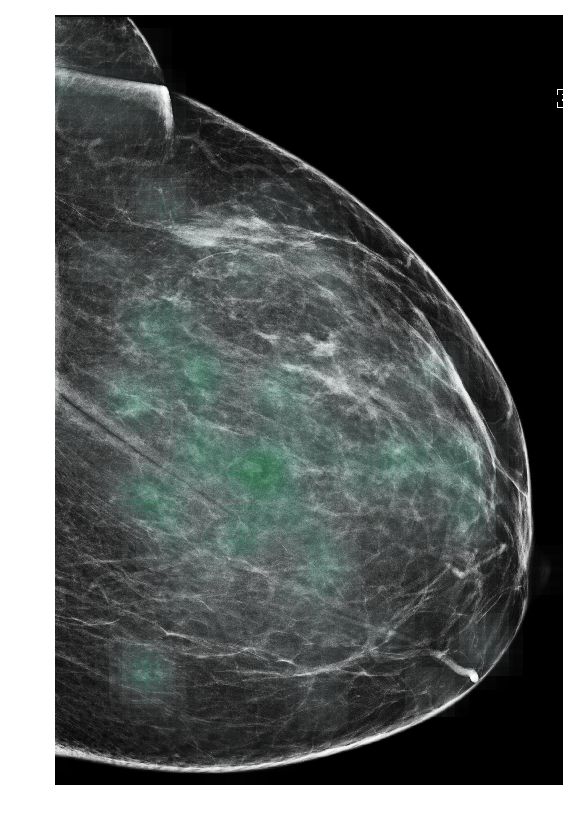}
    \end{tabular} & \begin{tabular}{c c c}
    \hspace{-5mm}
    \includegraphics[height = 0.15\textwidth, width = 0.10\textwidth, trim={0mm 0mm 0mm 0mm}]{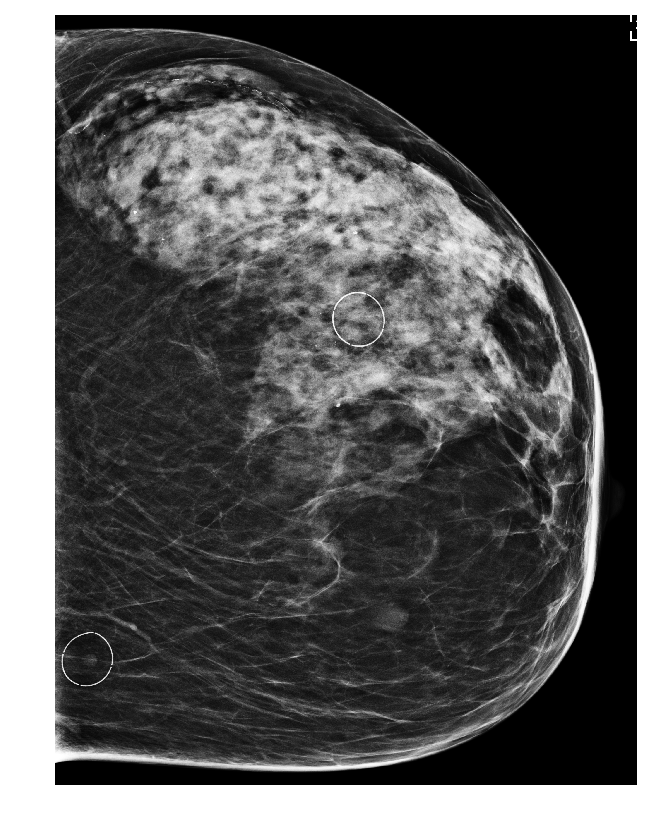}\hspace{-4mm}&
    \hspace{-5mm}\includegraphics[height = 0.15\textwidth, width = 0.10\textwidth, trim={0mm 0mm 0mm 0mm}]{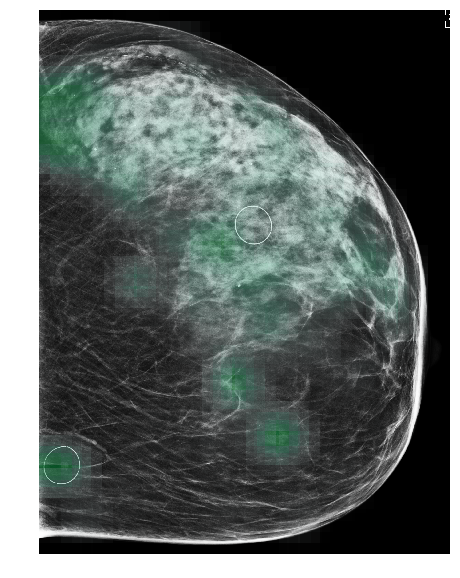}\hspace{-4mm}& \hspace{-5mm}\includegraphics[height = 0.15\textwidth, width = 0.10\textwidth, trim={0mm 0mm 0mm 0mm}]{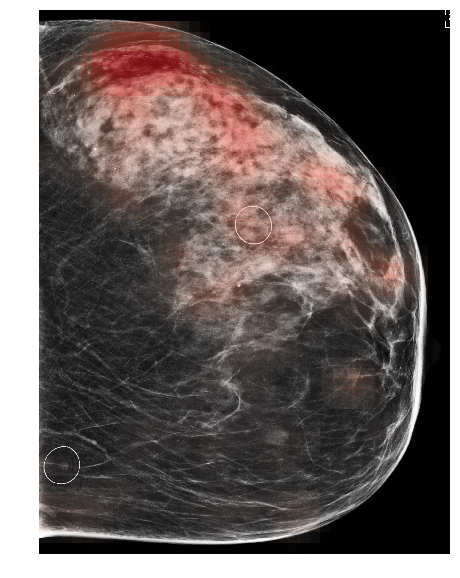}
    \end{tabular}&\begin{tabular}{c c c}
    \hspace{-5mm}
    \includegraphics[height = 0.15\textwidth, width = 0.10\textwidth, trim={0mm 0mm 0mm 0mm}]{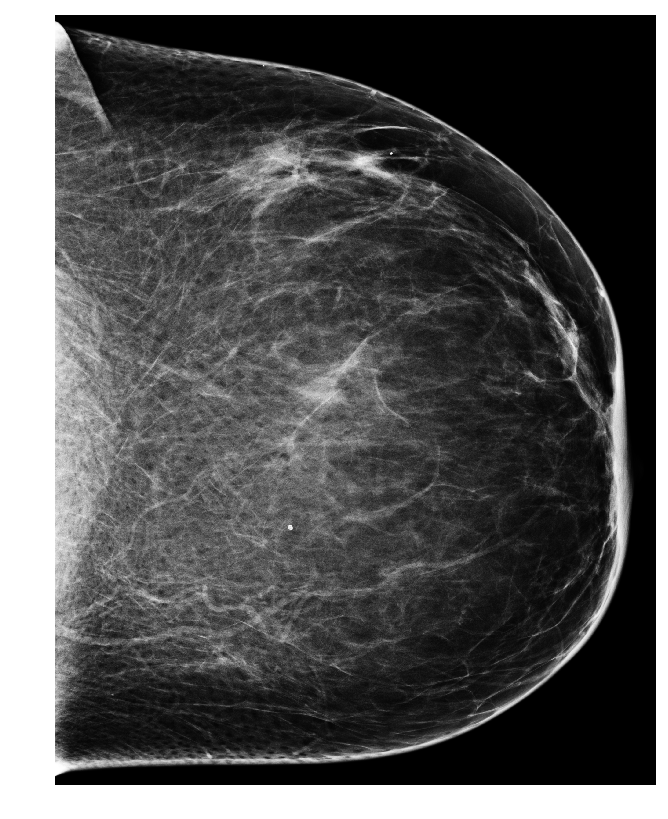}\hspace{-4mm}&
    \hspace{-5mm}\includegraphics[height = 0.15\textwidth, width = 0.10\textwidth, trim={0mm 0mm 0mm 0mm}]{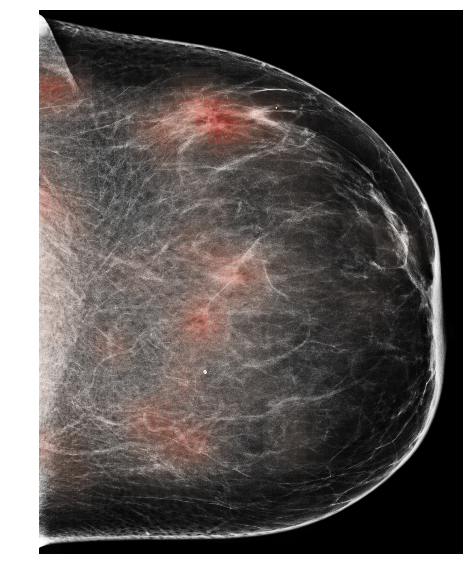}\hspace{-4mm}& \hspace{-5mm}\includegraphics[height = 0.15\textwidth, width = 0.10\textwidth, trim={0mm 0mm 0mm 0mm}]{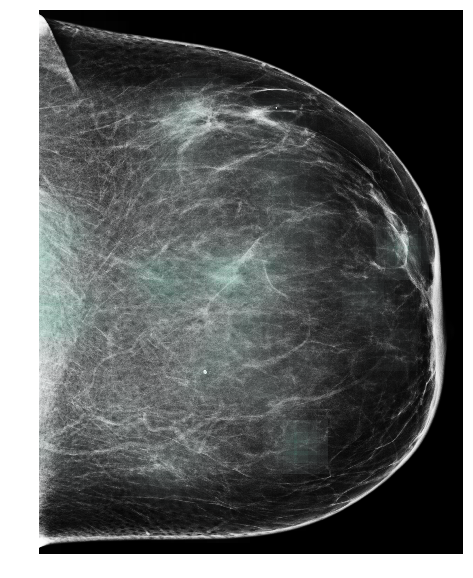}
    \end{tabular}
    \\
    
    \hspace{-6mm}\begin{tabular}{c}
    \rotatebox[origin=c]{270}{L-CC}
    \end{tabular}&
    \begin{tabular}{c c c}
    \hspace{-8mm}
    \includegraphics[height = 0.15\textwidth, width = 0.10\textwidth, trim={0mm 0mm 0mm 0mm}]{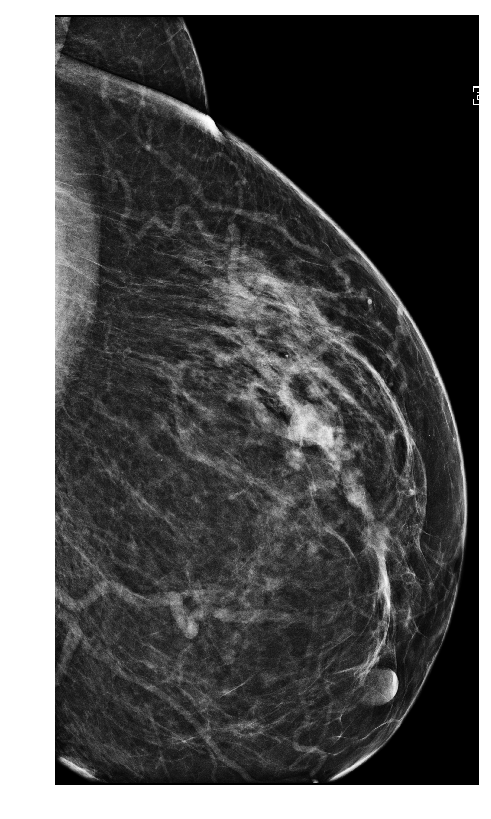}\hspace{-4mm}&
    \hspace{-5mm}\includegraphics[height = 0.15\textwidth, width = 0.10\textwidth, trim={0mm 0mm 0mm 0mm}]{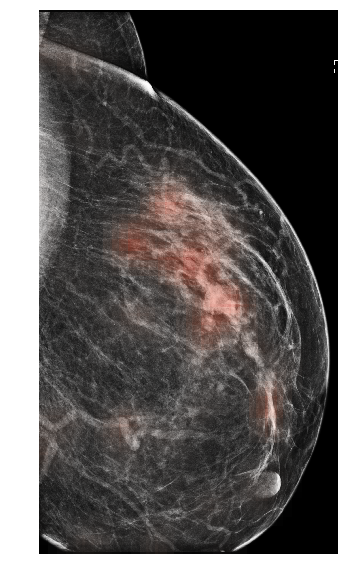}\hspace{-4mm}& \hspace{-5mm}\includegraphics[height = 0.15\textwidth, width = 0.10\textwidth, trim={0mm 0mm 0mm 0mm}]{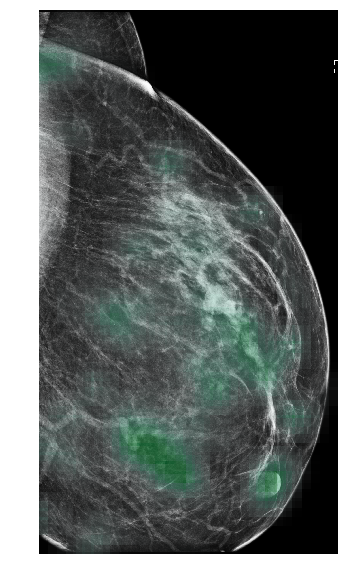}
    \end{tabular} & \begin{tabular}{c c c}
    \hspace{-5mm}
    \includegraphics[height = 0.15\textwidth, width = 0.10\textwidth, trim={0mm 0mm 0mm 0mm}]{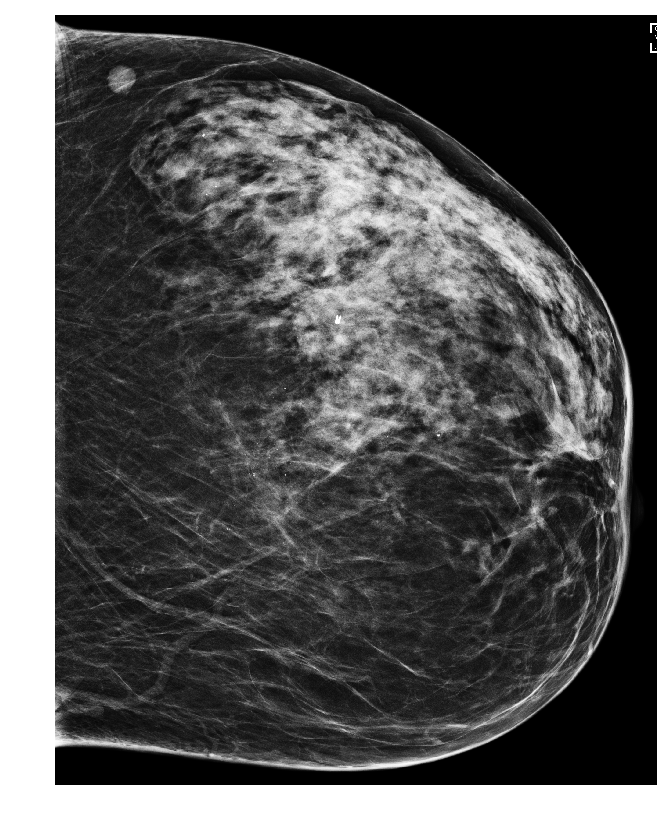}\hspace{-4mm}&
    \hspace{-5mm}\includegraphics[height = 0.15\textwidth, width = 0.10\textwidth, trim={0mm 0mm 0mm 0mm}]{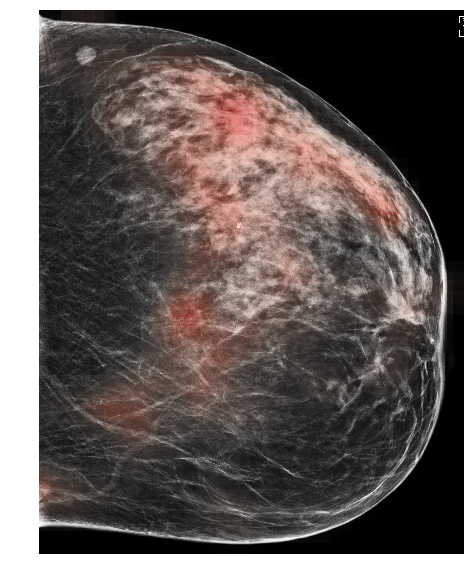}\hspace{-4mm}& \hspace{-5mm}\includegraphics[height = 0.15\textwidth, width = 0.10\textwidth, trim={0mm 0mm 0mm 0mm}]{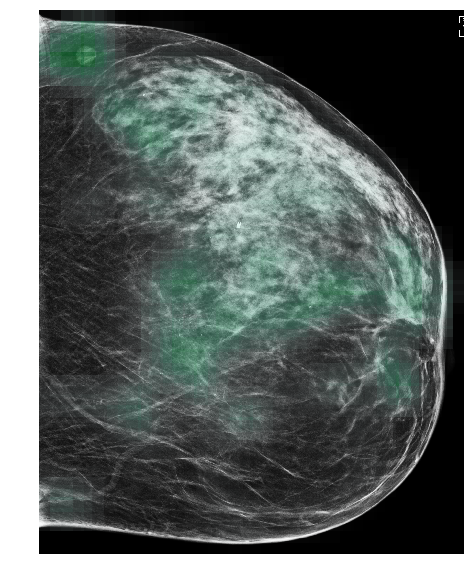}
    \end{tabular}&
    \begin{tabular}{c c c}
    \hspace{-5mm}
    \includegraphics[height = 0.15\textwidth, width = 0.10\textwidth, trim={0mm 0mm 0mm 0mm}]{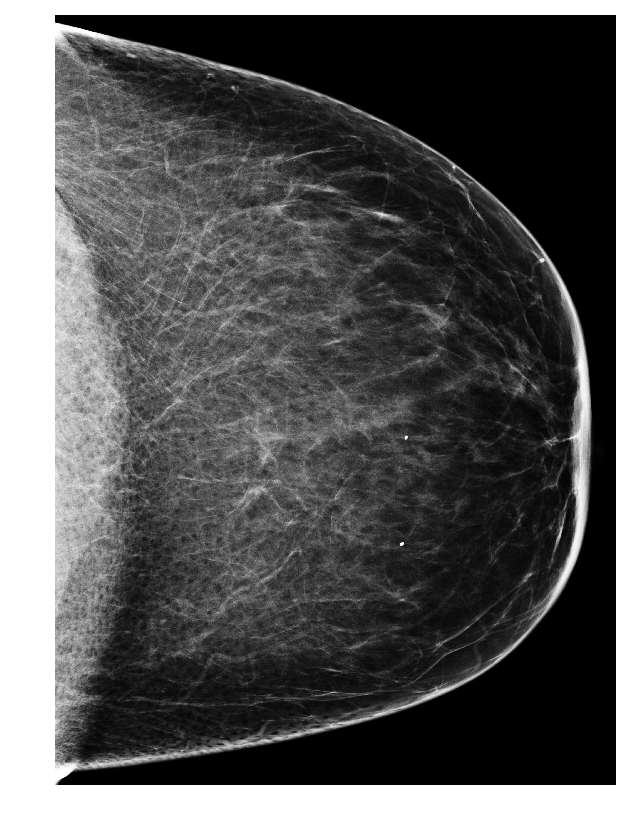}\hspace{-4mm}&
    \hspace{-5mm}\includegraphics[height = 0.15\textwidth, width = 0.10\textwidth, trim={0mm 0mm 0mm 0mm}]{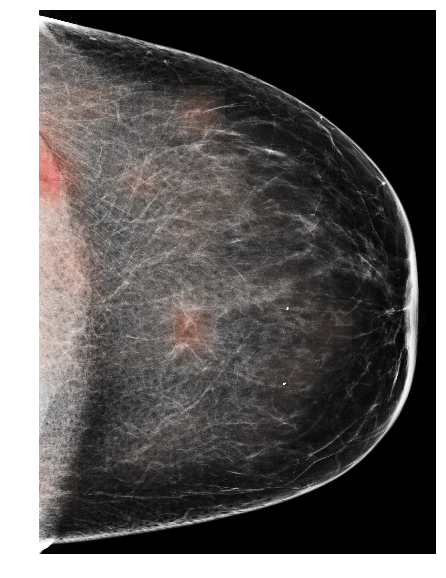}\hspace{-4mm}& \hspace{-5mm}\includegraphics[height = 0.15\textwidth, width = 0.10\textwidth, trim={0mm 0mm 0mm 0mm}]{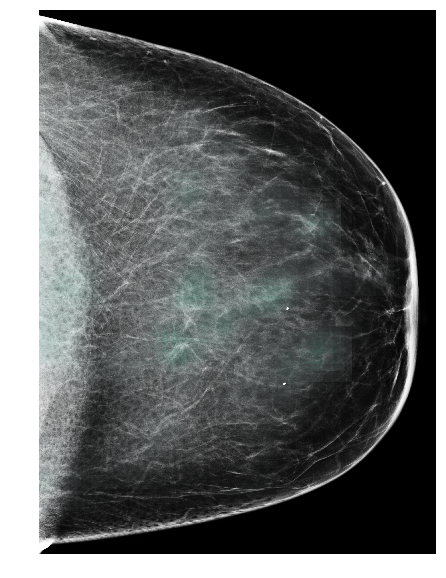}
    \end{tabular}
    \\
    
    \hspace{-6mm}\begin{tabular}{c}
    \rotatebox[origin=c]{270}{R-MLO}
    \end{tabular}&
    \begin{tabular}{c c c}
    \hspace{-8mm}
    \includegraphics[height = 0.15\textwidth, width = 0.10\textwidth,trim={0mm 0mm 0mm 0mm}]{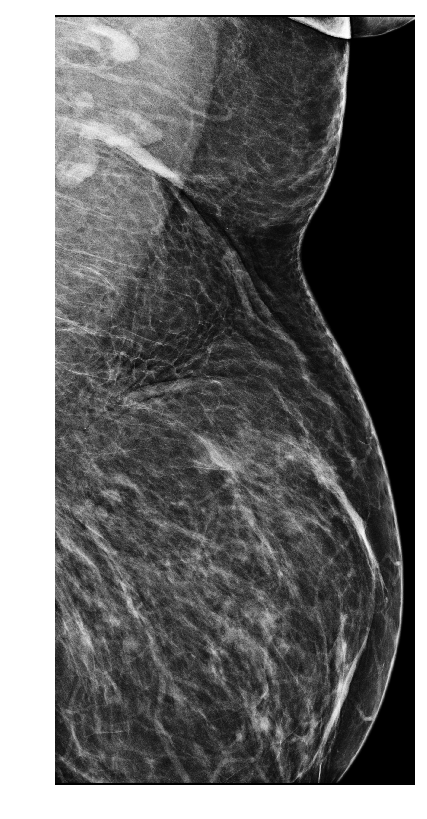}\hspace{-4mm}&
    \hspace{-5mm}\includegraphics[height = 0.15\textwidth, width = 0.10\textwidth, trim={0mm 0mm 0mm 0mm}]{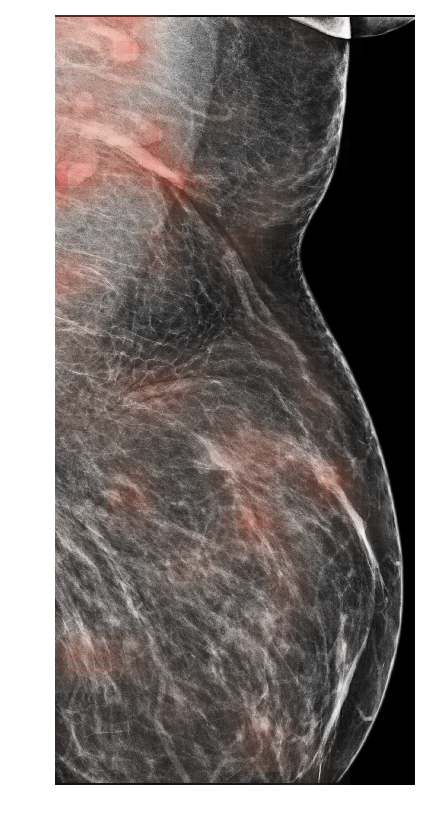}\hspace{-4mm}& \hspace{-5mm}\includegraphics[height = 0.15\textwidth, width = 0.10\textwidth, trim={0mm 0mm 0mm 0mm}]{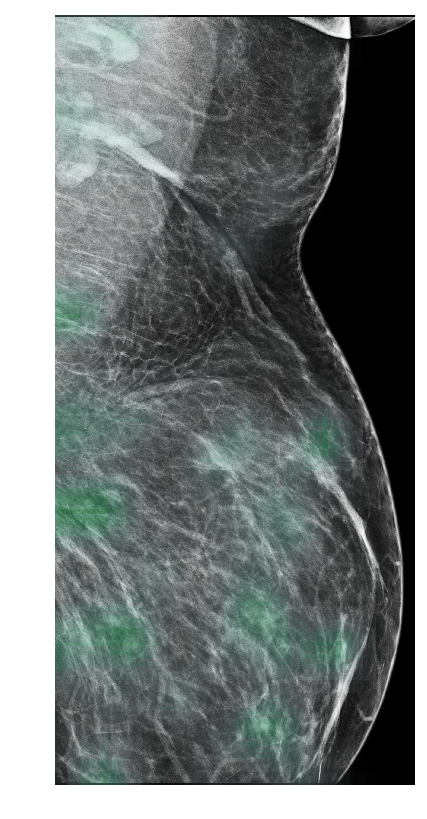}
    \end{tabular} & \begin{tabular}{c c c}
    \hspace{-5mm}
    \includegraphics[height = 0.15\textwidth, width = 0.10\textwidth,trim={0mm 0mm 0mm 0mm}]{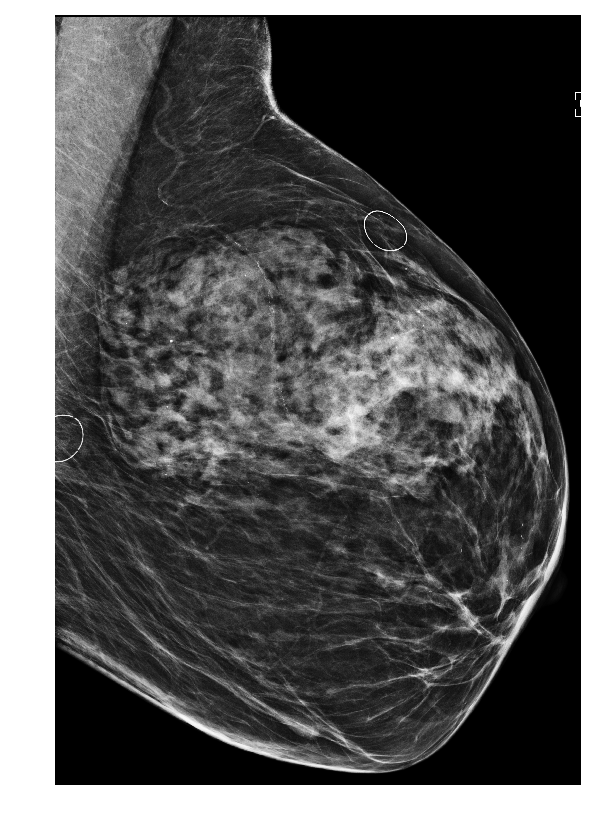}\hspace{-4mm}&
    \hspace{-5mm}\includegraphics[height = 0.15\textwidth, width = 0.10\textwidth,trim={0mm 0mm 0mm 0mm}]{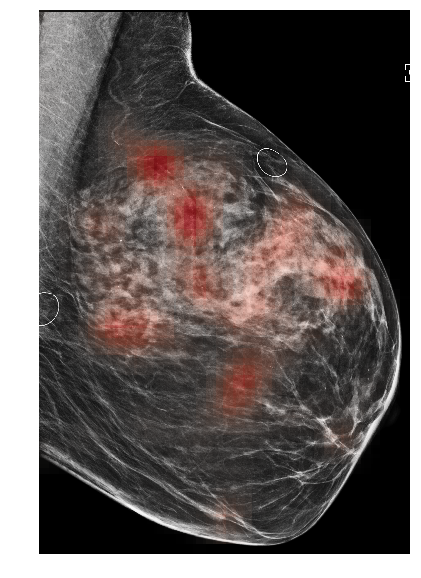}\hspace{-4mm}& \hspace{-5mm}\includegraphics[height = 0.15\textwidth, width = 0.10\textwidth, trim={0mm 0mm 0mm 0mm}]{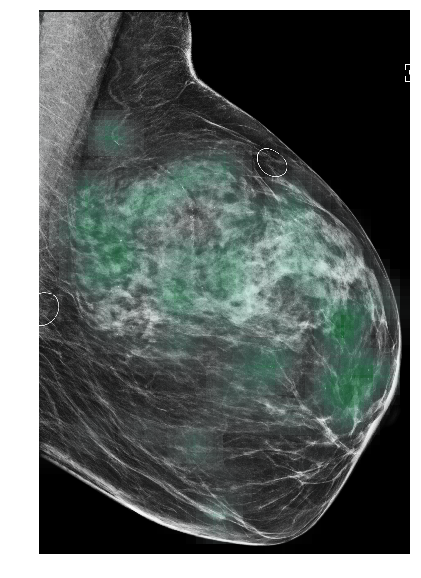}
    \end{tabular}& \begin{tabular}{c c c}
    \hspace{-5mm}
    \includegraphics[height = 0.15\textwidth, width = 0.10\textwidth, trim={0mm 0mm 0mm 0mm}]{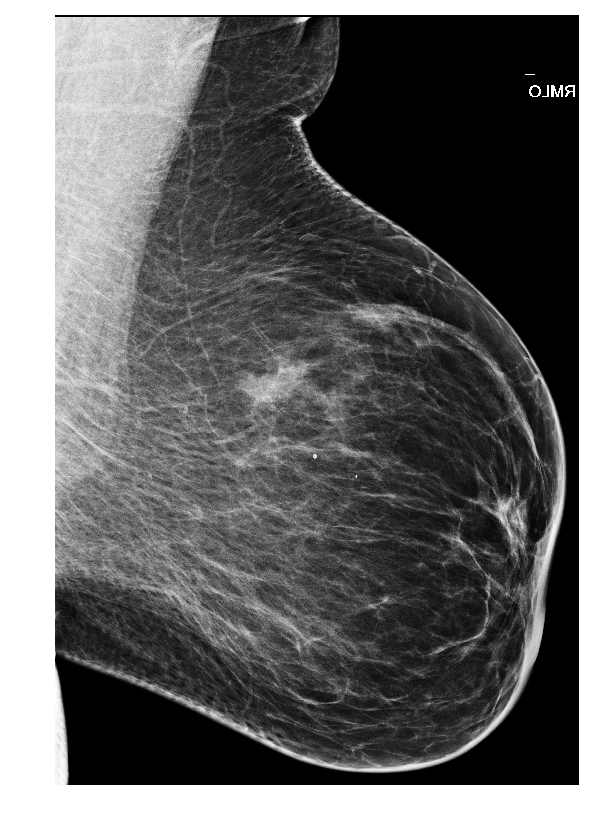}\hspace{-4mm}&
    \hspace{-5mm}\includegraphics[height = 0.15\textwidth, width = 0.10\textwidth, trim={0mm 0mm 0mm 0mm}]{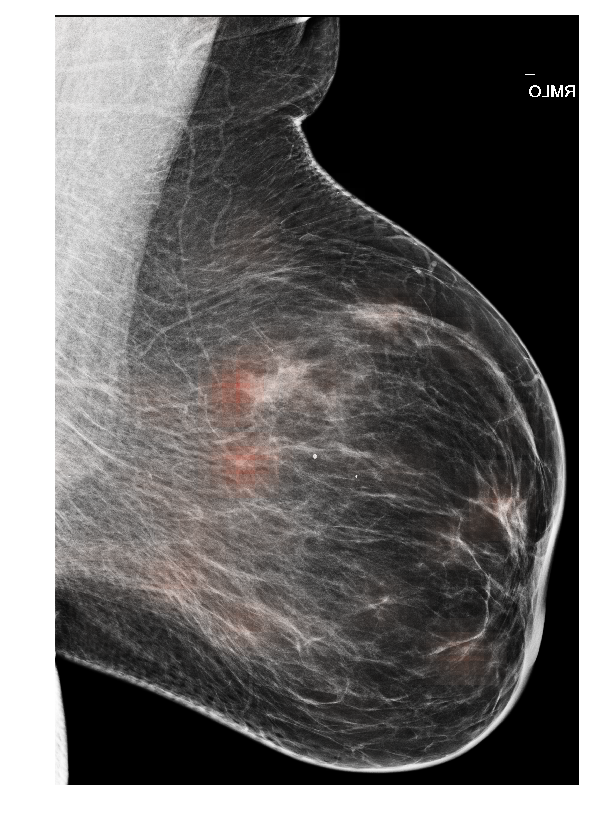}\hspace{-4mm}& \hspace{-5mm}\includegraphics[height = 0.15\textwidth, width = 0.10\textwidth, trim={0mm 0mm 0mm 0mm}]{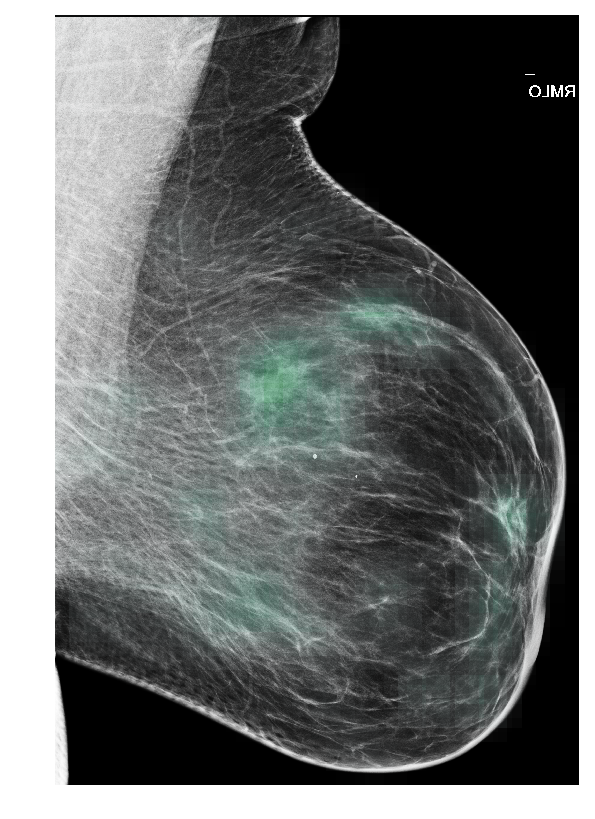}
    \end{tabular}\\
    
    \hspace{-6mm}\begin{tabular}{c}
    \rotatebox[origin=c]{270}{L-MLO}
    \end{tabular} &
    \begin{tabular}{c c c}
    \hspace{-8mm}
    \includegraphics[height = 0.15\textwidth, width = 0.10\textwidth, trim={0mm 0mm 0mm 0mm}]{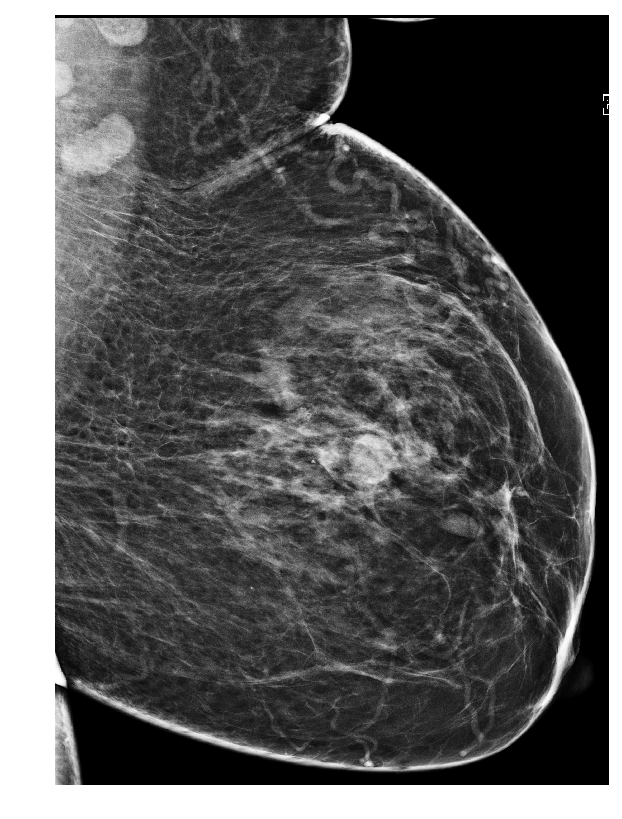}\hspace{-4mm}&
    \hspace{-5mm}\includegraphics[height = 0.15\textwidth, width = 0.10\textwidth, trim={0mm 0mm 0mm 0mm}]{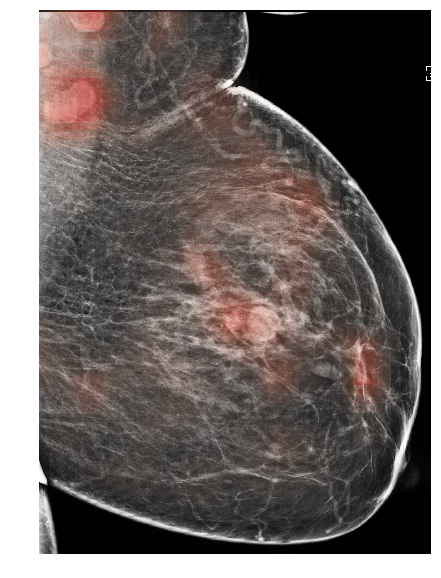}\hspace{-4mm}& \hspace{-5mm}\includegraphics[height = 0.15\textwidth, width = 0.10\textwidth, trim={0mm 0mm 0mm 0mm}]{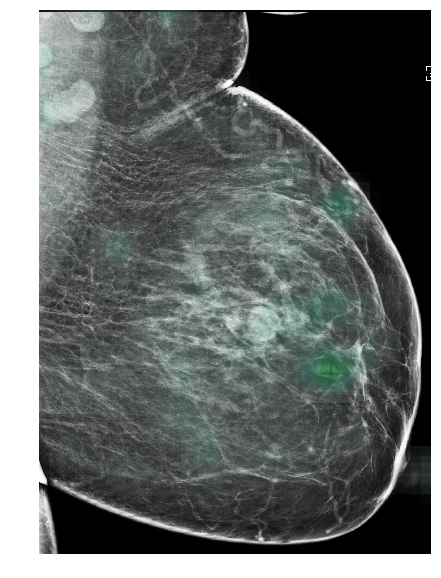}
    \end{tabular} & \begin{tabular}{c c c}
    \hspace{-4mm}
    \includegraphics[height = 0.15\textwidth, width = 0.10\textwidth, trim={0mm 0mm 0mm 0mm}]{figures/heatmaps/2769_L-MLO_i.png}\hspace{-4mm}&
    \hspace{-5mm}\includegraphics[height = 0.15\textwidth, width = 0.10\textwidth, trim={0mm 0mm 0mm 0mm}]{figures/heatmaps/2769_L-MLO_m.png}\hspace{-4mm}& \hspace{-5mm}\includegraphics[height = 0.15\textwidth, width = 0.10\textwidth, trim={0mm 0mm 0mm 0mm}]{figures/heatmaps/2769_L-MLO_b.png}
    \end{tabular}&
    \begin{tabular}{c c c}
    \hspace{-4mm}
    \includegraphics[height = 0.15\textwidth, width = 0.10\textwidth, trim={0mm 0mm 0mm 0mm}]{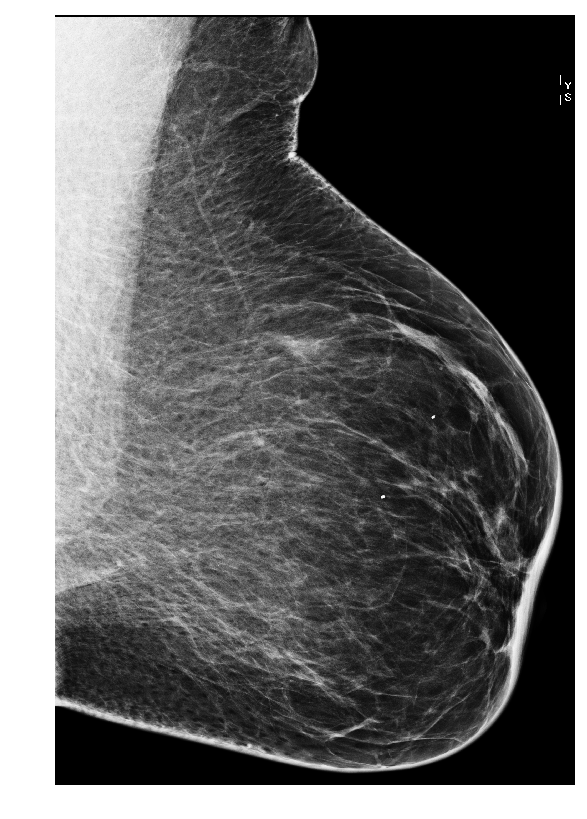}\hspace{-4mm}&
    \hspace{-5mm}\includegraphics[height = 0.15\textwidth, width = 0.10\textwidth, trim={0mm 0mm 0mm 0mm}]{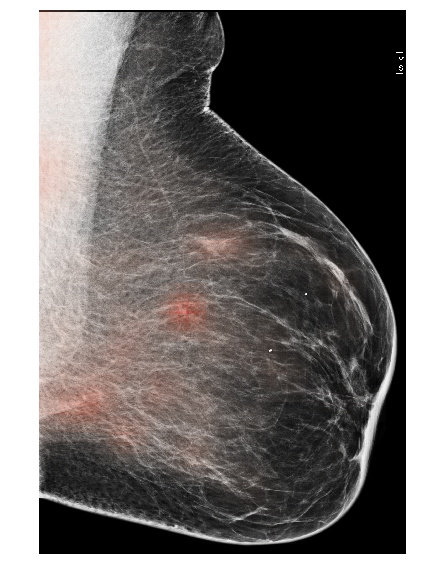}\hspace{-4mm}& \hspace{-5mm}\includegraphics[height = 0.15\textwidth, width = 0.10\textwidth, trim={0mm 0mm 0mm 0mm}]{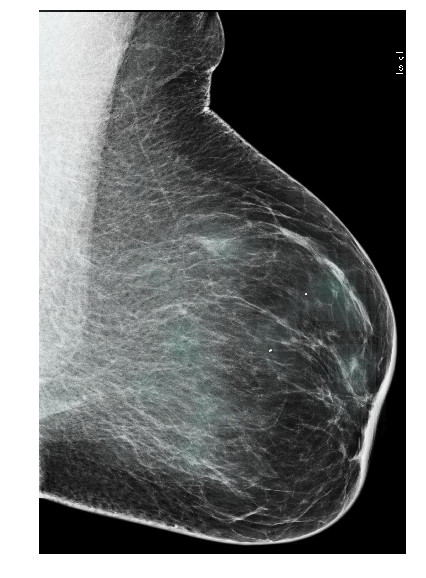}
    \end{tabular}
    \\
    
    & (a) & (b) & (c)
    \end{tabular}
    \vspace{-2mm}
    \caption{
    We select three exams from the test set and visualize the four standard views from each along with two heatmaps overlaid on the images. For each view, from left to right, we show: the original image, the image overlaid with a heatmap of the pixel-level prediction for malignancy, the image overlaid with a heatmap of the pixel-level prediction for benign findings. (a) An exam where the left breast was labeled as malignant as well as benign. (b) An exam in which there is a benign finding in the left breast. (c) An exam with benign findings in the right breast.}
    \label{fig:heatmaps_more}
\end{figure}

\subsubsection*{Model evaluation and selection} The main purpose of the patch-level classifier is to generate heatmaps which can be used as extra channels for the breast-level classifier. Unfortunately, it is hard to evaluate the patch-level classifier with respect to how it improves the breast-level model at each epoch.  We trained the patch-level network for  2,000 epochs, saving its parameters every 200 epochs. The 10 models saved were used to generate malignant and benign heatmaps for all images in the validation set. To form a breast-level prediction for the malignant/not malignant task from the heatmaps, we took the maximum value across the malignant heatmaps for each breast. The breast-level predictions for the benign/not benign task were computed analogously. The model we used for generating heatmaps for the entire data set was selected based on the average of the AUCs (between the two tasks) we obtained using these predictions. The process of generating the heatmaps for the entire dataset took approximately 1,000 hours using an Nvidia V100 GPU (2.12 seconds per image). Examples of the images with two corresponding heatmaps are shown in \autoref{fig:heatmaps_more}. 

\newpage
\bibliography{appendix}